%% file: main.tex
\documentclass[lettersize,journal]{IEEEtran}
\usepackage{amsmath,amsfonts}
\usepackage{algorithmic}
\usepackage{algorithm}
\usepackage{array}
\usepackage[caption=false,font=normalsize,labelfont=sf,textfont=sf]{subfig}
\usepackage{textcomp}
\usepackage{stfloats}
\usepackage{url}
\usepackage{verbatim}
\usepackage{graphicx}
\usepackage{cite}

\usepackage{booktabs}
\usepackage{multirow}

\usepackage{CJKutf8}  
\usepackage{tabularx}
\usepackage{pifont}  
\usepackage{xltabular}
\usepackage{makecell}   
\usepackage{enumitem}
\usepackage{colortbl} 
\usepackage[colorlinks,linkcolor=red,anchorcolor=blue,citecolor=blue]{hyperref}

\definecolor{lightgray}{gray}{0.9} 

\newcolumntype{C}[1]{>{\centering\arraybackslash}p{#1}} 
\newcolumntype{L}{>{\arraybackslash}X} 
\newcolumntype{Y}{>{\centering\arraybackslash}X} 
\newcolumntype{H}[1]{>{\centering\arraybackslash}m{#1}}

\usepackage[dvipsnames, svgnames, x11names]{xcolor}
\usepackage{bm}
\usepackage{amssymb}
\usepackage{pifont}
\newcommand{\Checkmark}{\ding{51}}
\newcommand{\Crossmark}{\ding{55}}

\usepackage{makecell}

\begin{document}

\title{UniMPA: A Unified Memory-Prediction-Action Model via Action-Grounded Transition Modeling}


\author{
    Wei Li,~Rui Shao,~Jie He,~Lingsen Zhang,~Ziwei Liu,~Liqiang Nie
    \thanks{
    Wei Li, Rui Shao, Jie He, Lingsen Zhang, Liqiang Nie are with the School of Computer Science and Technology, Harbin Institute of Technology (Shenzhen), China, 518055. E-mail: liwei2024@stu.hit.edu.cn, shaorui@hit.edu.cn, nieliqiang@gmail.com. 
    Ziwei Liu is with the S-Lab, Nanyang Technological University, Singapore, 639798. E-mail: ziwei.liu@ntu.edu.sg. \\
    Rui Shao is corresponding author.}
}

\markboth{IEEE Transactions on Pattern Analysis and Machine Intelligence}%
{Shell \MakeLowercase{\textit{Li et al.}}: UniMPA: A Unified Memory-Prediction-Action Model via Action-Grounded Transition Modeling}


\maketitle



\input{Sections/0-abstract}


\begin{figure*}[t]
\vspace{-4mm}
    \centering
    \includegraphics[width=1.0\textwidth]{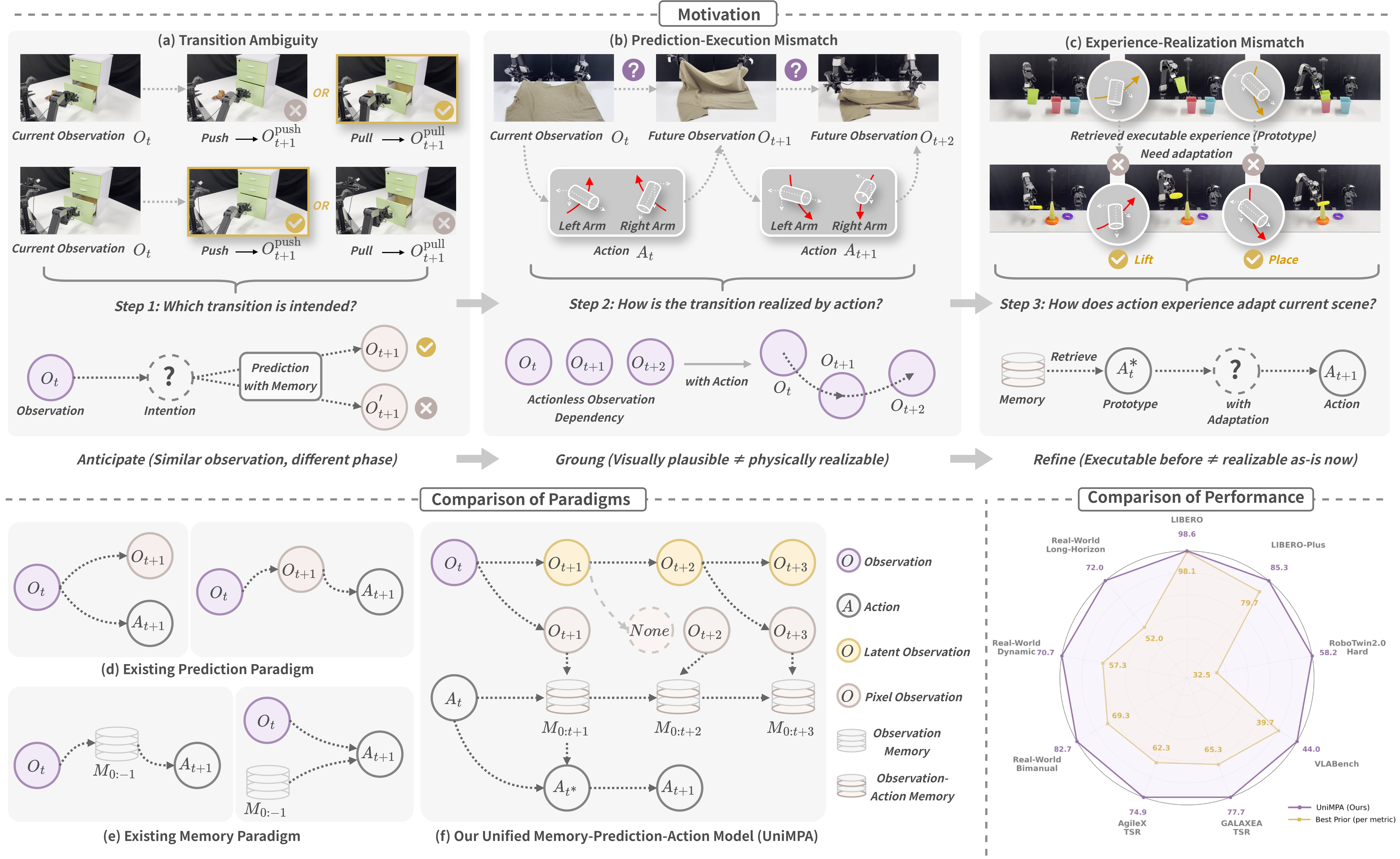}
    \vspace{-18pt}
     \caption{
    \textbf{Motivation and paradigm comparison.}
    Visually similar observations may imply different transitions, requiring anticipation of the intended change \textbf{(a)}; yet a visually plausible future may not be physically realizable and must be grounded in executable experience \textbf{(b)}, whose action prototype must further be adapted to the current scene rather than replayed as-is \textbf{(c)}.
    Existing prediction paradigms decouple predicted futures from historical executability \textbf{(d)}, whereas existing memory paradigms rely on observation-centric history without transition-action correspondence \textbf{(e)}.
    UniMPA \textbf{(f)} unifies Persistent-Selective Future Prediction, Bidirectional Memory, and Prototype-Biased Flow through an action-grounded transition interface, forming an \emph{anticipate--ground--refine} process.
    }
    \vspace{-1pt}
\label{fig:intro}
\end{figure*}

\input{Sections/1-introduction_v2}

\input{Sections/2-related_work}


\input{Sections/3-unimpa_v2}

\input{Sections/4-experiment}

\section{Conclusion}
\label{sec:con}
In this paper, we presented UniMPA, a unified Memory-Prediction-Action model
that reformulates robotic manipulation around action-grounded state transitions. 
By coupling future prediction with bidirectional vision-action memory, UniMPA grounds intended visual changes in executable experience and refines actions with retrieved prototypes, while persistent-selective prediction captures both long-horizon progress and interaction-critical details.
Extensive simulation and real-world experiments demonstrate strong robustness, generalization, and training efficiency.
More broadly, UniMPA suggests a shift from treating prediction, memory, and action generation as independent capabilities toward organizing them around a shared physical-transition interface. We hope this perspective inspires future embodied models to move beyond predicting \emph{what will happen} or recalling \emph{what happened}, and instead learn \emph{how intended world changes can be grounded in reusable executable experience}. 
Such transition-centric modeling may provide a promising foundation for more adaptive and continually improving robotic agents.

\bibliographystyle{IEEEtran}

\bibliography{main} 

\clearpage
\twocolumn[
\begin{center}
    {\Large\bfseries Appendix of UniMPA}
\end{center}
\vspace{1em}
]

\input{supp_arxiv}

\vfill

\end{document}

%% file: Sections/0-abstract.tex
\begin{abstract}
Recent advances in Vision-Language-Action (VLA) models have improved robotic manipulation, yet observation-to-action learning remains limited by a fundamental \textit{transition realizability gap}, manifested in three tightly coupled problems:
\textit{(i) Transition ambiguity.} 
Visually similar current observations may correspond to different manipulation phases and imply different subsequent transitions.
\textit{(ii) Prediction--execution mismatch.} A visually plausible predicted future observation does not necessarily correspond to a physically realizable transition. 
\textit{(iii) Experience--realization mismatch.}
A historically executable action pattern may not necessarily realize the intended transition in the current scene and therefore requires context-aware adaptation.
Accordingly, we propose \textit{UniMPA}, a Unified Memory-Prediction-Action model that addresses these problems through a shared \emph{action-grounded transition} interface. 
\textit{(i)} UniMPA introduces \textit{Persistent-Selective Future Prediction} to resolve transition ambiguity by modeling the intended future state evolution. A persistent latent stream continuously tracks task-level progress, while a transition-critical pixel stream selectively resolves fine-grained interaction changes through memory-grounded prediction.
\textit{(ii)} To assess the physical executability of the anticipated transition, the predicted transition queries a temporal \textit{Visual-Action Memory Bank}. The bank retrieves historically realized visual-action experience, grounding future prediction in executable evidence. 
\textit{(iii)} To adapt executable experience to the current scene, an \textit{Action-Visual Memory Bank} retrieves visually grounded action prototypes from historical action evolution. \textit{Prototype-Biased Flow} then shifts the flow source toward a historically supported action manifold for context-aware refinement. 
Together, these components form a unified \emph{anticipate--ground--refine} process from intended transition to executable action.
Extensive experiments show robust and generalizable manipulation with only \textit{25--50\%} of the training epochs of the $\pi_{0.5}$ baseline, outperforming it by 1.7, 11.7, 18.5, and 12.6 percentage points on LIBERO, LIBERO-Plus, RoboTwin 2.0 Hard, and real-world tasks, respectively. UniMPA Project page: \url{https://JiuTian-VL.github.io/UniMPA-page/}
\end{abstract}

%% file: Sections/1-introduction_v2.tex
\section{Introduction}

\IEEEPARstart{R}{ecent} Vision-Language-Action (VLA) models have substantially improved robotic manipulation through observation-conditioned action learning~\cite{Rt-2, openvla, pi05, song2026reconvla, hu2026abstraction, li2026cogvla, li2026global}. Yet their central bottleneck is not simply mapping a visual \textbf{observation} to an action, but inferring the task-relevant visual \textbf{transition} and realizing that change with an executable action pattern~\cite{wang2016actions, zeng2021transporter, souvcek2024genhowto}. This exposes a fundamental \emph{transition realizability gap}: \emph{how can a policy translate an intended future change into an action that can physically realize it in the current scene?} Fig.~\ref{fig:intro} decomposes this high-level question into three tightly coupled manifestations. \textit{\textbf{1) Transition ambiguity}}: similar observations may belong to different manipulation phases and imply different transitions (Fig.~\ref{fig:intro}(a)). \textit{\textbf{2) Prediction--execution mismatch}}: a visually plausible future need not be physically realizable through a coherent action sequence (Fig.~\ref{fig:intro}(b)). \textit{\textbf{3) Experience--realization mismatch}}: an action pattern executable in a historical scene may fail under the current geometry and require context-aware adaptation (Fig.~\ref{fig:intro}(c)). These are three manifestations of the same missing interface between intended state evolution and executable action generation.

Formally, let
$\mathbf{X}_t=\{\mathcal{V}_t,\boldsymbol{\ell},\mathbf{q}_t,\mathbf{H}_t\}$
denote the current visual, linguistic, proprioceptive, and action-history context, and let $\mathcal{V}_{t+\Delta}$ denote a future observation used as a training target. Rather than defining a transition as the additive image difference $\mathcal{V}_{t+\Delta}-\mathcal{V}_t$, we define the intended transition as a \emph{conditional state-evolution operator}
$\mathcal{T}^{\Delta}_t$ that maps the context toward a task-consistent future outcome. The action chunk $\mathbf{A}_t$ provides the physical realization of this conditional evolution:
\begin{equation}
\footnotesize
\mathcal{V}_t
\xrightarrow[
    \mathbf{A}_t(\cdot|\mathbf{X}_t)\,(\text{action realization})
]{
\mathcal{T}^{\Delta}_t(\cdot|\mathbf{X}_t)\,(\text{intended transition})
}
\mathcal{V}_{t+\Delta}\,(\text{supervised future outcome}).
\label{eq:transition_action_duality}
\end{equation}
This operator notation makes no linear or pixel-wise subtraction assumption. It instead characterizes \emph{how the current state is expected to evolve} under the task context.


However, existing methods generally fail to establish this coupling. Most VLA policies~\cite{diff, Rt-2, octo, cogact, openvla, oft, liu2024rdt, pi0, pi05, bjorck2025gr00t, spatialvla, li2026consisvla, zhu2026h} learn $\mathcal{V}_t\!\rightarrow\!\mathbf{A}_t$ directly and leave the intended transition implicit. Although prediction introduces future targets and memory incorporates historical context, they are typically treated as separate auxiliary mechanisms rather than jointly modeling the relation between intended transitions and executable action evolution. Consequently, increasing model capacity or temporal context alone cannot close the transition realizability gap.

First, observation-conditioned policies cannot resolve \emph{transition ambiguity}  (\textit{\textbf{limitation I}}). For example, in T-shirt folding, nearly identical layouts may correspond to fold initiation, continuation, grasp adjustment, or release. Because these phases require different transitions and action patterns~\cite{alayrac2017joint}, visual similarity alone provides an unreliable cue to the underlying task intention. The policy should model task evolution and identify the next transition, rather than forcing a static observation to simultaneously encode task phase and control intent.

Identifying the intended future transition, however, does not guarantee that it can be realized through feasible robot actions, giving rise to a \emph{prediction--execution mismatch} (\textit{\textbf{limitation II}}).
Bridging this mismatch requires future-supervised transition representations that capture both task-level progress and local interaction dynamics without collapsing the representation into a decoded future observation.
Existing future-prediction methods mainly operate at either the pixel or latent level. 
Pixel-level prediction~\cite{zhang2026dreamvla, cotvla, yuan2026fast} preserves local action effects but is dominated by static visual content, whereas latent-level prediction~\cite{assran2025v, sun2026vlajepa} captures task intention but may discard critical contact and geometric constraints. 
Neither alone reliably associates the predicted transition with action evolution. We use future latent and pixel targets to shape the pre-head World-Expert tokens, combining global task progress with local action-induced dynamics while exposing the resulting latent transition representation directly to action generation.

Beyond associating the intended transition with action evolution, historical experience can further improve realization by providing reusable priors from previously successful transitions. However, directly transferring such experience introduces an \emph{experience--realization mismatch} (\textit{\textbf{limitation III}}), since actions that were executable in past scenes may not remain realizable under the current poses or contacts. 
Prior memory methods mainly store observations~\cite{shi2025memoryvla, li2025map, shi2026memoryvla++}, keyframes or maps~\cite{zeng2026kemo, lin2025echovla}, and implicit histories~\cite{li2026remem, xiao2026ava}, while overlooking the transition--action correspondences required for phase-consistent executable experience retrieval.

These limitations point to a unified design principle in which prediction, memory, and action generation operate on the same action-grounded transition. Prediction disambiguates intent and forms the retrieval query, memory grounds it in executable visual--action evolution, and retrieved experience provides an action prior that is contextually refined. We therefore propose \textit{\textbf{UniMPA}}, a \textit{\textbf{Unified Memory-Prediction-Action model}} that follows an \emph{anticipate--ground--refine} process. Unlike the separated paradigms in Fig.~\ref{fig:intro}(d) and (e), UniMPA closes the loop among future, experience, and action in Fig.~\ref{fig:intro}(f), with its paradigm-level advantages summarized in Tab.~\ref{tab:intro}.

Specifically, UniMPA resolves the three aspects of the transition realizability gap with three correspondingly coupled designs. \textit{\textbf{1)}} To resolve transition ambiguity, \textit{\textbf{Persistent-Selective Future Prediction}} uses persistent latent supervision to encode semantic progress and long-term state evolution into the pre-head transition tokens, while a complementary \textit{\textbf{Trigger Gate}} activates pixel decoding only at transition-critical moments. The former prevents phase information from collapsing into a single observation, whereas the latter injects interaction-time-focused supervision around contact and release without the redundancy and reconstruction bias of dense pixel prediction. The decoded future is used only to supervise the transition tokens; the Action Expert directly consumes the pre-head tokens rather than the decoded future observation.
\textit{\textbf{2)}} To bridge the prediction--execution mismatch, the future-supervised transition representation is used to query a \textit{\textbf{Visual-Action Memory Bank}} of temporally aligned visual-action trajectories.
This transition-conditioned retrieval departs from conventional appearance-based matching by retrieving experience according to the expected state change and using historically realized action evolution as executable evidence to ground the predicted future.
\textit{\textbf{3)}} To bridge the experience--realization mismatch, an \textit{\textbf{Action-Visual Memory Bank}} exploits action history together with its aligned visual evolution to retrieve a visually grounded action prototype. 
Rather than directly executing the retrieved prototype, \textit{\textbf{Prototype-Biased Flow}} shifts the flow source toward a historically supported action manifold and lets the policy refine it for the current observation, task specification, and robot state.
UniMPA unifies transition prediction, experience grounding, and prototype-guided refinement into a coherent action-generation process that is both executable and context-adaptive.
Our contributions are summarized as follows:

\input{Tables/intro}

\begin{itemize}
\item We identify the \textit{\textbf{transition realizability gap}} as a central obstacle to observation-to-action learning and decompose this high-level scientific problem into transition ambiguity, prediction--execution mismatch, and experience--realization mismatch. Based on this analysis, we propose \textit{\textbf{UniMPA}}, a unified memory-prediction-action model that unifies anticipation, experience grounding, and action refinement through an \textit{\textbf{action-grounded transition}}.
\item We design \textit{\textbf{Persistent-Selective Future Prediction}}, which uses latent and selectively triggered pixel targets to shape content-free World-Expert queries into future-supervised transition tokens. These pre-head tokens encode manipulation progress and interaction-critical changes and condition the Action Expert during training and inference.
\item We introduce bidirectional \textit{\textbf{Visual-Action}} and \textit{\textbf{Action-Visual Memory Banks}}, together with \textit{\textbf{Prototype-Biased Flow}}. They ground predicted transitions in historically executable evidence, construct visually grounded action prototypes, and adapt those prototypes from a supported action manifold to the current scene.
\item Extensive simulation and real-world experiments demonstrate that UniMPA achieves robust and generalizable manipulation, attaining state-of-the-art performance across the evaluated benchmarks while requiring only $25$--$50\%$ of the training epochs used by the $\pi_{0.5}$ baseline.

\end{itemize}


\begin{figure*}[t]
\vspace{-4mm}
\centering
\includegraphics[width=1.0\textwidth]{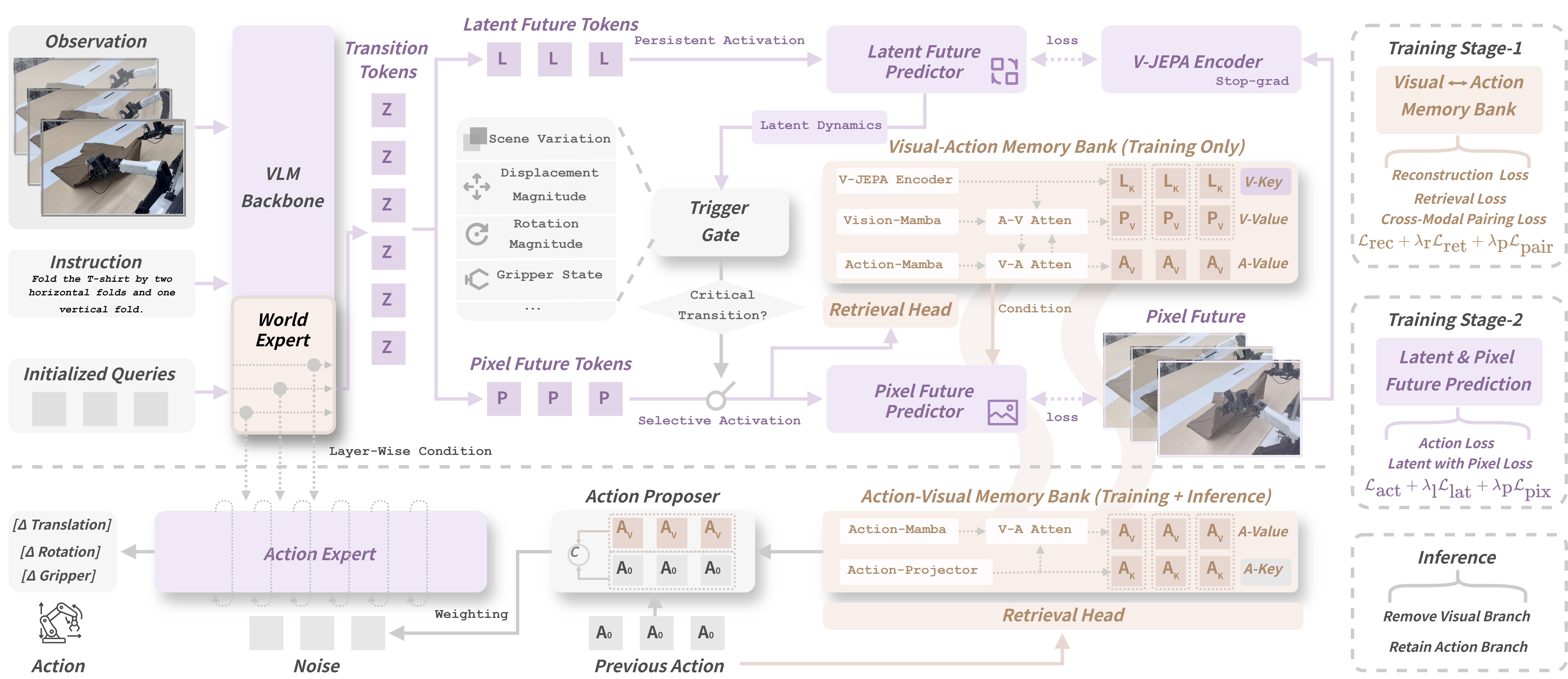} 
\caption{
\textbf{Overall framework of UniMPA.}
UniMPA couples future-supervised transition modeling, bidirectional visual-action memory, and action generation through an action-grounded transition.
\textit{\textbf{1)}} The World Expert transforms zero-initialized transition queries into latent transition tokens conditioned on the current context; training-only latent and trigger-gated pixel heads supervise these tokens with future outcomes.
\textit{\textbf{2)}} The future-supervised transition representation queries the Visual-Action Memory Bank for action-grounded visual experience, while historical actions query the Action-Visual Memory Bank to construct executable action prototypes.
\textit{\textbf{3)}} The retrieved prior biases the initial flow distribution and is subsequently refined by the action expert.
\textit{\textbf{4)}} The memory banks are pretrained in Stage~1 and frozen during policy training in Stage~2. During inference, only the explicit latent/pixel decoding heads are removed; the World Expert and its transition tokens remain active and, together with the retrieved action-manifold prior, condition action generation.
}
\label{fig:framework}
\end{figure*}

%% file: Tables/intro.tex
\begin{table*}[t]
\vspace{-5mm}
\caption{Paradigm-level comparison in modeling design and performance (libero and libero-plus). Data Cost denotes that UniMPA achieves superior performance using only 25-50\% of the training epochs. P and L denote pixel- and latent-level prediction.}
\label{tab:intro}
\centering
\setlength{\tabcolsep}{3.3pt}
\begin{tabular}{lcccccc|ccccc}
\toprule
{Paradigm}
& \makecell{{Action}\\{Basis}}
& \makecell{{Future}\\{Supervision}}
& \makecell{{Experience}\\{Memory}}
& \makecell{{Pre-Mem}\\{Coupling}}
& \makecell{{Trans-Act}\\{Grounding}}
& \makecell{{Temporal}\\{Modeling}}
& \makecell{{Representative}\\{Baseline}}
& \makecell{{Base}\\{Performance}}
& \makecell{{Generalization}\\{Performance}}
& \makecell{{Training}\\{Budget}}
\\
\midrule

General
& $\mathcal{V}_t$
& \Crossmark
& \Crossmark
& \Crossmark
& \Crossmark
& \Crossmark
& $\pi_{0.5}$~\cite{pi05}
& 96.9\%
& 73.6\%
& 100\%
\\

Prediction-Based
& $\mathcal{V}_t$
& \Checkmark (P or L)
& \Crossmark
& \Crossmark
& \Crossmark
& \Checkmark~(\textit{heavy})
& Fast-WAM~\cite{yuan2026fast}
& 97.6\%
& 59.0\%
& $\sim$266\%
\\

Memory-Based
& $\mathcal{V}_t$
& \Crossmark
& \Checkmark ($\mathcal{V}$)
& \Crossmark
& \Crossmark
& \Crossmark
& MemoryVLA~\cite{shi2025memoryvla}
& 96.5\%
& 70.2\%
& $\sim$333\%
\\

\rowcolor[HTML]{EFEFEF}
{UniMPA (Ours)}
& $\mathcal{T}^{\Delta}_t$\
& \Checkmark (P + L)
& \Checkmark ($\mathcal{V}$ + $\mathbf{A}$)
& \Checkmark
& \Checkmark
& \Checkmark~(\textit{lightweight})
& -
& \textbf{98.6\%}
& \textbf{85.3\%}
& \textbf{25-50\%}
\\

\bottomrule
\end{tabular}
\vspace{-4pt}

\end{table*}

%% file: Sections/2-related_work.tex
\section{Related Work}
\label{sec:rw}
\subsection{Single- and Dual-System VLA Models}
 Vision-Language-Action models can generally be categorized into single- and dual-system architectures. 
\textit{\textbf{1) Single-system models}}~\cite{shao2025large, Rt-2, openvla, oft, zheng2024tracevla, spatialvla, li2026semanticvla} directly extend pretrained  Vision-Language models (VLMs)~\cite{team2026gemma, peng2026survey, yang2025qwen3, llama3, shao2024detecting, shao2023detecting, shao2019multi, xiao2026survey, zhou2026hiconagent, zhu2026uniemo, shao2026hats, lyu2026personalalign, li2025lion} to predict robot actions autoregressively~\cite{openvla, hung2025nora, tan2025FLASHVLA, zhang2026vtla} or in parallel~\cite{oft, pdvla, li2025cogvla, molevla}, preserving semantic consistency within a unified vision-language-action representation space~\cite{Rt-2, openvla, spatialvla}. However, autoregressive action decoding incurs substantial computational overhead, while modeling semantic reasoning and continuous control within the same representation space introduces a pronounced modality gap. 
\textit{\textbf{2) Dual-system models}} instead couple a VLM backbone for vision-language understanding with a dedicated diffusion~\cite{bu2024towards, cogact, cui2025openhelix, wen2024diffusionvla} or flow-matching action expert~\cite{pi0, pi05, bjorck2025gr00t, shukor2025smolvla, yuan2026qwen}, improving the efficiency and expressiveness of continuous action modeling. Nevertheless, the two modules are typically linked only through generic hidden features, leaving a representational gap between high-level reasoning and low-level control.
UniMPA extends this paradigm with a world expert, forming a \textit{\textbf{three-system architecture}} in which an action-grounded transition interface bridges the VLM and action expert using predicted physical changes and executable historical experience.

\subsection{Pixel- and Latent-Level Future Prediction}
Future prediction has been introduced into robot policies to expose task progress and physical dynamics beyond the current observation. 
\textit{\textbf{1) Pixel-level methods}} predict future images, videos, or dynamic regions~\cite{du2023learning, hu2024video, zhang2026dreamvla, cotvla, zhang2025up, liu2026oa}, providing explicit supervision for local interaction changes but often allocating substantial capacity to static backgrounds, textures, and appearance variations. 
\textit{\textbf{2) Latent-level methods}} instead predict compact future representations in pretrained visual or multimodal feature spaces~\cite{assran2025v, sun2026vlajepa, luo2026being, miao2026jepavla, wang2026latentvla, bai2026latent}, offering more stable and efficient modeling of semantic progress and long-horizon dynamics, while potentially overlooking fine-grained changes such as contact and release. 
Existing approaches generally adopt one prediction level with a uniform activation strategy. 
UniMPA instead combines persistent latent prediction with trigger-gated pixel prediction, continuously tracking execution progress while selectively introducing fine-grained visual supervision only at critical transition moments.

\subsection{From Vision-Language-Action to World-Action Modeling}
Conventional Vision-Language-Action models primarily learn observation-conditioned action generation, whereas recent World-Action Models (WAMs)~\cite{wu2024unleashing, cen2025worldvla, cotvla, jang2025dreamgen, cen2025rynnvla} enhance robot control by explicitly modeling future environmental states. DreamZero~\cite{ye2026world} instantiates a WAM using a pretrained video diffusion backbone to jointly generate future videos and robot actions. More generally, Wang \emph{et al.}~\cite{wang2026world} define WAMs as models that jointly characterize future states and actions:
\begin{equation}
\footnotesize
\mathcal{L}_{\mathrm{WAM}}
=
\mathbb{E}_{(\mathbf{O_t},\mathbf{L},\mathbf{O_{t+1}},\mathbf{A_t})\sim\mathcal{D}}
\left[
-\log p(\mathbf{O_{t+1}},\mathbf{A_t} \mid \mathbf{O_t},\mathbf{L})
\right]
\end{equation}
where $\mathbf{O_t}$, $\mathbf{L}$, $\mathbf{O_{t+1}}$, and $\mathbf{A_t}$ denote the current observation, language instruction, future observation, and corresponding action sequence, respectively. 
This architecture-agnostic definition also encompasses VLM-based approaches.
Despite this broader formulation, many existing predictive WAMs employ future prediction primarily as an auxiliary~\cite{bi2026motus, yuan2026fast} or cascaded objective~\cite{li2026causal, feng2025vidar, guo2025ctrl}, rather than as a retrieval interface grounded in historically executable experience.
UniMPA shifts future modeling from generic future reconstruction toward transition modeling. 
It further couples action generation with the anticipated transition through temporal alignment with historically executable transition-action experience, transforming future modeling into an action-grounded interface.

\subsection{Memory-Augmented Robot Manipulation}
Memory mechanisms have recently been introduced into VLA models to support long-horizon, history-dependent, and non-Markovian manipulation. 
Existing methods retrieve historical observations or demonstration-derived prompts~\cite{shi2025memoryvla, li2025map, shi2026memoryvla++}, construct keyframes or spatial-semantic memories~\cite{zeng2026kemo, lin2025echovla, chen2026non, li2026spatial}, and recurrent or episodic history representations~\cite{torne2026mem, sridhar2025memer, li2026remem, xiao2026ava}. 
These methods provide useful temporal context, but memory retrieval is predominantly organized around observations, semantic states, or implicit history representations rather than the correspondence between a visual transition and the action evolution that physically realizes it. Such observation-centric retrieval can be insufficient when visually similar configurations occur at different manipulation phases and therefore require different subsequent transitions. Moreover, even a historically successful action sequence cannot generally be transferred directly to a new scene because object poses, contact conditions, and geometric constraints may have changed.
UniMPA instead builds complementary Visual-Action and Action-Visual Memory Banks that store temporally aligned transition-action experience, enabling predicted visual transitions to
retrieve executable action evidence and its associated historical action evolution.

%% file: Sections/3-unimpa_v2.tex
\begin{figure*}[t]
\vspace{-4mm}
\centering
\includegraphics[width=1.0\textwidth]{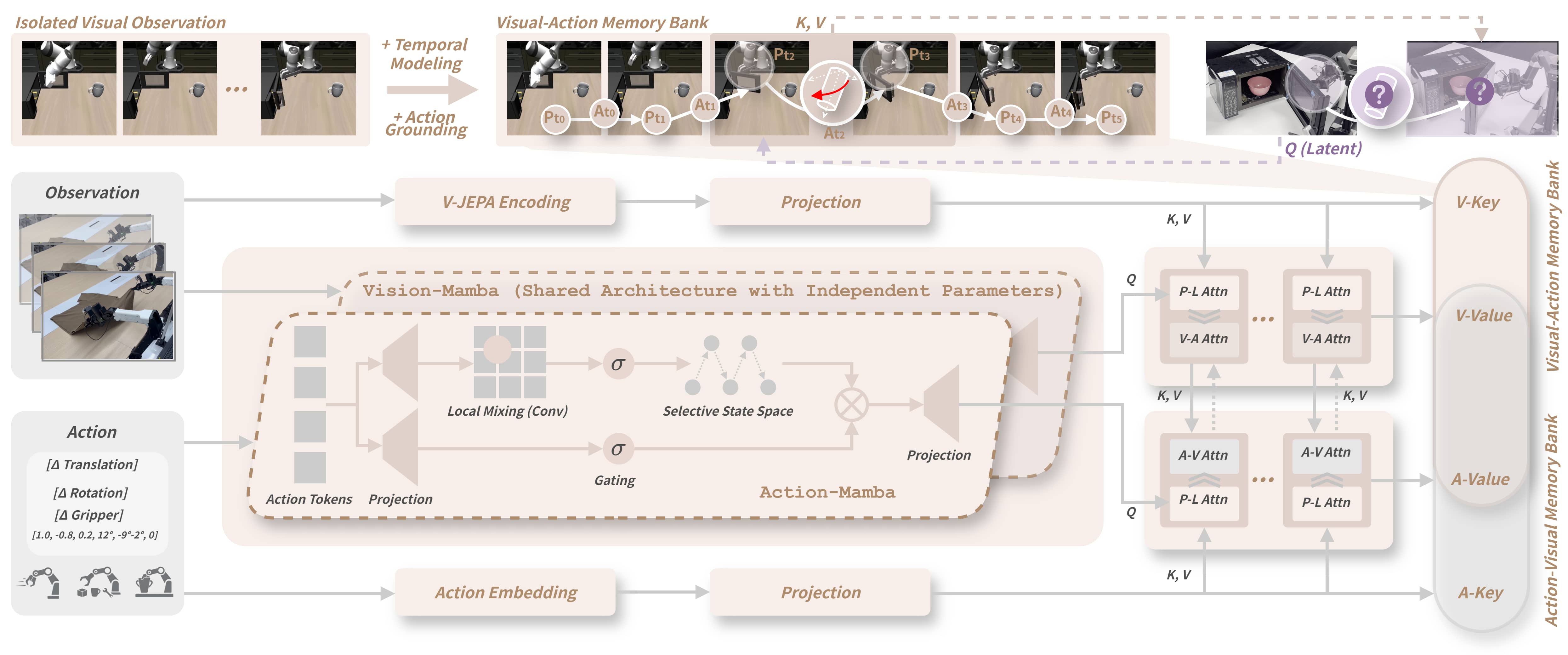} 
\vspace{-24pt}
    \caption{\textbf{Transition-aligned bidirectional memory construction.} 
    Historical trajectories are organized into complementary \textit{\textbf{Visual-Action}} and \textit{\textbf{Action-Visual Memory Banks}}, where transition-aware keys support retrieval and temporally modeled values preserve aligned visual-action evolution.
    Stateful Vision-Mamba and Action-Mamba modules encode episode-level temporal correspondence, while cross-stream transition grounding injects action-conditioned dynamics into visual memories and visual-transition semantics into action memories.
    The resulting bidirectional memories establish a transition-level interface: future-supervised transition queries retrieve executable visual--action evolution, while historical actions provide visually grounded action priors.
    }
\label{fig:bank}
\end{figure*}

\section{UniMPA}
\label{sec:UniMPA}
\subsection{Problem Formulation and Framework Overview}
\label{sec:framework}
Given a robot manipulation trajectory, UniMPA takes as input the current multi-view observation, language instruction, robot proprioceptive state, and historical action context. Formally, at timestep $t$, we define the input context as
\begin{equation}
\small
    \mathbf{X}_t=\{\mathcal{V}_t,\boldsymbol{\ell},\mathbf{q}_t,\mathbf{H}_t\},
    \quad 
    \mathbf{H}_t=[\mathbf{a}_{t-M},\dots,\mathbf{a}_{t-1}],
\end{equation}
where $\mathcal{V}_t=\{\mathbf{I}^{\nu}_t\}_{\nu\in\mathcal{U}}$ contains the RGB observations from the camera set $\mathcal{U}$, $\mathbf{q}_t$ denotes the robot proprioceptive state, $\boldsymbol{\ell}$ is the language instruction, and $\mathbf{H}_t\in\mathbb{R}^{M\times D}$ denotes the historical action window. We use $\mathbf{C}_t=\operatorname{Enc}_{c}(\mathcal{V}_t,\boldsymbol{\ell})$ to denote the current vision-language prefix produced by the VLM backbone. The target future action chunk is
\begin{equation}
\small
\mathbf{A}_t
=
\left[
\mathbf{a}_t,
\dots,
\mathbf{a}_{t+K-1}
\right]
\in
\mathbb{R}^{K\times D},
\label{eq:target_action}
\end{equation}
where $M$ and $K$ are the history length and prediction horizon, respectively. In our implementation, $D=7$, corresponding to the end-effector translation, rotation, and gripper command.
For embodiments controlled in joint space rather than Cartesian end-effector space, we represent actions using joint-position commands, as detailed in \textbf{Appendix A}.

Following Eq.~\eqref{eq:transition_action_duality}, UniMPA does not instantiate the intended transition as an observation difference. Instead, it learns a latent realization of the conditional operator $\mathcal{T}^{\Delta}_t$: the current context $\mathbf{X}_t$ is the model input, the future observation $\mathcal{V}_{t+\Delta}$ is used only through training losses, and the World-Expert features form the representation supplied to action generation.

As illustrated in Fig.~\ref{fig:framework}, UniMPA contains three tightly coupled components:
\textit{\textbf{1)}} Persistent-Selective Future Prediction, which uses future targets to shape a latent transition representation throughout the trajectory and selectively decodes pixels at critical transitions;
\textit{\textbf{2)}} Visual-Action Memory Bank, which uses a future-supervised transition query to retrieve historically executable visual-transition evidence; and
\textit{\textbf{3)}} Action-Visual Memory Bank, which retrieves visually grounded action prototypes and uses them to bias the action manifold. For a flow timestep $\rho\in[0,1]$, the noisy action input $\mathbf{x}_{\rho}$ is embedded as
\begin{equation}
\small
\mathbf{E}^{a}_t
=
\phi_a
\left(
\mathbf{x}_{\rho},
\rho,
\mathbf{q}_t
\right),
\qquad
\mathbf{E}^{w}_t
=
\mathbf{Q}_0
=
\mathbf{0}
\in
\mathbb{R}^{N_w\times d},
\label{eq:expert_inputs}
\end{equation}
where $\mathbf{E}^{a}_t$ denotes the action tokens and $\mathbf{E}^{w}_t=\mathbf{Q}_0$ denotes $N_w$ content-free, zero-initialized transition queries. 

UniMPA organizes the VLM backbone, World Expert, and Action Expert as three interacting Transformer streams with hidden states $\mathbf{H}^{c}_{l}$, $\mathbf{H}^{w}_{l}$, and $\mathbf{H}^{a}_{l}$. Each stream $r\in\{c,w,a\}$ forms standard projections $\mathbf{Q}^{r}_{l}=\mathbf{H}^{r}_{l}\mathbf{W}^{r}_{Q,l}$, $\mathbf{K}^{r}_{l}=\mathbf{H}^{r}_{l}\mathbf{W}^{r}_{K,l}$, and $\mathbf{V}^{r}_{l}=\mathbf{H}^{r}_{l}\mathbf{W}^{r}_{V,l}$. Rather than introducing separate equations for these standard operations, we directly write the cross-stream update with concatenated keys and values as
\begin{equation}
{
\footnotesize
\left\{
\begin{array}{@{}l@{}}
\widetilde{\mathbf{H}}^{r}_{l}
=
\mathbf{H}^{r}_{l}
+
\operatorname{Softmax}\!\left(
\frac{
\mathbf{Q}^{r}_{l}
[\mathbf{K}^{c}_{l};\mathbf{K}^{w}_{l};\mathbf{K}^{a}_{l}]^{\top}
}{
\sqrt{d}
}
+\mathbf{M}^{r}_{l}
\right)
[\mathbf{V}^{c}_{l};\mathbf{V}^{w}_{l};\mathbf{V}^{a}_{l}],
\\[3pt]
\mathbf{H}^{r}_{l+1}
=
\widetilde{\mathbf{H}}^{r}_{l}
+
\operatorname{FFN}^{r}_{l}
\!\left(
\widetilde{\mathbf{H}}^{r}_{l}
\right),
\end{array}
\right.
}
\label{eq:tri_attention}
\end{equation}
where $\mathbf{M}^{r}_{l}$ is a block mask controlling information exchange among the three systems. This interaction implements the proposed action-grounded transition interface. At each layer, the World-Expert stream reads the current vision-language context $\mathbf{C}_t$ and transforms it into world keys and values. 
We denote the final transition tokens and action tokens as
\begin{equation}
\small
\mathbf{Z}^{\mathrm{tr}}_t
=
\mathbf{H}^{w}_{L},
\qquad
\mathbf{Z}^{a}_t
=
\mathbf{H}^{a}_{L}.
\label{eq:transition_and_action_tokens}
\end{equation}


\subsection{Transition-Aligned Bidirectional Memory Pretraining}

UniMPA first pretrains a structured memory bank from historical robot trajectories $\mathcal{T}^{e}=\{\mathcal{V}^{e}_{t},\mathbf{q}^{e}_{t},\mathbf{a}^{e}_{t}\}_{t=1}^{T_e}$. Unlike conventional memories that store observations, keyframes, or implicit history features independently, UniMPA explicitly associates visual state evolution with the action patterns that induce it. The resulting memory therefore acts as a transition-level interface between future prediction and action generation, as shown in Fig.~\ref{fig:bank}. At timestep $t$, the bidirectional entry is
\begin{equation}
\small
\mathcal{M}_{e,t}
=
\left\{
(\mathbf{k}^{\nu}_{v,e,t},\mathbf{u}^{\nu}_{v,e,t})_{\nu\in\mathcal{U}},
(\mathbf{k}_{a,e,t},\mathbf{u}_{a,e,t}),e,t
\right\},
\label{eq:memory_entry}
\end{equation}
where $\mathbf{k}^{\nu}_{v,e,t}$ and $\mathbf{u}^{\nu}_{v,e,t}$ denote the visual key and value for view $\nu$, while $\mathbf{k}_{a,e,t}$ and $\mathbf{u}_{a,e,t}$ denote the action key and value. 
Episode and timestep indices are retained for temporally coherent retrieval. The {{Visual-Action Memory Bank}} retrieves action-aware transition experience from future-supervised transition queries, while the {{Action-Visual Memory Bank}} retrieves visually grounded action prototypes from historical action evolution.

\subsubsection{Visual and Action Key Construction}
\label{sec:memory_keys}
For each view $\nu$, a frozen V-JEPA~2 encoder~\cite{assran2025v} first extracts compact visual latent representations, which are subsequently projected into $N_\nu$ memory tokens of dimension $d_m$. Mean pooling followed by $\ell_2$ normalization yields the visual retrieval key:
\begin{equation}
\small
\mathbf{P}^{\nu}_t
=
\psi_v^\nu(f_{\mathrm{JEPA}}(\mathbf{I}^{\nu}_t))
\in\mathbb{R}^{N_\nu\times d_m},
~~
\mathbf{k}^{\nu}_{v,t}
=
\operatorname{norm}_2(
\frac{1}{N_\nu}\sum_{i=1}^{N_\nu}\mathbf{P}^{\nu}_{t,i}
).
\label{eq:visual_key}
\end{equation}
Although computed from the current observation, this key is optimized by the future-oriented reconstruction and retrieval-simulation objectives below, rather than serving as a generic appearance descriptor. Likewise, the recent action history is projected into memory tokens and summarized as
\begin{equation}
\small
\mathbf{P}^{a}_t
=
\psi_a(\mathbf{H}_t)
\in\mathbb{R}^{M\times d_m},
~~
\mathbf{k}_{a,t}
=
\operatorname{norm}_2(
\frac{1}{M}\sum_{i=1}^{M}\mathbf{P}^{a}_{t,i}
).
\label{eq:action_key}
\end{equation}
The visual key identifies the observation-conditioned origin of a physical transition, while the action key characterizes its recent execution phase. Both key spaces are subsequently shaped by future reconstruction and simulated retrieval rather than appearance or action similarity alone.

\subsubsection{Temporal Value Encoding}
Keys provide retrieval indices, while memory values preserve temporal dynamics. To model episode-level state evolution, UniMPA employs stateful Mamba modules~\cite{gu2023mamba} for both visual and action streams. For camera view $\nu$, the visual temporal state is updated as
\begin{equation}
\small
\left(
\mathbf{S}^{\nu}_{v,t},
\mathbf{h}^{\nu}_{v,t}
\right)
=
\operatorname{Mamba}^{\nu}_{v}
\left(
\operatorname{Patch}
\left(
\mathbf{I}^{\nu}_{t}
\right),
\mathbf{h}^{\nu}_{v,t-1}
\right),
\label{eq:visual_mamba}
\end{equation}
where $\mathbf{h}^{\nu}_{v,t}$ denotes the recurrent state maintained throughout the episode, and the action temporal state is computed by
\begin{equation}
\small
\left(
\mathbf{S}_{a,t},
\mathbf{h}_{a,t}
\right)
=
\operatorname{Mamba}_{a}
\left(
\mathbf{P}^{a}_{t},
\mathbf{h}_{a,t-1}
\right).
\label{eq:action_mamba}
\end{equation}

This stateful formulation enables the memory values to distinguish visually similar observations according to their preceding interaction history. To further inject action dynamics into the visual memory, UniMPA first conditions the visual temporal states on both the action states and semantic visual tokens, and then fuses the resulting action-conditioned representation with the original visual temporal state:
\begin{equation}
\small
\left\{
\begin{aligned}
\mathbf{R}^{\nu}_{v,t}
&=
\operatorname{Attn}
\left(
\mathbf{S}^{\nu}_{v,t},
\operatorname{Concat}
\left[
\mathbf{S}_{a,t},
\mathbf{P}^{\nu}_t
\right]
\right),
\\
\mathbf{u}^{\nu}_{v,t}
&=
\Phi_v^\nu
\left(
\operatorname{Concat}
\left[
\mathbf{S}^{\nu}_{v,t},
\mathbf{R}^{\nu}_{v,t}
\right]
\right).
\label{eq:visual_value}
\end{aligned}
\right.
\end{equation}
Here, $\mathbf{S}^{\nu}_{v,t}$ serves as the queries, while the concatenated action-visual representation provides the keys and values. The resulting $\mathbf{R}^{\nu}_{v,t}$ captures action-conditioned visual context, which is further integrated with the original temporal state to construct the memory value $\mathbf{u}^{\nu}_{v,t}$. 

Conversely, UniMPA grounds the action memory in multi-view visual dynamics. Specifically, it first conditions the action temporal state on the multi-view visual temporal representations and historical action tokens, and then fuses the resulting visually grounded context with the original action state:
\begin{equation}
\small
\left\{
\begin{aligned}
\mathbf{R}_{a,t}
&=
\operatorname{Attn}
\left(
\mathbf{S}_{a,t},
\operatorname{Concat}
\left[
\operatorname{Concat}_{\nu\in\mathcal{U}}
\left(
\mathbf{S}^{\nu}_{v,t}
\right),
\mathbf{P}^{a}_t
\right]
\right),
\\
\mathbf{u}_{a,t}
&=
\Phi_a
\left(
\operatorname{Concat}
\left[
\mathbf{S}_{a,t},
\mathbf{R}_{a,t}
\right]
\right).
\label{eq:action_value}
\end{aligned}
\right.
\end{equation}


\subsubsection{Future-Oriented Memory Reconstruction}
The memory bank is trained with decoding and retrieval simulation objectives. Visual values are decoded to future images, and action values are decoded to future action chunks:
\begin{equation}
    \hat{\mathbf{I}}^{\nu}_{t+\Delta}
    =
    D_v^\nu(\mathbf{u}^{\nu}_{v,t}),
    \quad
    \hat{\mathbf{A}}_{t}
    =
    D_a(\mathbf{u}_{a,t}),
\end{equation}
where $\Delta$ denotes the future prediction offset.

The corresponding reconstruction objective is
\begin{equation}
\small
    \mathcal{L}_{\text{rec}}
    =
    \sum_{\nu\in\mathcal{U}}
    \frac{1}{|\Omega_\nu|}
    \left\|
    \hat{\mathbf{I}}^{\nu}_{t+\Delta}
    -
    \mathbf{I}^{\nu}_{t+\Delta}
    \right\|_1
    +
    \lambda_a
    \frac{1}{KD}
    \left\|
    \hat{\mathbf{A}}_{t}
    -
    \mathbf{A}_{t}
    \right\|_1,
\end{equation}
where $|\Omega_\nu|$ is the number of image elements in view $\nu$. The visual reconstruction term encourages visual values to retain future evolution, while the action reconstruction term requires action values to preserve executable future motion patterns.

\subsubsection{Retrieval-Simulation Training}
Reconstruction alone does not ensure effective retrieval. UniMPA therefore simulates retrieval during pretraining to jointly optimize the key and value spaces.
For a query memory entry $i$, let $\mathcal{C}(i)$ denote its candidate set. For memory modality $r\in\{v,a\}$, the similarity between query $i$ and candidate $j$ is
\begin{equation}
\small
s^{r}_{ij}
=
\operatorname{sim}
\left(
\mathbf{k}^{r}_{i},
\mathbf{k}^{r}_{j}
\right)
=
\mathbf{k}^{r\top}_{i}
\mathbf{k}^{r}_{j},
\label{eq:key_similarity}
\end{equation}
where all keys are $\ell_2$-normalized. 
A soft-retrieval distribution and its corresponding retrieved value are then defined as
\begin{equation}
\small
\begin{aligned}
\omega^{r}_{ij}
&=
\frac{
\exp\left(s^{r}_{ij}/\tau_r\right)
}{
\sum_{m\in\mathcal{C}(i)}
\exp\left(s^{r}_{im}/\tau_r\right)
},
\qquad
\overline{\mathbf{u}}^{r}_{i}
=
\sum_{j\in\mathcal{C}(i)}
\omega^{r}_{ij}
\mathbf{u}^{r}_{j},
\end{aligned}
\label{eq:soft_retrieval}
\end{equation}
where $\tau_r$ denotes the modality-specific retrieval temperature.

For the visual memory, Eq.~\eqref{eq:soft_retrieval} is independently applied to each camera view, yielding $\overline{\mathbf{u}}^{\nu}_{v,t}$. For the action memory, it produces $\overline{\mathbf{u}}_{a,t}$. 
The retrieved values are decoded using the same future decoders employed for memory reconstruction:
\begin{equation}
\footnotesize
    \mathcal{L}_{\text{ret}}
    =
    \sum_{\nu\in\mathcal{U}}
    \frac{1}{|\Omega_\nu|}
    \left\|
    D_v^\nu(\bar{\mathbf{u}}^{\nu}_{v,t})
    -
    \mathbf{I}^{\nu}_{t+\Delta}
    \right\|_1
    +
    \lambda_{ra}
    \frac{1}{KD}
    \left\|
    D_a(\bar{\mathbf{u}}_{a,t})
    -
    \mathbf{A}_{t}
    \right\|_1 .
\end{equation}

Unlike conventional metric-learning objectives that directly impose pairwise similarity in the embedding space, $\mathcal{L}_{\mathrm{ret}}$ evaluates whether the retrieved value predicts the correct future. 

To further align the two memory directions, we introduce a cross-modal pairing objective that explicitly associates visual and action values corresponding to the same physical transition. Since both memory values are token sequences, we first mean-pool and $\ell_2$-normalize them:
\begin{equation}
\footnotesize
    \widetilde{\mathbf{u}}^{\nu}_{v,i}
    =
    \operatorname{norm}_{2}(
    \frac{1}{N^{v}_{i}}
    \sum_{n=1}^{N^{v}_{i}}
    \mathbf{u}^{\nu}_{v,i,n}
    ),
    ~~
    \widetilde{\mathbf{u}}_{a,i}
    =
    \operatorname{norm}_{2}(
    \frac{1}{N^{a}_{i}}
    \sum_{n=1}^{N^{a}_{i}}
    \mathbf{u}_{a,i,n}
    ).
    \label{eq:value_pair_pooling}
\end{equation}
For each visual view $\nu$, the similarity between visual entry $i$ and action entry $j$ is defined as
$S^{\nu}_{ij}=\operatorname{cos}(\widetilde{\mathbf{u}}^{\nu}_{v,i},\widetilde{\mathbf{u}}_{a,j})/\tau_{c}$,
where $\tau_c$ is the cross-modal temperature. The temporally aligned pair $(i,i)$ serves as the positive sample, while the remaining valid entries in the current training batch serve as negatives. 
We optimize the correspondence bidirectionally:
\begin{equation}
\footnotesize
    \left\{
    \begin{aligned}
    \mathcal{L}^{\nu}_{v\rightarrow a}
    &=
    -\frac{1}{N}
    \sum_{i=1}^{N}
    \log
    \frac{\exp(S^{\nu}_{ii})}
    {\sum_{j=1}^{N}\exp(S^{\nu}_{ij})},\\
    \mathcal{L}^{\nu}_{a\rightarrow v}
    &=
    -\frac{1}{N}
    \sum_{i=1}^{N}
    \log
    \frac{\exp(S^{\nu}_{ii})}
    {\sum_{j=1}^{N}\exp(S^{\nu}_{ji})}.
    \end{aligned}
    \right.
\end{equation}
Here, $\mathcal{L}^{\nu}_{v\rightarrow a}$ matches each visual representation to its temporally paired action representation, while $\mathcal{L}^{\nu}_{a\rightarrow v}$ performs the reverse matching.
The cross-modal pairing loss averages these two directional objectives across all camera views:
\begin{equation}
\small
\mathcal{L}_{\mathrm{pair}}
=
\frac{1}{2|\mathcal{U}|}
\sum_{\nu\in\mathcal{U}}
\left(
\mathcal{L}^{\nu}_{v\rightarrow a}
+
\mathcal{L}^{\nu}_{a\rightarrow v}
\right).
\label{eq:cross_modal_pairing}
\end{equation}
The two objectives provide complementary supervision, with $\mathcal{L}_{\mathrm{ret}}$ encouraging future-consistent retrieval and $\mathcal{L}_{\mathrm{pair}}$ aligning visual and action values associated with the same physical transition.
We jointly optimize future reconstruction, retrieval simulation, and cross-modal pairing to learn predictive and cross-modally aligned memory representations.
The complete Stage-1 memory pretraining objective is
\begin{equation}
\mathcal{L}_{\mathrm{bank}}
=
\mathcal{L}_{\mathrm{rec}}
+
\lambda_{\mathrm{ret}}\mathcal{L}_{\mathrm{ret}}
+
\lambda_{\mathrm{pair}}\mathcal{L}_{\mathrm{pair}}.
\end{equation}



\begin{figure}[t]
\centering
\includegraphics[width=0.5\textwidth]{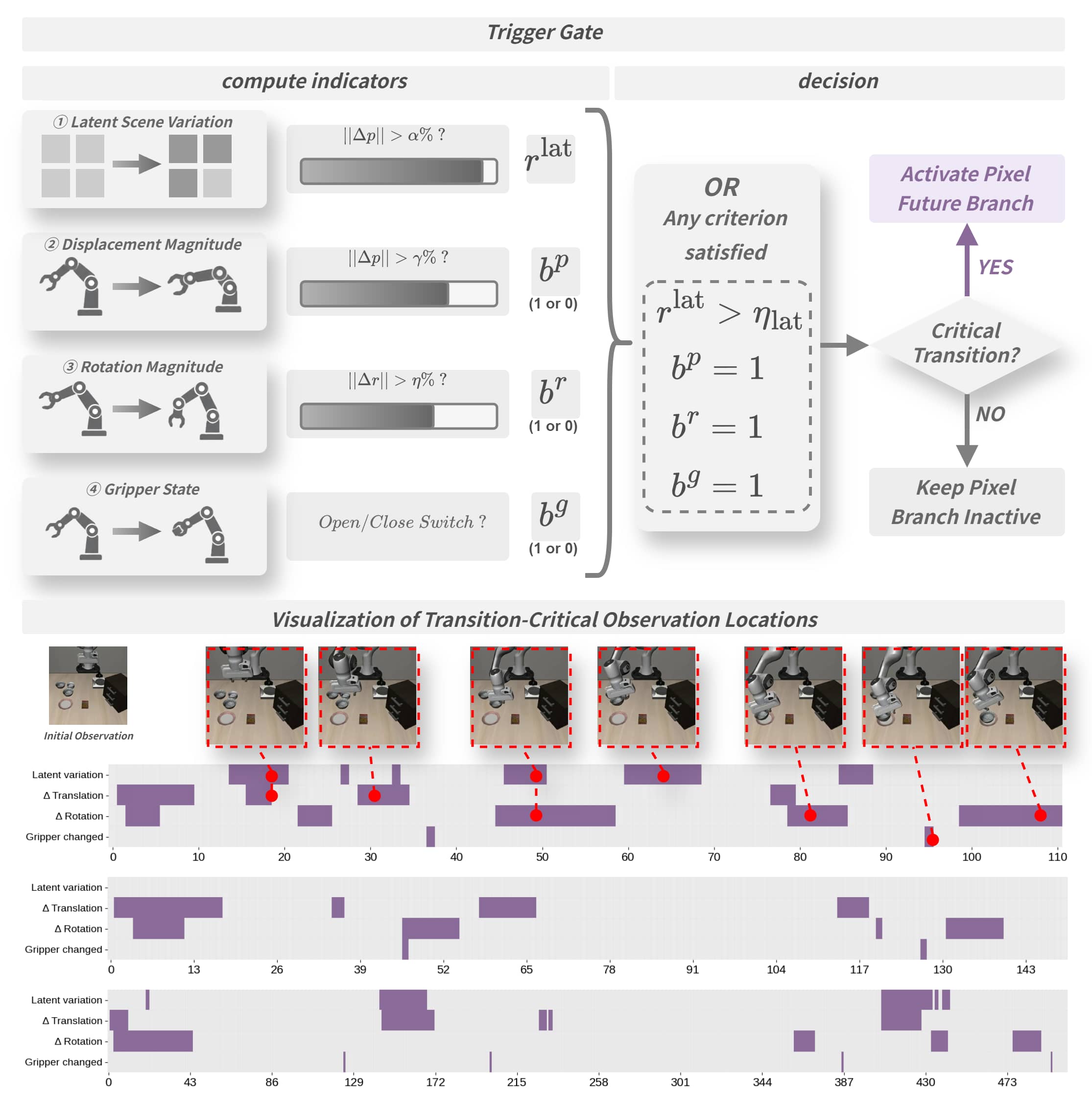} 
\vspace{-18pt}
    \caption{\textbf{Transition-Critical Trigger Gate.} \textbf{Top:} The gate combines latent semantic variation with action-derived translation, rotation, and gripper-state indicators. Pixel-level future prediction is activated when any criterion indicates a critical transition. 
    \textbf{Bottom:} Visualization of transition-critical observation locations on representative trajectories. The highlighted timestamps show frames selected by the Trigger Gate. The selected frames cluster around interaction-intensive moments, such as reaching, contact, pose adjustment, and gripper switching, showing that the trigger gate concentrates pixel-level supervision on task-relevant transitions while avoiding redundant prediction. }
\label{fig:gate}
\end{figure}

\subsubsection{Coarse-to-Fine Temporal Indexing}
\label{sec:memory_indexing}

A global index is constructed over all memory keys to efficiently identify relevant episodes. However, globally aggregating values from unrelated trajectories may mix different task phases or execution strategies. UniMPA therefore retains the episode identities and temporal indices of all memory entries and adopts a coarse-to-fine retrieval strategy. Given a query $\mathbf{q}_t$, the coarse stage first selects the most relevant episode:
\begin{equation}
\small
e^{\ast}
=
\operatorname*{arg\,max}_{e}
\;
\max_{s}
\operatorname{sim}
\left(
\mathbf{q}_t,
\mathbf{k}_{e,s}
\right).
\label{eq:coarse_episode_retrieval}
\end{equation}
The fine stage subsequently performs soft retrieval only over the selected episode:
\begin{equation}
\small
\overline{\mathbf{u}}_t
=
\sum_{s=1}^{T_{e^{\ast}}}
\frac{
\exp
\left(
\operatorname{sim}
\left(
\mathbf{q}_t,
\mathbf{k}_{e^{\ast},s}
\right)
/\tau
\right)
}{
\sum_{j=1}^{T_{e^{\ast}}}
\exp
\left(
\operatorname{sim}
\left(
\mathbf{q}_t,
\mathbf{k}_{e^{\ast},j}
\right)
/\tau
\right)
}
\mathbf{u}_{e^{\ast},s}.
\label{eq:fine_episode_retrieval}
\end{equation}

This episode-constrained aggregation prevents the retrieved memory from averaging visually similar but temporally incompatible states across unrelated demonstrations. 
This episode constraint improves temporal consistency while reducing the fine-search domain from the full memory bank to $T_{e^\ast}$ entries.


\subsection{Persistent-Selective Future Prediction}

In contrast to methods that treat a decoded future as the representation supplied to the policy, UniMPA uses future prediction to supervise the transition tokens. Latent supervision remains active throughout the trajectory, whereas the pixel-decoding head is selectively activated at critical moments. The resulting $\mathbf{Z}^{\mathrm{tr}}_t$ remains part of action generation even when the explicit decoding heads are removed at inference.

\subsubsection{Future-Supervised Latent Transition Modeling}
\label{sec:latent_transition_modeling}
The World Expert outputs pre-head transition tokens $\mathbf{Z}^{\mathrm{tr}}_t$. These tokens are not themselves a decoded future state. During training, a lightweight prediction head projects them into the V-JEPA~2 latent space for each camera view:
\begin{equation}
\small
    \hat{\mathbf{F}}^{\nu}_{t+\Delta}
    =
    h_{\text{lat}}^\nu(\mathbf{Z}^{\mathrm{tr}}_t),
    \quad
    \nu\in\mathcal{U}.
\end{equation}
The latent future prediction loss is defined by aligning the predicted latent with the target future V-JEPA~2 representation:
\begin{equation}
\footnotesize
    \mathcal{L}_{\text{lat}}
    =
    \sum_{\nu\in\mathcal{U}}
    \frac{1}{N_\nu d_f}
    \left\|
    \hat{\mathbf{F}}^{\nu}_{t+\Delta}
    -
    \text{sg}\left( f_{\mathrm{JEPA}}
    \left(
    \mathbf{I}^{\nu}_{t+\Delta}
    \right)
    \right)
    \right\|_2^2.
\end{equation}
where $N_\nu$ denotes the number of latent tokens, $d_f$ is their feature dimension, and $\operatorname{sg}(\cdot)$ denotes stop-gradient. Since the zero-initialized queries can minimize this objective only by reading the current context and encoding information predictive of its subsequent evolution, $\mathcal{L}_{\mathrm{lat}}$ shapes $\mathbf{Z}^{\mathrm{tr}}_t$ into a future-supervised latent transition representation. 

\subsubsection{Transition-Critical Trigger Gate}
Pixel changes become particularly important near physical interaction events, such as establishing contact, closing the gripper, adjusting orientation, or releasing the object. UniMPA introduces a training-only Trigger Gate that combines predicted semantic change with action-derived motion events, as illustrated in Fig.~\ref{fig:gate}.

The semantic transition score is computed from the predicted latent displacement of the main camera view:
\begin{equation}
\small
r^{\mathrm{lat}}_t
=
\frac{
\left\|
\hat{\mathbf{F}}^{\mathrm{main}}_{t+\Delta}
-
\mathbf{F}^{\mathrm{main}}_t
\right\|_2
}{
\left\|
\mathbf{F}^{\mathrm{main}}_t
\right\|_2
+
\epsilon
},
\label{eq:latent_transition_score}
\end{equation}
where $\epsilon$ prevents numerical instability; a large $r^{\mathrm{lat}}_t$ indicates substantial predicted visual-state change. Decomposing each action as $\mathbf{a}_{t+k}=[\Delta\mathbf{p}_{t+k},\Delta\mathbf{r}_{t+k},g_{t+k}]$, with translation, rotation, and gripper components respectively, the corresponding transition indicators are
\begin{equation}
    \small
    \left\{
    \begin{aligned}
    b^{p}_t
    &=
    \mathbf{1}
    [
    \max_{0\leq k<K}
    \left\|
    \Delta\mathbf{p}_{t+k}
    \right\|_2
    >
    \eta_p
    ],
    \\
    b^{r}_t
    &=
    \mathbf{1}
    [
    \max_{0\leq k<K}
    \left\|
    \Delta\mathbf{r}_{t+k}
    \right\|_2
    >
    \eta_r
    ],
    \\
    b^{g}_t
    &=
    \mathbf{1}
    [
    \max_{0\leq k<K}
    \left|
    g_{t+k}
    -
    g_{t+k-1}
    \right|
    >
    \eta_g
    ].
    \end{aligned}
    \right.
    \label{eq:transition_trigger}
\end{equation}
where $\eta_p$, $\eta_r$, and $\eta_g$ are modality-specific thresholds.

The final binary gate is
\begin{equation}
\small
m_t
=
\mathbf{1}
\left[
r^{\mathrm{lat}}_t
>
\eta_{\mathrm{lat}}
\;\lor\;
b^{p}_t=1
\;\lor\;
b^{r}_t=1
\;\lor\;
b^{g}_t=1
\right],
\label{eq:binary_trigger_gate}
\end{equation}
where $\eta_{\mathrm{lat}}$ controls the sensitivity to semantic state changes. 


The latent score and action indicators provide complementary semantic and motor evidence, enabling the gate to more reliably capture diverse critical transitions beyond either visual change or action magnitude alone.

\begin{figure}[t]
\centering
\includegraphics[width=0.48\textwidth]{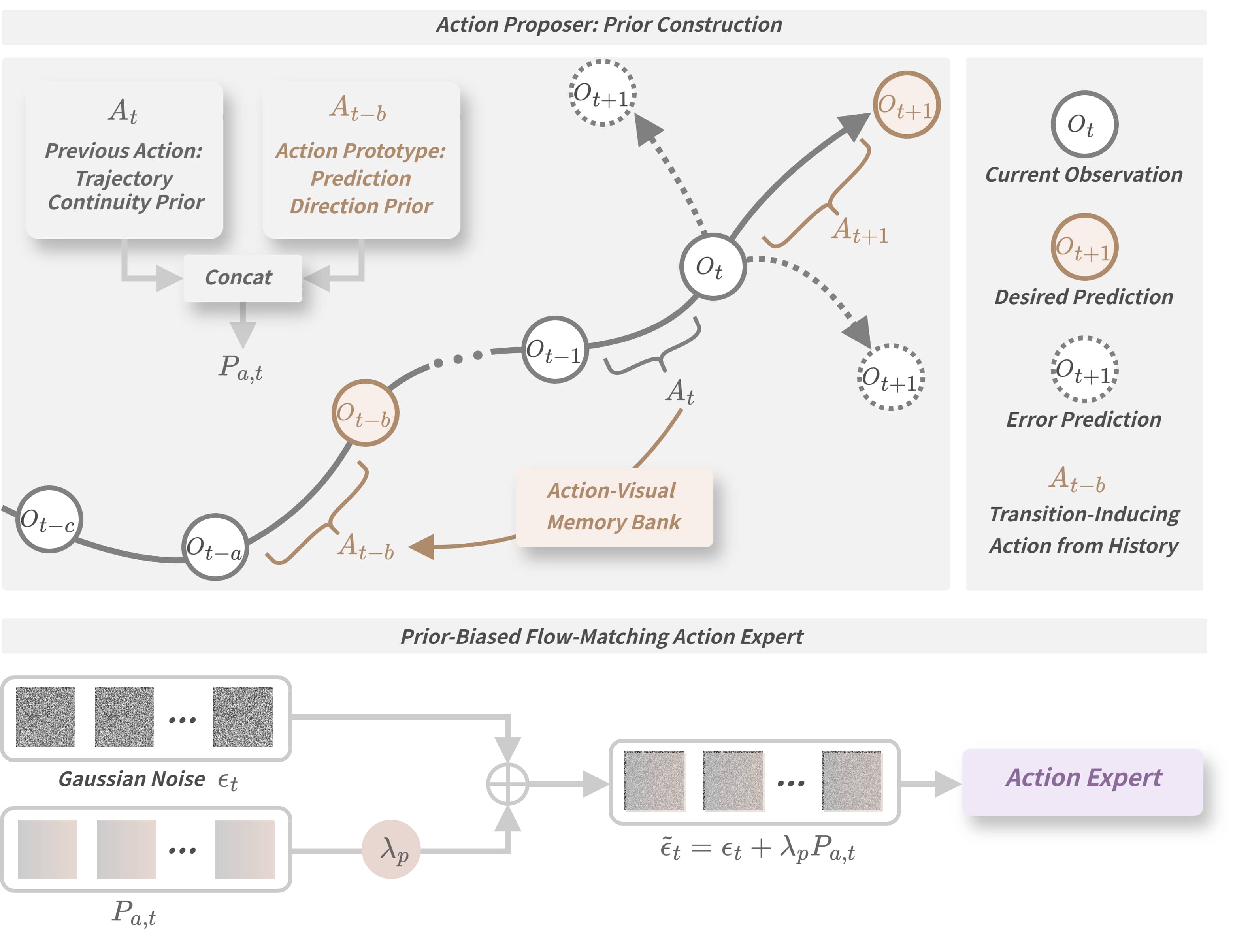} 
\vspace{-12pt}
    \caption{\textbf{Action-manifold prototype refinement.} Historical action evolution retrieves a visually grounded action prototype from the Action-Visual Memory Bank. The retrieved prototype is fused with recent action history to form an action-manifold prior, which translates the Gaussian source distribution toward a historically supported motion region. The Action Expert then refines the prior-biased samples through flow matching according to the current vision-language context and future-supervised latent transition representation.}
\label{fig:proposer}
\end{figure}

\subsubsection{Memory-Grounded Pixel Future Calibration}
\label{sec:pixel_future}
The transition tokens are not treated as an isolated auxiliary representation. Their decoded latent endpoint is used to construct an explicit transition-aware query for retrieving executable visual--action experience. For each view $\nu$, the future decoded from $\mathbf{Z}^{\mathrm{tr}}_t$ is projected into the visual memory key space:
\begin{equation}
\small
\widetilde{\mathbf{q}}^{\nu}_{v,t}
=
g_v^\nu
\left(
\hat{\mathbf{F}}^{\nu}_{t+\Delta}
\right),
\qquad
\mathbf{q}^{\nu}_{v,t}
=
\frac{
\widetilde{\mathbf{q}}^{\nu}_{v,t}
}{
\left\|
\widetilde{\mathbf{q}}^{\nu}_{v,t}
\right\|_2
}.
\label{eq:visual_memory_query}
\end{equation}
Because $\hat{\mathbf{F}}^{\nu}_{t+\Delta}$ is decoded from the current-conditioned transition tokens, $\mathbf{q}^{\nu}_{v,t}$ is an endpoint-oriented address of the anticipated transition rather than a generic descriptor of either the current or future observation. 
Following the coarse-to-fine temporal retrieval described in Sec.~\ref{sec:memory_indexing}, the query retrieves the action-aware transition value $\overline{\mathbf{u}}^{\nu}_{v,t}$ from the Visual-Action Memory Bank.

The pixel future predictor then uses the world latent tokens as queries and the retrieved visual values as decoder memory:
\begin{equation}
    \hat{\mathbf{I}}^{\nu}_{t+\Delta}
    =
    h_{\text{pix}}^\nu
    \left(
    \text{Up}(\mathbf{Z}^{\mathrm{tr}}_t),
    \Pi_v(\bar{\mathbf{u}}^{\nu}_{v,t})
    \right),
\end{equation}
where $\text{Up}(\cdot)$ upsamples the world tokens into pixel query tokens, and $\Pi_v$ projects the retrieved memory values into the pixel decoder space. The pixel future loss is selectively imposed only when the Trigger Gate is activated:
\begin{equation}
\small
    \mathcal{L}_{\text{pix}}
    =
    m_t
    \sum_{\nu\in\mathcal{U}}
    \frac{1}{|\Omega_\nu|}
    \left\|
    \hat{\mathbf{I}}^{\nu}_{t+\Delta}
    -
    \mathbf{I}^{\nu}_{t+\Delta}
    \right\|_1 .
\end{equation}
Through this mechanism, pixel prediction is not the representation consumed by the Action Expert. It serves as a selectively applied, memory-grounded supervision signal that calibrates $\mathbf{Z}^{\mathrm{tr}}_t$ around critical visual transitions.

\begin{figure*}[t]
\vspace{-5mm}
\centering
\includegraphics[width=1.0\textwidth]{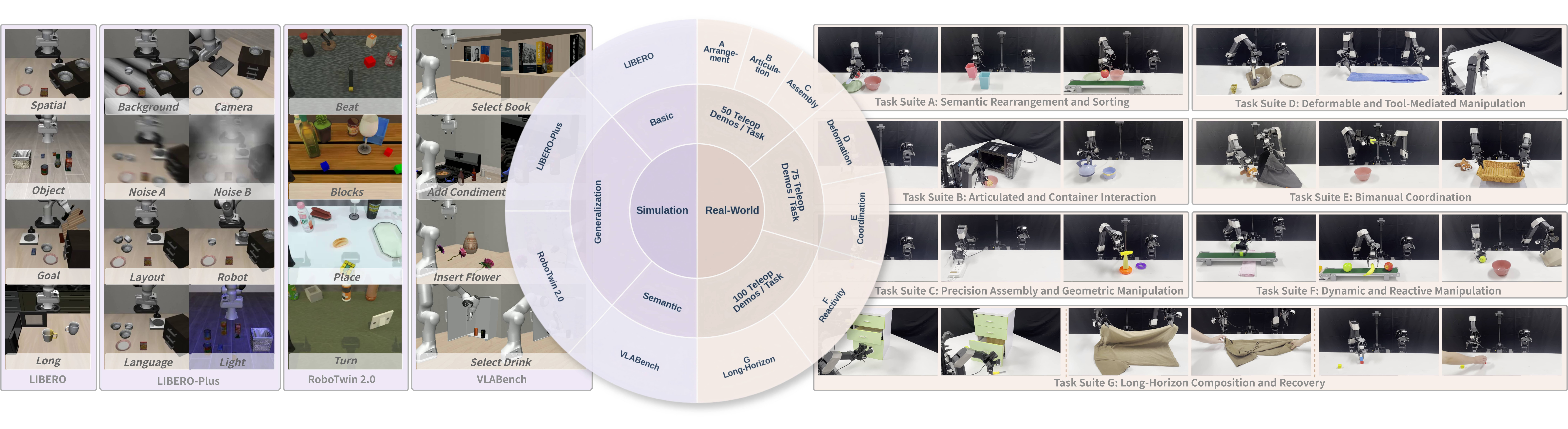} 
\vspace{-24pt}
    \caption{\textbf{Overview of the evaluation settings.} UniMPA is evaluated on complementary simulation benchmarks covering basic manipulation, controlled distribution shifts, randomized bimanual manipulation, and language-conditioned reasoning, together with real-world evaluation spanning seven task suites.}
\label{fig:setting}
\end{figure*}

\begin{figure}[t]
\centering
\includegraphics[width=0.5\textwidth]{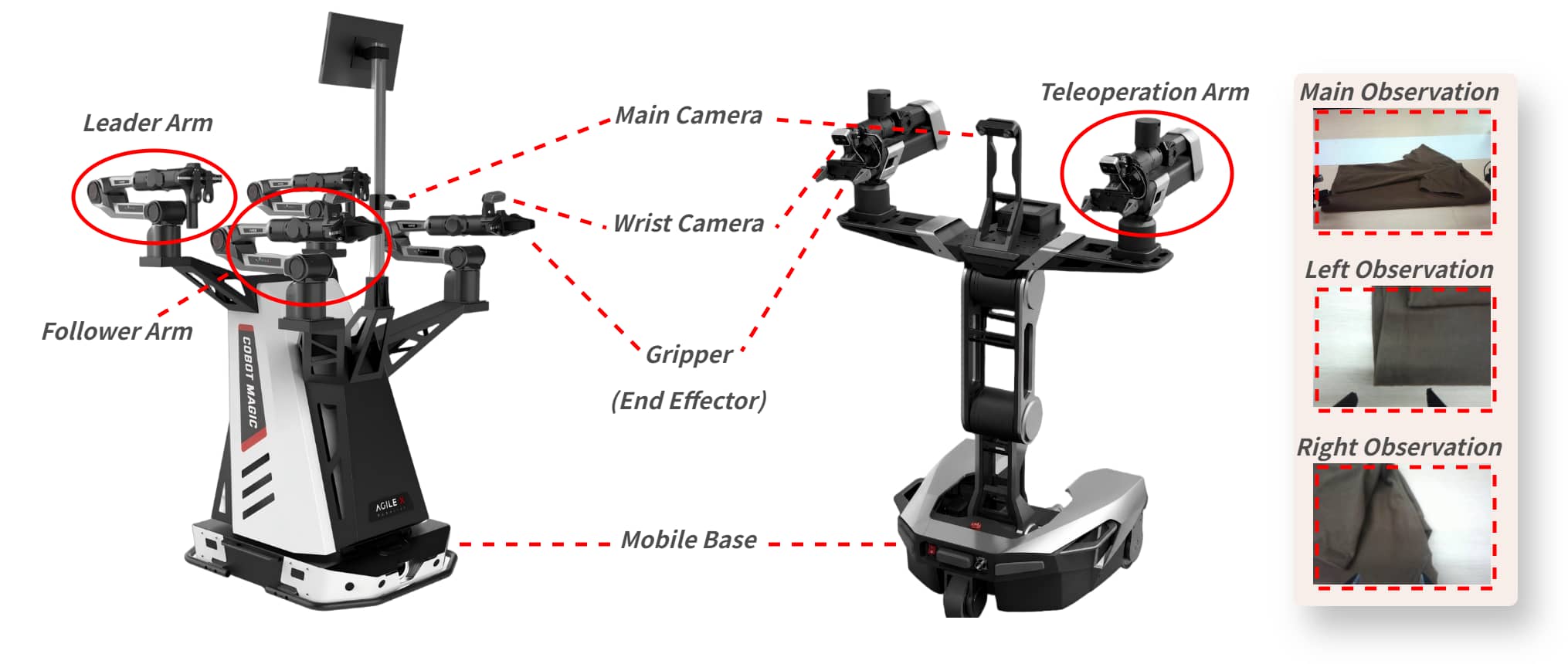} 
\vspace{-24pt}
    \caption{AgileX Cobot Magic (left) and  GALAXEA R1 Lite (right) platforms.}
\label{fig:robot}
\end{figure}

\subsection{Action-Manifold Prototype Refinement}
As shown in Fig.~\ref{fig:proposer}, unlike nearest-neighbor action copying, the retrieved historical experience is not treated as the final prediction. Instead, it defines an action prototype that biases the initial flow distribution toward a feasible region of the action space. The action expert subsequently extends, adjusts, or switches this historical motion tendency according to the current vision-language context and world-transition representation. Within {Prototype-Biased Flow}, the {Action Proposer} constructs an action-manifold prior from retrieved historical experience and recent action history, which shifts the initial Gaussian flow source before refinement by the action expert.

Given the recent action history $\mathbf{H}_t$, UniMPA first computes an action query using the frozen action-key encoder:
\begin{equation}
\small
\widetilde{\mathbf{q}}_{a,t}
=
g_a
\left(
\mathbf{H}_t
\right),
\qquad
\mathbf{q}_{a,t}
=
\frac{
\widetilde{\mathbf{q}}_{a,t}
}{
\left\|
\widetilde{\mathbf{q}}_{a,t}
\right\|_2
}.
\label{eq:action_memory_query}
\end{equation}
The query $\mathbf{q}_{a,t}$ represents the current execution tendency, including the recent translation, rotation, and gripper evolution.

Similar to visual retrieval, UniMPA applies the coarse-to-fine temporal retrieval described in Sec.~\ref{sec:memory_indexing} to the Action-Visual Memory Bank. Given the action query $\mathbf{q}_{a,t}$, this procedure retrieves the visually grounded action value $\overline{\mathbf{u}}_{a,t}$. Since each stored action value is conditioned on temporally aligned visual dynamics during memory pretraining, $\overline{\mathbf{u}}_{a,t}$ encodes both an executable motion pattern and the visual-transition semantics associated with its execution.

The retrieved value $\bar{\mathbf{u}}_{a,t}$ serves as an action-manifold prior rather than the final action, and is fused with a projection of the recent action history to produce prior tokens:
\begin{equation}
    \mathbf{P}_{a,t}
    =
    h_{\text{prior}}
    \left(
    [
    \phi_{\text{prev}}(\mathbf{H}_t),
    \bar{\mathbf{u}}_{a,t}
    ]
    \right),
    \quad
    \tilde{\boldsymbol{\epsilon}}_t
    =
    \boldsymbol{\epsilon}_t
    +
    \lambda_p\mathbf{P}_{a,t}.
\end{equation}
where $\lambda_p$ controls the strength of the retrieved action prior. Conditioned on $\mathbf{P}_{a,t}$, the resulting source distribution becomes
\begin{equation}
\small
p
\left(
\widetilde{\boldsymbol{\epsilon}}_t
\mid
\mathbf{P}_{a,t}
\right)
=
\mathcal{N}
\left(
\lambda_p\mathbf{P}_{a,t},
\mathbf{I}
\right).
\label{eq:prototype_source_distribution}
\end{equation}
Equation~\eqref{eq:prototype_source_distribution} formalizes this prototype bias as a translation of the zero-mean Gaussian source toward a historically supported action region while preserving its covariance and stochasticity.

Then, for a sampled flow timestep $\rho$, the noisy action input and target velocity are defined as
\begin{equation}
    \mathbf{x}_{\rho}
    =
    \rho\tilde{\boldsymbol{\epsilon}}_t
    +
    (1-\rho)\mathbf{A}_t,
    \quad
    \mathbf{u}_{\rho}
    =
    \tilde{\boldsymbol{\epsilon}}_t
    -
    \mathbf{A}_t .
\end{equation}
Through the three-system interaction described in Sec.~\ref{sec:framework}, the Action Expert receives the vision-language prefix $\mathbf{C}_t$, future-supervised transition representation $\mathbf{Z}^{\mathrm{tr}}_t$, and prototype-conditioned action context. Its predicted velocity field is
\begin{equation}
\small
\widehat{\mathbf{u}}_{\rho}
=
v_{\theta}
\left(
\mathbf{x}_{\rho},
\rho
\mid
\mathbf{C}_t,
\mathbf{Z}^{\mathrm{tr}}_t,
\mathbf{P}_{a,t}
\right)
=
\mathbf{W}_{o}
\mathbf{Z}^{a}_t.
\label{eq:action_velocity_prediction}
\end{equation}
The corresponding flow-matching objective is
\begin{equation}
\small
\mathcal{L}_{\mathrm{act}}
=
\mathbb{E}_{
\substack{
(\mathbf{X}_t,\mathbf{A}_t)\sim\mathcal{D},~
\rho\sim\mathcal{U}(0,1),~
\boldsymbol{\epsilon}_t\sim\mathcal{N}(\mathbf{0},\mathbf{I})
}
}
\left[
\frac{1}{KD}
\left\|
\hat{\mathbf{u}}_{\rho}
    -
    \mathbf{u}_{\rho}
\right\|_2^2
\right].
\label{eq:action_flow_loss}
\end{equation}

The Stage-2 objective combines action denoising, persistent latent prediction, and trigger-gated pixel prediction:
\begin{equation}
    \mathcal{L}_{\text{UniMPA}}
    =
    \mathcal{L}_{\text{act}}
    +
    \lambda_{\text{lat}}\mathcal{L}_{\text{lat}}
    +
    \lambda_{\text{pix}}\mathcal{L}_{\text{pix}} .
\end{equation}

During training, visual-memory retrieval supports pixel-level future supervision of the World Expert. At inference, we retain the World Expert and the Action–Visual Memory Bank, while disabling the latent and pixel prediction heads and their associated visual-retrieval path.
The World Expert processes the transition queries to produce $\mathbf{Z}^{\mathrm{tr}}_t$. Its layer-wise keys and values are cached and remain available to the Action Expert throughout iterative action generation. The Action-Visual Memory Bank remains active and produces $\mathbf{P}_{a,t}$. Starting from the prior-biased noise $\mathbf{x}_{1}=\boldsymbol{\epsilon}+\lambda_p\mathbf{P}_{a,t}$, the Action Expert iteratively denoises it into the final action chunk $\hat{\mathbf{A}}_t$ using both the current vision-language context and the latent transition. Deployment avoids explicit future-feature and future-image decoding while preserving future-supervised representations and retrieval-guided action refinement.


%% file: Sections/4-experiment.tex
\section{Experiments}
\label{sec:exp}

\subsection{Experimental Setup}
\label{sec:exp_setup}

\subsubsection{Simulation Benchmarks}
We evaluate UniMPA on four simulation benchmarks covering zero-shot robustness, cross-setting generalization, and semantic reasoning (Fig.~\ref{fig:setting} Left).

\input{Tables/libero-plus}

\input{Tables/libero}

\input{Tables/robotwinv2}

\input{Tables/libero-plus-long}

\textbf{LIBERO}~\cite{libero} measures manipulation performance across four suites (Spatial, Object, {{Goal}}, and {{Long}}) which respectively emphasize spatial relationships, object-centric transfer, goal-conditioned behavior, and long-horizon composition.

\textbf{LIBERO-Plus}~\cite{fei2025libero} evaluates zero-shot robustness under seven controlled perturbations: {{camera}}, {{robot}}, {{language}}, {{illumination}}, {{background}}, observation {{noise}}, and object {{layout}}. We report original performance, perturbation-averaged success rate, and the corresponding performance drop.

\textbf{RoboTwin~2.0}~\cite{chen2025robotwin} contains diverse object configurations and bimanual manipulation tasks. Models are trained only on the {{Clean (Easy) setting}} and evaluated on the {{Randomized (Hard) setting}} to assess out-of-distribution generalization.

\textbf{VLABench}~\cite{zhang2024vlabench} evaluates language-conditioned manipulation with diverse objects and implicit language instructions, emphasizing semantic understanding, commonsense transfer, spatial reasoning, and long-horizon manipulation.

\subsubsection{Real-World Task Suite}
We construct seven real-world manipulation suites according to their dominant physical interaction patterns (Fig.~\ref{fig:setting} Right): \textit{\textbf{(A)}} semantic rearrangement and sorting, \textit{\textbf{(B)}} articulated and container interaction, \textit{\textbf{(C)}} precision assembly and geometric manipulation, \textit{\textbf{(D)}} deformable and tool-mediated manipulation, \textit{\textbf{(E)}} bimanual coordination, \textit{\textbf{(F)}} dynamic and reactive manipulation, and \textit{\textbf{(G)}} long-horizon composition and recovery. The complete task definitions, language instructions, progressive checkpoints, and task-specific completion criteria are provided in \textbf{Appendix B}. 
We conduct 25 independent trials per task.
We evaluate both final task completion and intermediate execution progress using progressive checkpoints. \textit{\textbf{Task Success Rate (TSR)}} is the success rate at the final checkpoint, while \textit{\textbf{Cumulative Success Rate (CSR)}} summarizes progress across all checkpoints.

\subsubsection{Real-World Robot Platforms}
We conduct real-world experiments on two complementary bimanual platforms, as illustrated in Fig.~\ref{fig:robot}.
The {GALAXEA R1 Lite}~\cite{galaxea} is a 23-DoF mobile bimanual robot with a wheeled base, articulated torso, dual 6-DoF arms, and multi-view cameras for global and wrist-level observations.
The {AgileX Cobot Magic}~\cite{aloha}, built upon Mobile ALOHA, features a mobile base, dual 6-DoF arms, parallel grippers, and multi-view sensing. 

\subsection{Implementation Details}

\subsubsection{Model Details}
UniMPA is built on the $\pi_{0.5}$ backbone with a PaliGemma-2B vision-language backbone and a 311M-parameter Gemma action expert. The world expert follows the same 18-layer Transformer architecture as the action expert, with hidden dimension 1024, 8 heads, and FFN dimension 4096. We use 16 learnable world tokens. Latent prediction targets frozen V-JEPA~2 features with 16 tokens per view and dimension 1408. For pixel prediction, world tokens are expanded to 64 tokens and decoded by a 4-layer, 8-head Transformer with $32\times32$ patches into $256\times256$ images. We set the action horizon and history length to $K=M=10~\text{or}~50$, and use 10 Euler steps for flow-matching inference.

\subsubsection{Memory and Prediction Details}
The bidirectional memory uses a hidden dimension of 512. Visual keys project V-JEPA~2 features from 1408 to 512 dimensions, while visual and action values are modeled by 2-layer stateful Mamba encoders with state dimension 16. Vision-Mamba uses $32\times32$ patches and produces 64 temporal tokens per view; cross-modal fusion uses 8-head attention. Retrieval uses cosine similarity with temperature $\tau=0.1$ and a maximum causal history of 32 entries. The Trigger Gate activates pixel supervision when the latent-change score exceeds 20-50\% or a translation, rotation, or gripper-change indicator is active. We set $\lambda_{\mathrm{lat}}=\lambda_{\mathrm{pix}}=1.0$ for LIBERO and $0.01$ for RoboTwin~2.0, VLABench and real-world task suites.

\subsubsection{Training Details}
UniMPA is trained in two stages. Stage~1 pretrains the memory bank using AdamW with learning rate $1\times10^{-4}$. We use batch sizes of 16 for LIBERO and 8 for RoboTwin~2.0/VLABench/real-world tasks. Reconstruction and retrieval losses use unit weights, while the cross-modal alignment weight is 0.5. 
Stage~2 freezes the memory bank and fine-tunes the policy using AdamW with a peak learning rate of $5\times10^{-5}$ ($5\times10^{-6}$ for VLABench) after 5k or 10k warm-up steps. 
Training uses bfloat16 on 4-16 NVIDIA Pro6000 GPUs. We adopt benchmark-specific simulation training and task-suite-specific real-world training. 



\subsection{Evaluation on Simulation Benchmarks}
\label{sec:simulation_results}

\subsubsection{LIBERO}
Table~\ref{tab:libero_main} reports results on the four LIBERO suites. UniMPA achieves an average success rate of {98.6\%}, outperforming the strongest reported baseline, X-VLA, despite the near-saturated performance of this benchmark.
More importantly, the reported result is obtained using only 25\% of the training epochs required by $\pi_0$ and $\pi_{0.5}$. This training advantage supports the central motivation of UniMPA: predicted transitions and retrieved executable experience provide structured supervision and action priors that reduce the burden on direct observation-to-action fitting.

\subsubsection{LIBERO-Plus}
Table~\ref{tab:libero-plus} evaluates zero-shot generalization under seven perturbation dimensions. UniMPA achieves an average success rate of {85.3\%}, ranking first among all compared general, prediction-based, and memory-based methods. It exceeds the strongest competing average of 79.7\% by 5.6 points and improves over $\pi_{0.5}$ by 11.7 points. Meanwhile, its performance drop from the original setting is only 13.3 points, which is the smallest degradation among all compared methods.
The strong gain under visual noise shows that compact latent future representations provide robust transition queries. Improvements under camera, background, and layout shifts further indicate that memory retrieval relies on expected state transitions rather than frame-level appearance.

Across the four LIBERO task suites, UniMPA achieves 91.0\%, 89.7\%, 84.3\%, and 76.5\% average success under the perturbations for Spatial, Object, Goal, and Long, respectively. Long-horizon tasks remain the most challenging, but UniMPA preserves substantially stronger performance than direct observation-to-action baselines, as shown in Table~\ref{tab:libero-plus-long}. This supports the role of persistent latent future modeling and temporally constrained retrieval in maintaining phase-consistent execution under distribution shift.


\subsubsection{RoboTwin 2.0}
Table~\ref{tab:robotwin_random_subset} evaluates zero-shot generalization under the Randomized (Hard) setting, with all policies trained only on Clean demonstrations. Across the 11 tasks, UniMPA achieves an average success rate of {58.2\%}, outperforming HALO and $\pi_{0.5}$ by {25.7} and {18.5} percentage points, respectively, and ranking first on {8/11} tasks. Compared with $\pi_{0.5}$, UniMPA shows clear gains on challenging tasks such as Move PA ({46\%} vs.\ 14\%), Place OSc ({38\%} vs.\ 20\%), and Place PS ({37\%} vs.\ 7\%). These results show that action-grounded transition modeling generalizes effectively under scene and object randomization. Transition-conditioned memory retrieves executable experience, while Prototype-Biased Flow adapts the action prior to the current scene. Results on additional RoboTwin 2.0 tasks are provided in \textbf{Appendix C}.

\subsubsection{VLABench}
VLABench provides a complementary evaluation under semantically diverse manipulation tasks requiring stronger task understanding.
As shown in Table~\ref{tab:vlabench}, UniMPA achieves the highest average success rate of {44.0\%}, outperforming $\pi_{0.5}$ by {4.3} percentage points.
UniMPA achieves the best or tied-best performance on four of the six tasks, including Select Book, Select Drink, Select Chemistry Tube, and Insert Flower.
These results demonstrate that the proposed transition interface generalizes beyond visually specified goals: future-transition prediction provides a task-conditioned representation of the desired evolution, while retrieved vision-action experience supplies executable priors for realizing such transitions.

\input{Tables/vlabench}

\input{Tables/real-world}

\begin{figure*}[t]
\vspace{-4mm}
\centering
\includegraphics[width=1.0\textwidth]{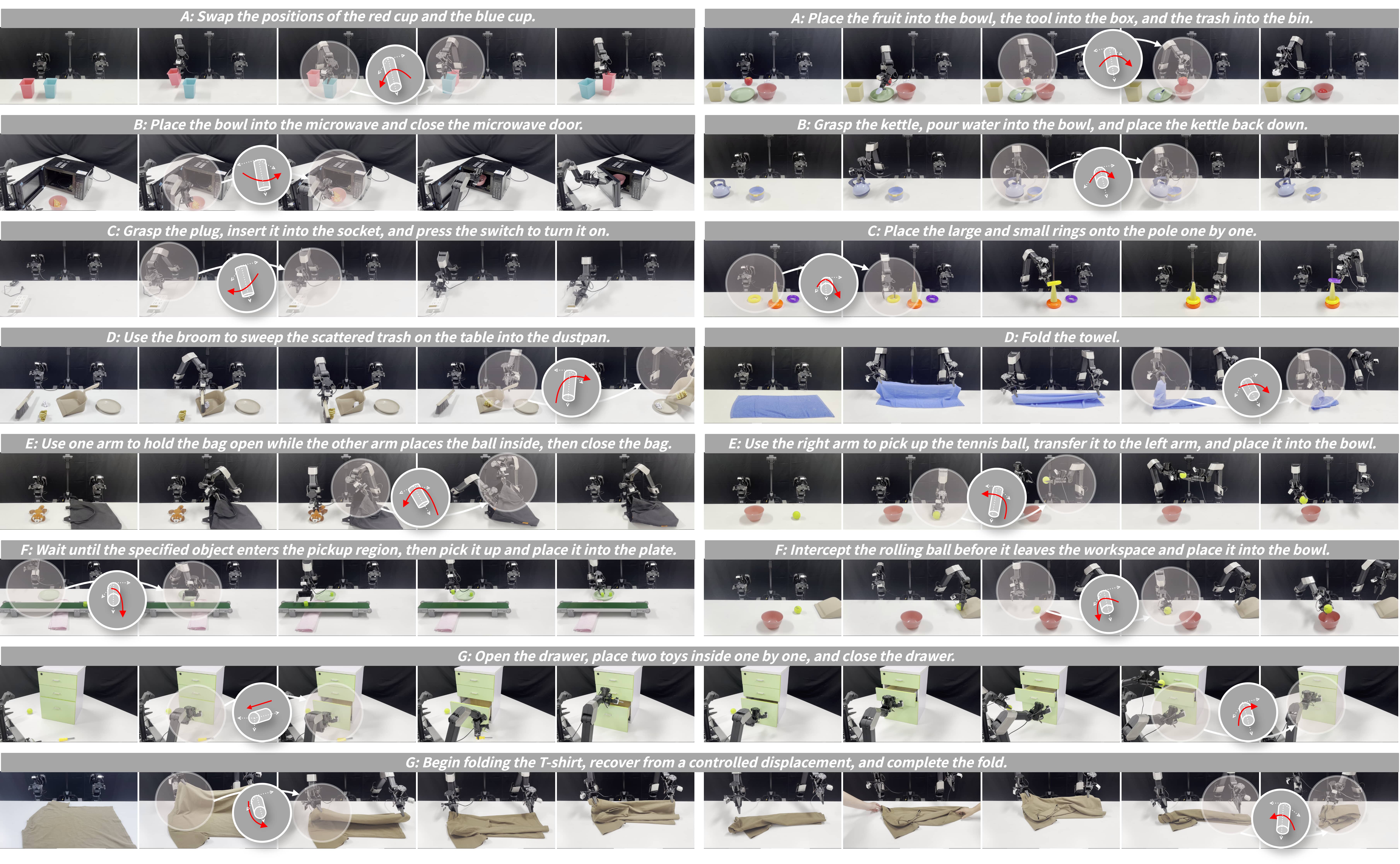} 
\vspace{-18pt}
    \caption{\textbf{Representative real-world manipulation trajectories.} Examples are shown across the seven task suites, including semantic rearrangement, articulated interaction, precision assembly, deformable and tool-mediated manipulation, bimanual coordination, dynamic interception, and long-horizon recovery. The highlighted regions indicate representative contact-sensitive or transition-critical interaction moments.}
\label{fig:realworld_vis}
\vspace{-8pt}
\end{figure*}

\subsection{Evaluation on Real-World Tasks}

Fig.~\ref{fig:realworld_vis} presents representative real-world manipulation trajectories across the seven task suites, illustrating diverse physical interaction patterns and highlighting contact-sensitive and transition-critical moments during execution.

\subsubsection{Evaluation on GALAXEA R1 Lite}
We conduct a large-scale evaluation on the
GALAXEA R1 Lite platform. As shown in
Table~\ref{tab:real_world}, the evaluation contains
21 tasks organized into seven manipulation suites,
with progressive checkpoints designed to characterize execution quality throughout different stages of long-horizon manipulation.
Across all tasks, UniMPA achieves {77.7\%} TSR and
{86.3\%} CSR, compared with 65.3\%/75.6\% for
$\pi_{0.5}$ and 45.7\%/58.1\% for OpenVLA-OFT. These results demonstrate that the proposed design remains effective across diverse manipulation regimes. Further analysis of the real-world tasks is provided in \textbf{Appendix B}.

\subsubsection{Evaluation on AgileX Cobot Magic}
Table~\ref{tab:aloha} reports results on seven representative tasks conducted on the AgileX Cobot Magic platform. UniMPA achieves an average {74.9\%} Task Success Rate (TSR) and {86.4\%} Cumulative Success Rate (CSR), outperforming $\pi_{0.5}$ by {12.6} and {11.5} percentage points, respectively. Together, the two platforms demonstrate the applicability of UniMPA across distinct bimanual embodiments under platform-specific training. The real-world ablation studies are conducted on these seven representative tasks on this platform, with detailed results provided in \textbf{Appendix D}.

\subsubsection{Dynamic and Long-Horizon Tasks}
UniMPA shows clear advantages on tasks requiring temporal coordination and adaptation. It achieves {82.7\%/86.7\%} TSR/CSR on bimanual coordination and {70.7\%/79.2\%} on dynamic and reactive tasks. On the long-horizon composition and recovery tasks, UniMPA reaches {72.0\%/83.3\%}, outperforming $\pi_{0.5}$ by {20.0/14.9} percentage points. These results support the complementary roles of persistent future modeling for tracking task progress and transition-aligned memory for retrieving phase-relevant executable experience, particularly in recovery scenarios.

\subsection{Ablation Study}

\subsubsection{Persistent-Selective Future Prediction}
Table~\ref{tab:ablation_prediction} validates the complementary roles of persistent latent and selective pixel prediction. Compared with the full model, latent-only prediction reduces the LIBERO average by 3.3 points and real-world TSR/CSR by 10.3/8.2 points, while pixel-only prediction drops by 1.7 and 6.3/5.2 points. Dense pixel prediction performs better than either single branch but still trails the full model by 1.3 points and 4.6/3.4 points. These results show that persistent latent prediction captures task progress, while selectively activated pixel prediction complements it with fine-grained transition cues at action-critical moments.

\input{Tables/real-world-aloha}

\input{Tables/ablation-perdition}

\input{Tables/ablation-gate}

\begin{figure}[t]
\vspace{-4mm}
\centering
\includegraphics[width=0.5\textwidth]{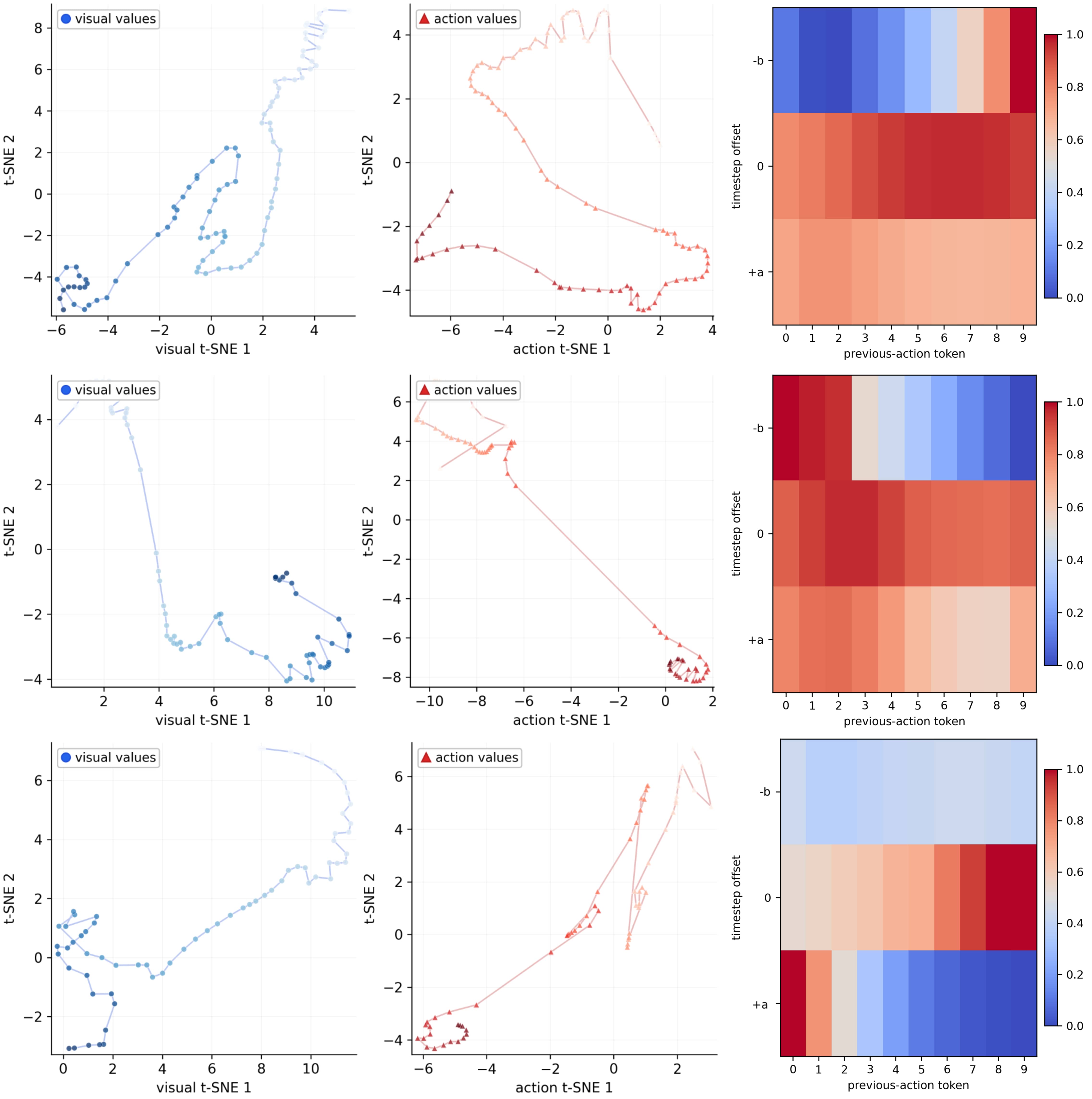} 
\caption{ \textbf{Visualization of temporal memory structure and cross-modal retrieval.} Each row shows one trajectory. The left and middle panels visualize the visual and action memory values of the same trajectory using t-SNE. Color intensity indicates temporal progression, showing that both modalities form smooth time-ordered trajectories after temporal modeling. The right panels show cross-modal retrieval similarity, where the temporally aligned pair at zero offset gives the strongest response, while randomly sampled negative offsets yield lower similarity. This demonstrates that UniMPA organizes visual and action memories around temporally aligned physical transitions. } 
\label{fig:memory_alignment_vis}
\vspace{-8pt}
\end{figure}

\begin{figure}[t]
\vspace{-4mm}
\centering
\includegraphics[width=0.5\textwidth]{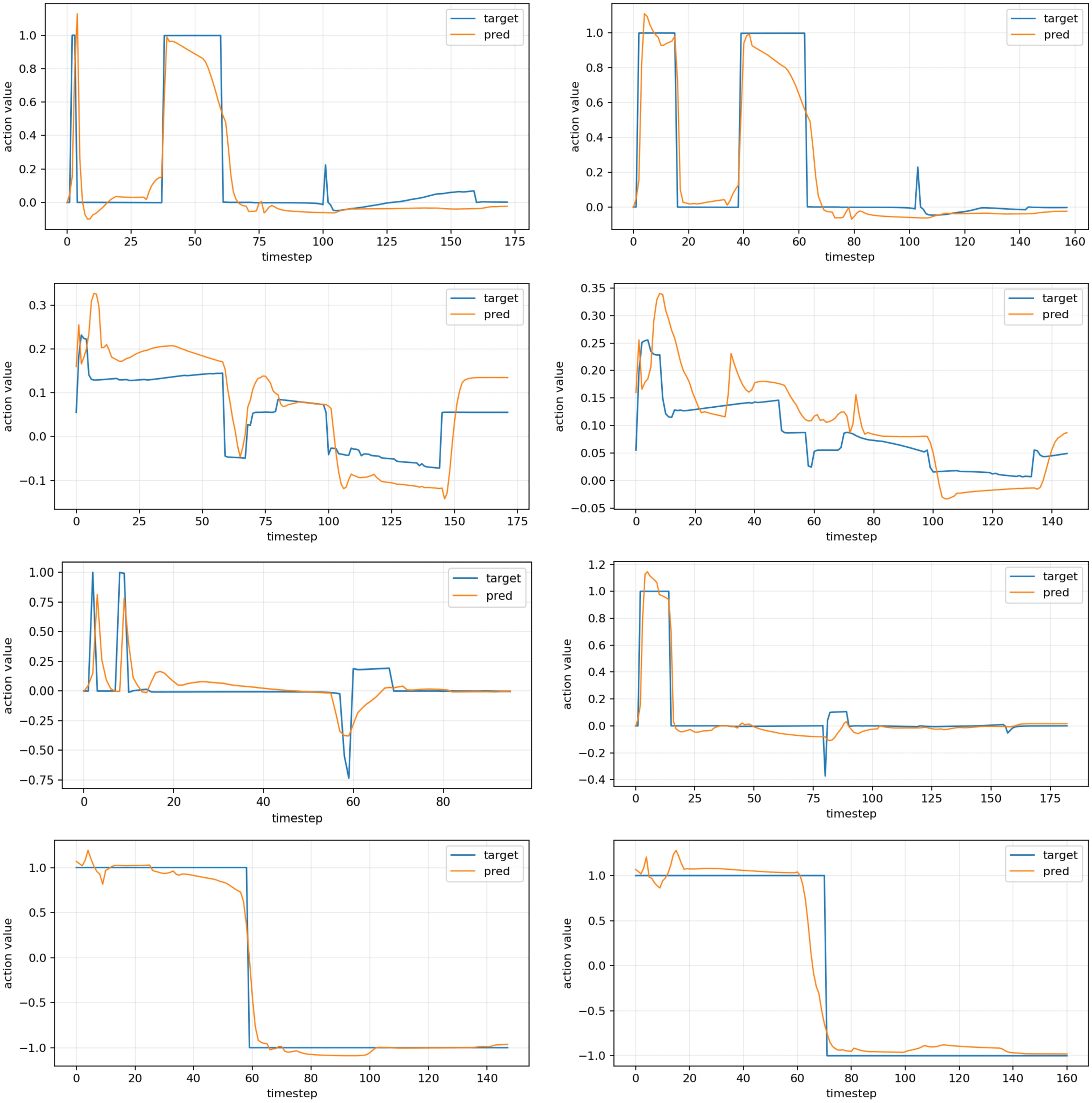} 
\caption{
\textbf{Visualization of executable action prototypes from the Action-Visual Memory Bank.}
Each row presents two dimensions of the 7-DoF action trajectory across different action types.
The prototype is obtained with a single forward pass and already captures the major motion trends of the target action, including direction, transition, and gripper-state changes.
Rather than being executed, it provides a coarse executable action manifold that the action expert further adapts to the current scene through iterative flow refinement.
}
\label{fig:action_prototype_vis}
\vspace{-8pt}
\end{figure}

\input{Tables/ablation-memory}
\input{Tables/ablation-coarse2fine}

\input{Tables/ablation-bank-pretraining}
\input{Tables/ablation-proposer}

\begin{figure}[t]
\vspace{-4mm}
\centering
\includegraphics[width=0.5\textwidth]{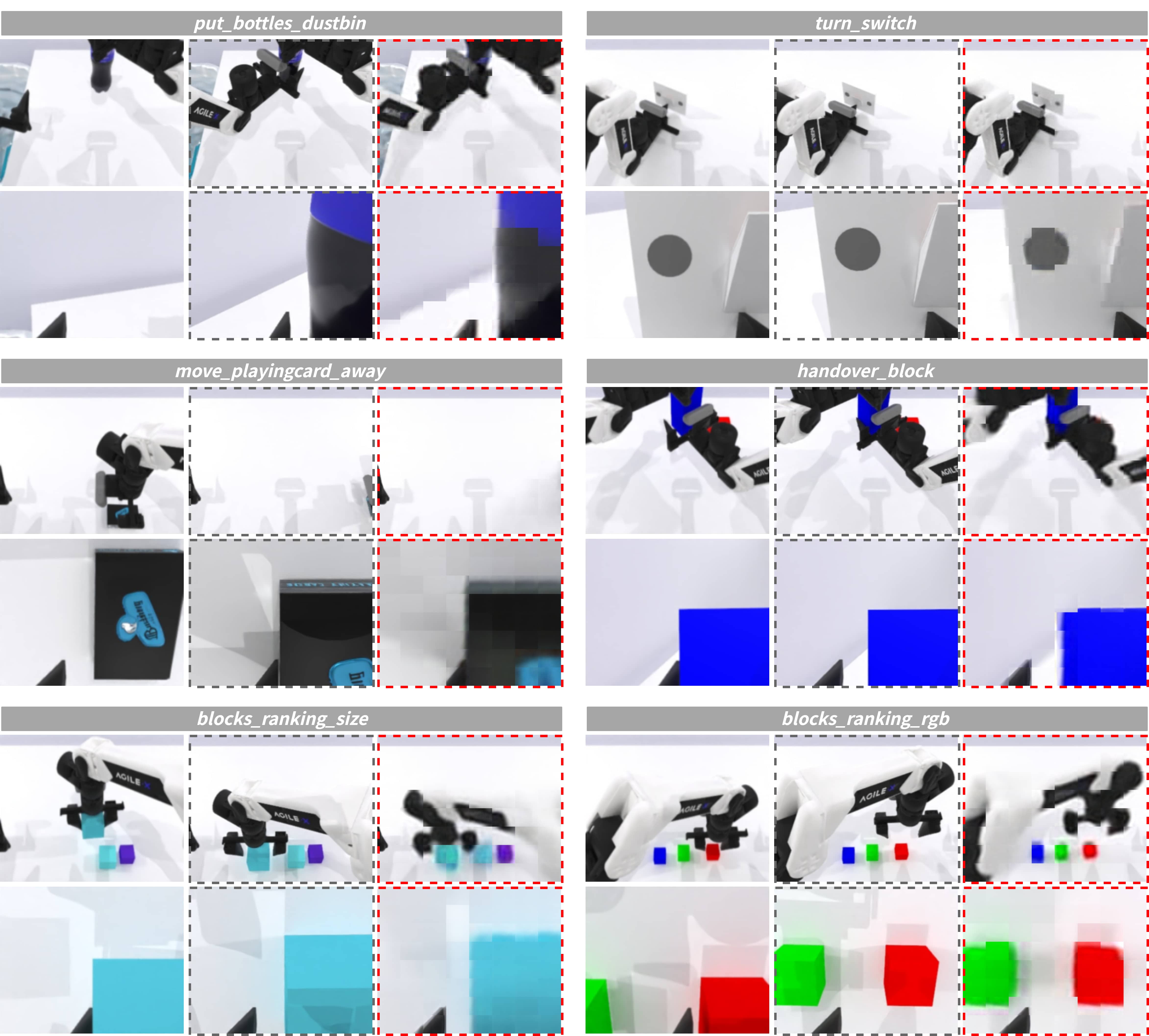} 
\caption{
\textbf{Visualization of pixel-level future prediction.}
For each example, the left panel is the current observation, the middle panel is the ground-truth future frame, and the right panel is the predicted future obtained in a single forward pass.
The predictions provide fine-grained action-relevant future cues.
}
\label{fig:pixel_future_vis}
\vspace{-5pt}
\end{figure}

Table~\ref{tab:ablation_trigger} further validates the transition-aware trigger. Compared with the combined Latent+Action Gate, random triggering reduces LIBERO performance by 2.6 points and real-world TSR/CSR by 9.8/7.3 points. Action-only and latent-only triggering also underperform the combined gate by 1.6 and 1.2 points on LIBERO, and by 6.3/5.1 and 5.2/3.4 points in real-world TSR/CSR. These results demonstrate that latent and action cues provide complementary signals for identifying transition-critical moments.

\subsubsection{Transition-Aligned Bidirectional Memory}
Table~\ref{tab:ablation_memory} verifies that UniMPA benefits from transition-aligned memory rather than generic historical context. Removing memory produces a {4.1}-point drop on LIBERO and larger {13.8/10.7}-point drops in real-world TSR/CSR, showing that future prediction alone does not reliably specify an executable realization. Removing the Visual-Action Bank reduces LIBERO and real-world TSR/CSR by {2.1} and {8.0/6.3} points, whereas removing the Action-Visual Bank causes corresponding drops of {1.5} and {6.9/5.0} points. The former confirms that predicted futures benefit from historical actions associated with compatible visual transitions, while the latter shows that action generation benefits from prototypes carrying visual-transition semantics. 

Temporal organization is also critical for phase-consistent retrieval. As shown in Table~\ref{tab:ablation_temporal_retrieval}, removing temporal modeling decreases the LIBERO average and real-world TSR/CSR by {2.7} and {9.2/7.0} points. Replacing coarse-to-fine retrieval with a fixed local window narrows the gap but still incurs drops of {1.9} and {6.3/4.8} points. The difference is particularly pronounced on LIBERO-Long, where removing temporal modeling and using fixed local retrieval decrease success by {6.0} and {3.8} points. By first identifying a compatible trajectory and then retrieving within its context, UniMPA avoids mixing visually similar but execution-incompatible phases across demonstrations.

\subsubsection{Future-Oriented Memory Pretraining}
Table~\ref{tab:ablation_memory_pretraining} shows that the effectiveness of memory depends on how its representation space is learned. Replacing future-oriented training with current-state reconstruction causes a {2.5} points degradation on LIBERO and {12.0/9.7} points drops in real-world TSR/CSR, confirming that observation reconstruction primarily preserves appearance rather than transition semantics. 
Removing visual future reconstruction decreases performance by {2.2} and {10.3/7.9} points, while removing action future reconstruction causes corresponding drops of {1.9} and {8.0/5.8} points. This shows that both sides of a physical transition must be retained. Visual values encode expected state evolution, whereas action values preserve the executable motion that realizes it. Removing the cross-modal pairing objective further reduces performance by {1.5} and {5.8/4.1} points, demonstrating that explicit vision-action correspondence is important for shaping the two memory directions around compatible physical transitions.

\subsubsection{Action-Manifold Prototype Refinement}
As shown in Table~\ref{tab:ablation_action_prior}, directly copying the nearest retrieved action produces the largest degradation, falling {7.3} points on LIBERO and {22.9/18.9} points in real-world TSR/CSR relative to Prototype-Biased Flow. This confirms that historical actions cannot be directly reused under the current scene geometry. Removing the Action Proposer reduces performance by {2.4} and {10.3/7.8} points, while using the retrieved prototype only as a conditioning feature still incurs drops of {1.6} and {5.8/4.8} points. These results show that retrieved experience is most effective when it directly biases the flow source toward an executable action region and is subsequently refined according to the current vision-language and world-transition context.

\subsection{Qualitative Analysis}

\subsubsection{Temporal and Cross-Modal Memory Alignment}

Fig.~\ref{fig:memory_alignment_vis} visualizes the temporal structure and cross-modal correspondence learned by the memory bank.
Each row corresponds to one trajectory.
The left two panels project the visual and action memory values into two-dimensional t-SNE spaces, with darker colors indicating later timesteps.
In both modalities, the memory values form smooth ordered trajectories rather than scattered points, showing that temporal modeling preserves episode-level evolution. The right panels show cross-modal retrieval similarity.
For each example, the aligned visual-action pair at zero offset consistently produces the highest similarity, while randomly sampled negative offsets are weaker.
This indicates that UniMPA does not match memories by appearance or action similarity alone, but aligns visual state evolution with the action dynamics that realize the same physical transition.

\subsubsection{Executable Action Prototype Visualization}

Fig.~\ref{fig:action_prototype_vis} visualizes representative action prototypes produced by the Action-Visual Memory Bank.
Across translation, rotation, and gripper dimensions, the one-pass prototypes closely follow the coarse evolution of the target trajectories, preserving major motion directions, stage transitions, and action-state switches despite local deviations in magnitude.
This behavior is consistent with the intended role of the retrieved action memory.
The prototype is not required to reproduce the target action precisely, but to identify a historically executable motion region before iterative action generation.
Compared with generating actions from an unstructured Gaussian source through multiple denoising steps, the retrieved prototype provides an informed starting tendency with only a single forward pass.

\begin{figure}[t]
\vspace{-4mm}
\centering
\includegraphics[width=0.5\textwidth]{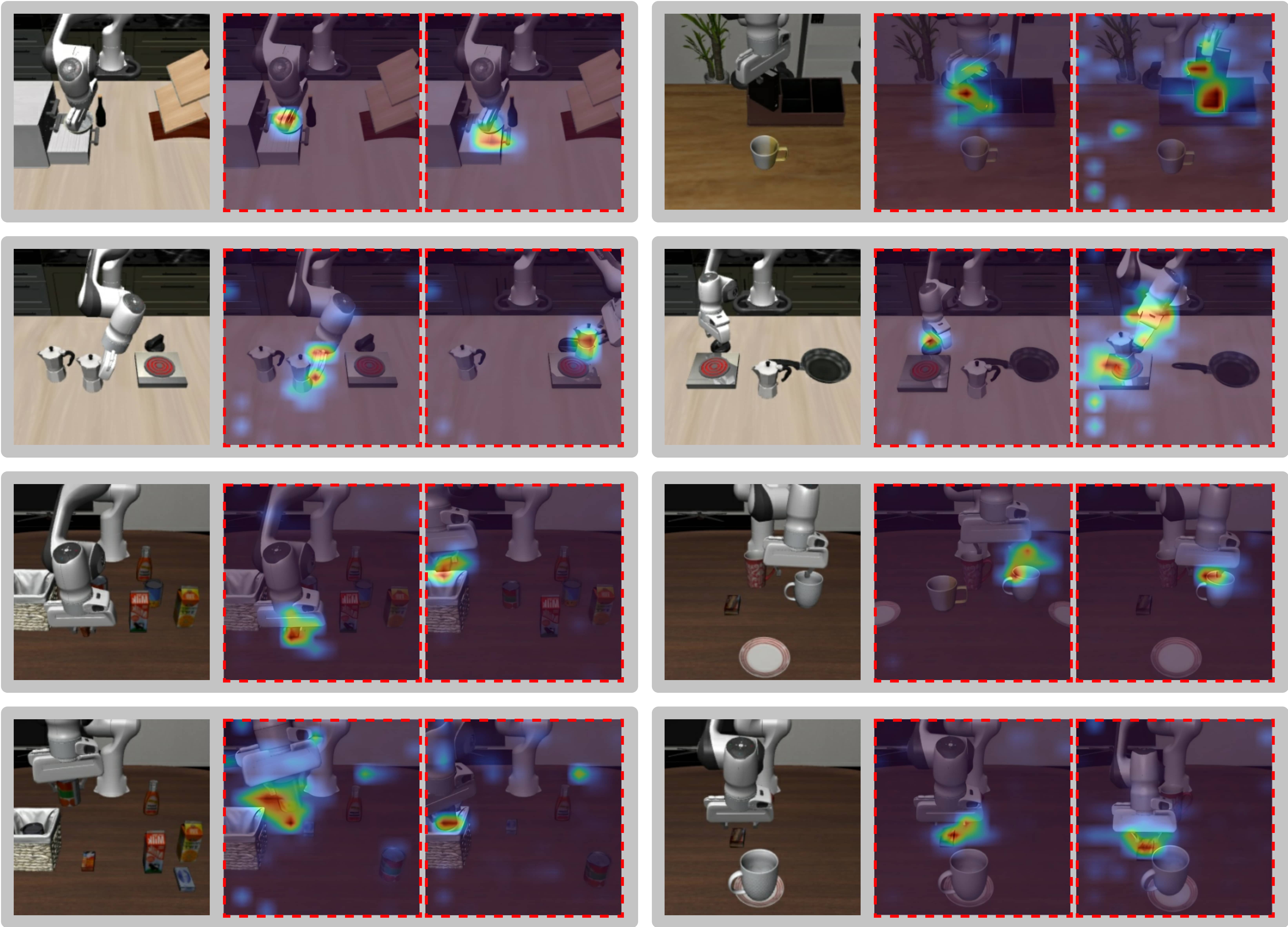} 
\caption{
\textbf{Visualization of world-transition attention maps.}
For each example, the left panel is the current observation.
The middle and right panels visualize attention maps between the intermediate feature produced by the world expert and all patch tokens from two observations sampled at different timesteps along the trajectory, where the right panel corresponds to a later timestep than the middle panel.
The attention consistently concentrates on manipulation-relevant regions such as the robot end-effector, target objects, and contact areas, and evolves with the task progression.
}
\label{fig:attention_map_vis}
\vspace{-8pt}
\end{figure}

\subsubsection{Pixel-Level Future Prediction Visualization}

As shown in Fig.~\ref{fig:pixel_future_vis}, given the current observation, the model generates the future frame in a single forward pass.
Across diverse tasks, the predicted results reflect key future changes, including robot motion, object displacement, and contact-related scene evolution.
These examples show that the pixel-level future prediction provides informative visual cues at critical interaction moments and serves as an effective complement.

\subsubsection{World-Transition Attention Visualization}

The attention behavior of the world expert is illustrated in Fig.~\ref{fig:attention_map_vis}.
Each row contains two examples, and each example includes the current observation together with two attention maps computed against observations from different timesteps of the same trajectory.
The maps show that the intermediate feature output by the world expert mainly attends to regions closely related to manipulation, including the gripper, the manipulated object, and the surrounding contact area.
More importantly, the attended regions evolve across timesteps in a temporally coherent manner.
This suggests that the world expert does not merely encode static scene appearance, but captures action-relevant transition cues associated with task progression.

%% file: Tables/libero-plus.tex
\begin{table*}[t]
\vspace{-4mm}
\caption{\textbf{Results on LIBERO-Plus benchmark~\cite{fei2025libero}.} 
We report generalization performance across seven perturbation dimensions. UniMPA is evaluated in a {zero-shot} setting with 10,030 total rollouts. }
\label{tab:libero-plus}
\vspace{-5pt}

\centering
\setlength{\tabcolsep}{4.5pt}
\begin{tabular}{lcccccccc|cc|cc}
\toprule
{Method}
& {Origin}
& {Camera}
& {Robot}
& {Language}
& {Light}
& {Background}
& {Noise}
& {Layout}
& {Average} $\uparrow$
& {Rank} $\downarrow$
& {$\Delta$} $\downarrow$
& {Rank} $\downarrow$ \\
\midrule

\rowcolor[HTML]{EFEFEF}
\multicolumn{13}{c}{\textit{\scriptsize General VLA Models}} \\
\midrule

OpenVLA~\textit{[CoRL'24]}~\cite{openvla}
& 75.6 & 0.8 & 3.5 & 23.0 & 8.1 & 34.8 & 15.2 & 28.5
& 15.6 & 22 & 60.0 & 21 \\

OpenVLA-OFT~\textit{[RSS'25]}~\cite{oft}
& 97.1 & 56.4 & 31.9 & 79.5 & 88.7 & 93.3 & 75.8 & 74.2
& 69.6 & 14 & 27.5 & 14 \\


UniVLA~\textit{[RSS'25]}~\cite{univla}
& 95.2 & 1.8 & 46.2 & 69.6 & 69.0 & 81.0 & 21.2 & 31.9
& 42.9 & 20 & 52.3 & 19 \\

$\pi_0$~\textit{[RSS'25]}~\cite{pi0}
& 94.2 & 13.8 & 6.0 & 58.8 & 85.0 & 81.4 & 79.0 & 68.9
& 53.6 & 18 & 40.6 & 17 \\

$\pi_0$-Fast~\textit{[RSS'25]}~\cite{fast}
& 85.5 & 65.1 & 21.6 & 61.0 & 73.2 & 73.2 & 74.4 & 68.8
& 61.6 & 15 & 23.9 & 9 \\

$\pi_{0.5}$~\textit{[CoRL'25]}~\cite{pi05}
& 96.9 & 59.7 & 65.5 & 75.3 & 87.0 & 82.4 & 72.1 & 80.3
& 73.6 & 8 & 23.3 & 8 \\

VLA-Adapter~\textit{[AAAI'26]}~\cite{wang2025vlaadapter}
& 97.3 & 36.2 & 37.9 & 74.6 & 70.6 & 76.1 & 58.0 & 69.7
& 59.1 & 16 & 38.2 & 15 \\

HoloBrain~\textit{[arXiv'26]}~\cite{lin2026holobrain}
& 97.4 & 66.3 & 49.0 & 65.9 & 94.9 & 93.3 & 73.1 & 78.2
& 72.6 & 10 & 24.8 & 10 \\

GuidedVLA~\textit{[RSS'26]}~\cite{jia2026guidedvla}
& -
& {73.7}
& 51.4
& 62.6
& 94.6
& 89.0
& 85.2
& 79.9
& 75.4
& 7
& -
& - \\

X-VLA~\textit{[ICLR'26]}~\cite{zheng2025x}
& 98.1
& 23.4
& {89.7}
& 75.7
& 88.2
& 96.0
& 62.7
& 71.8
& 71.4
& 11
& 26.7
& 13 \\

\midrule
\rowcolor[HTML]{EFEFEF}
\multicolumn{13}{c}{\textit{\scriptsize Prediction-Based VLA/WAM Methods}} \\
\midrule

WorldVLA~\textit{[arXiv'25]}~\cite{cen2025worldvla}
& 79.1 & 0.1 & 27.9 & 41.6 & 43.7 & 17.1 & 10.9 & 38.0
& 25.0 & 21 & 54.1 & 20 \\

DreamVLA~\textit{[NeurIPS'25]}~\cite{zhang2026dreamvla}
& 92.6 & 65.0 & 40.8 & 63.5 & 85.7 & 82.6 & 84.9 & 74.0
& 69.9 & 13 & 22.7 & 7 \\

VLA-JEPA~\textit{[arXiv'26]}~\cite{sun2026vlajepa}
& 97.2
& 63.3
& 67.1
& 85.4
& 95.6
& 93.6
& 66.3
& {85.1}
& 79.5
& 4
& {17.7}
& 2 \\

FutureVLA~\textit{[arXiv'26]}~\cite{xu2026futurevla}
& 98.3
& 59.7
& 66.0
& 88.2
& 97.4
& {97.8}
& 77.3
& 82.6
& {79.7}
& 2
& 18.6
& 4 \\

LaMP~\textit{[arXiv'26]}~\cite{wang2026lamp}
& 98.3
& 64.5
& 69.6
& 88.2
& 95.3
& {97.4}
& 76.9
& 73.8
& 79.3
& 6
& 19.0
& 6 \\

Cosmos Policy~\textit{[arXiv'26]}~\cite{kim2026cosmos}
& {98.5}
& 69.6
& 51.0
& {89.6}
& {97.7}
& 85.7
& {87.3}
& 83.7
& {79.7}
& 2
& 18.8
& 5 \\

Fast-WAM~\textit{[arXiv'26]}~\cite{yuan2026fast}
& {98.5}
& 34.0
& 55.1
& 88.9
& 90.0
& 44.9
& 33.3
& 73.4
& 59.0
& 17
& 39.5
& 16 \\

\midrule
\rowcolor[HTML]{EFEFEF}
\multicolumn{13}{c}{\textit{\scriptsize Memory-Based VLA Models}} \\
\midrule

MemoryVLA~\textit{[ICLR'26]}~\cite{shi2025memoryvla}
& 96.5 & 42.7 & 44.9 & 84.4 & 92.8 & 95.0 & 62.1 & 84.7
& 70.2 & 12 & 26.3 & 12 \\

MemoryVLA++~\textit{[arXiv'26]}~\cite{shi2026memoryvla++}
& 98.4 & 36.4 & 68.9 & 88.7 & 93.8 & 90.6 & 63.5 & 83.8
& 73.1 & 9 & 25.3 & 11 \\

VF-VLA~\textit{[arXiv'26]}~\cite{park2026test}
& 96.7 & 31.8 & 49.2 & 48.2 & 44.9 & 63.0 & 55.3 & 72.4
& 52.1 & 19 & 44.6 & 18 \\

Libra-VLA~\textit{[ACL'26]}~\cite{wei2026libra}
& 97.2
& 68.9
& 48.8
& {92.7}
& {97.9}
& 93.4
& 86.3
& 77.5
& 79.5
& 4
& {17.7}
& 2 \\

\midrule
\rowcolor[HTML]{EFE9F7}
UniMPA
& {98.6}
& {75.7}
& {73.0}
& 84.6
& 96.8
& 94.5
& {90.9}
& {87.3}
& \textbf{85.3}
& {1}
& \textbf{13.3}
& {1} \\

\rowcolor[HTML]{EFE9F7}
\textit{~~~~~~Spatial}
& 99.6
& 82.7
& 79.7
& 91.5
& 96.2
& 96.9
& 95.2
& 96.9
& 91.0
& --
& 8.6
& -- \\

\rowcolor[HTML]{EFE9F7}
\textit{~~~~~~Object}
& 99.8
& 86.6
& 71.6
& 88.4
& 98.3
& 98.4
& 96.9
& 92.3
& 89.7
& --
& 10.1
& -- \\

\rowcolor[HTML]{EFE9F7}
\textit{~~~~~~Goal}
& 98.8
& 78.9
& 75.3
& 80.7
& 99.3
& 96.1
& 94.2
& 75.3
& 84.3
& --
& 14.5
& -- \\

\rowcolor[HTML]{EFE9F7}
\textit{~~~~~~Long}
& 96.0
& 56.1
& 66.2
& 78.3
& 93.1
& 87.5
& 79.3
& 85.6
& 76.5
& --
& 19.5
& -- \\

\bottomrule
\end{tabular}
\vspace{-3pt}

\end{table*}

%% file: Tables/libero.tex
\begin{table}[t]
\centering
\caption{\textbf{Results on the LIBERO benchmark~\cite{libero}.} We report success rates on four task suites with 2,000 total rollouts. UniMPA achieves state-of-the-art performance using only \textbf{25\%} of the training epochs required by $\pi_0$ and $\pi_{0.5}$.}
\label{tab:libero_main}
\vspace{-5pt}
\resizebox{\columnwidth}{!}{%
    \begin{tabular}{l|cccc|c}
        \toprule
        {Method} & {Spatial} & {Object} & {Goal} & {Long} & {Avg.} \\
        \midrule

        Octo~\textit{[RSS'24]}~\cite{octo} & 78.9 & 85.7 & 84.6 & 51.1 & 75.1 \\

        OpenVLA~\textit{[CoRL'24]}~\cite{openvla} & 84.7 & 88.4 & 79.2 & 53.7 & 76.5 \\
        OpenVLA-OFT~\textit{[RSS'25]}~\cite{oft} & 97.6 & 98.4 & 97.9 & 94.5 & 97.1 \\



        UniVLA~\textit{[RSS'25]}~\cite{univla} & 95.4 & 98.8 & 93.6 & 94.0 & 95.4 \\

        SpatialVLA~\textit{[RSS'25]}~\cite{spatialvla} & 88.2 & 89.9 & 78.6 & 55.5 & 78.1 \\

        CoT-VLA~\textit{[CVPR'25]}~\cite{cotvla} & 87.5 & 91.6 & 87.6 & 69.0 & 83.9 \\

        TraceVLA~\textit{[ICLR'25]}~\cite{zheng2024tracevla} & 84.6 & 85.2 & 75.1 & 54.1 & 74.8 \\

        $\pi_0$~\textit{[RSS'25]}~\cite{pi0} & 96.8 & 98.8 & 95.8 & 85.2 & 94.2 \\
        $\pi_0$-Fast~\textit{[RSS'25]}~\cite{fast}  & 96.4 & 96.8 & 88.6 & 60.2 & 85.5 \\
        $\pi_{0.5}$~\textit{[CoRL'25]}~\cite{pi05} & {98.8} & 98.2 & 98.0 & 92.4 & 96.9 \\
        $\pi_{0.5}$-KI~\textit{[arXiv'25]}~\cite{driess2026knowledge} & 98.0 & 97.8 & 95.6 & 85.8 & 94.3 \\

        DreamVLA~\textit{[NeurIPS'25]}~\cite{zhang2026dreamvla} & 97.5 & 94.0 & 89.5 & 89.5 & 92.6 \\
        
        Fast-WAM~\textit{[arXiv'26]}~\cite{yuan2026fast}  & 98.2 & \textbf{100.0} & {97.0} & {95.2} & {97.6} \\

        VLA-Adapter~\textit{[AAAI'26]}~\cite{wang2025vlaadapter} & 97.8 & 99.2 & 97.2 & {95.0} & 97.3 \\

        X-VLA~\textit{[ICLR'26]}~\cite{zheng2025x} & 98.2 & 98.6 & 97.8 & \textbf{97.6} & 98.1 \\

        Motus~\textit{[arXiv'26]}~\cite{bi2026motus}  & 96.8 & 99.8 & 96.6 & \textbf{97.6} & 97.7 \\

        GeoVLA~\textit{[IROS'26]}~\cite{sun2025geovla}  & 98.4 & 99.0 & 96.6 & {96.6} & 97.7 \\

        VLA-JEPA~\textit{[arXiv'26]}~\cite{sun2026vlajepa}  & 96.2 & 99.6 & 97.2 & {95.8} & 97.2 \\

        \midrule
        \rowcolor[HTML]{EFE9F7} 
        \textbf{UniMPA} & \textbf{99.6} & {99.8} & \textbf{98.8} & {96.0} & \textbf{98.6} \\
        \bottomrule
    \end{tabular}%
}
\vspace{-8pt}
\end{table}

%% file: Tables/robotwinv2.tex
\begin{table*}[t]
\vspace{-4mm}
\caption{\textbf{Results on RoboTwin 2.0 benchmark~\cite{chen2025robotwin}}. 
All policies are trained on the Clean setting and evaluated without task-specific adaptation under the \textbf{Randomized (hard) zero-shot} setting, with 100 rollouts per task.}
\vspace{-5pt}
\centering
\footnotesize
\setlength{\tabcolsep}{4pt}
\label{tab:robotwin_random_subset}

\resizebox{\textwidth}{!}{
\begin{tabular}{l|ccccccccccc}
\toprule
Method 
& Adjust B 
& Dump BB 
& Grab R 
& Move PA 
& Open L 
& Place BF 
& Place EC 
& Place OSc 
& Place PS 
& Shake BH 
& Turn S \\
\midrule

ACT~\textit{[RSS'23]}~\cite{zhao2023act} 
& 23\% & 1\% & 25\% & 0\% & 0\% 
& 0\% & 0\% & 0\% & 0\% & 4\% & 2\% \\

DP~\textit{[RSS'23]}~\cite{diff} 
& 0\% & 0\% & 0\% & 0\% & 0\% 
& 0\% & 0\% & 0\% & 0\% & 18\% & 1\% \\

DP3~\textit{[RSS'24]}~\cite{ze20243d} 
& 3\% & 53\% & 2\% & 3\% & 7\% 
& 18\% & 1\% & 0\% & 2\% & 25\% & 8\% \\

$\pi_{0}$~\textit{[RSS'25]}~\cite{pi0} 
& 56\% & 24\% & 80\% & 22\% & 46\% 
& 4\% & 11\% & 0\% & 7\% & 51\% & 23\% \\

$\pi_{0.5}$~\textit{[CoRL'25]}~\cite{pi05}
& 54\% & 30\% & 63\% & {14\%} & 37\% 
& \textbf{45\%} & 53\% & 20\% & 7\% & 94\% & 20\% \\

RDT~\textit{[ICLR'25]}~\cite{liu2024rdt} 
& 75\% & 32\% & 43\% & 11\% & 32\% 
& 27\% & 7\% & 0\% & 6\% & 51\% & 15\% \\

UP-VLA~\textit{[ICML'25]}~\cite{zhang2025up} 
& 17\% & 35\% & 28\% & 13\% & 21\% 
& 26\% & 27\% & 4\% & 0\% & 68\% & 26\% \\

KAM-WM~\textit{[arXiv'26]}~\cite{shao2026kam}
& - & 47\% & 80\% & {21\%} & \textbf{69\%} 
&{ 40\%} & 6\% & 1\% & 15\% & 55\% & 10\% \\

TwinVLA~\textit{[ICLR'26]}~\cite{im2025twinvla}
& 35\% & 34\% & 22\% & 35\% & 17\% 
& 13\% & 1\% & 0\% & 2\% & 55\% & 15\% \\

BagelVLA~\textit{[RSS'26]}~\cite{hu2026bagelvla}
& 14\% & 51\% & 41\% & 30\% & 37\% 
& 11\% & 34\% & 0\% & 2\% & 73\% & 30\% \\

HALO~\textit{[ICML'26]}~\cite{shou2026halo}
& 9\% & 28\% & 57\% & \textbf{53\%} & 37\% 
& 37\% & 28\% & 5\% & 10\% & 66\% & 27\% \\

\midrule

\rowcolor[HTML]{EFE9F7}
\textbf{UniMPA} 
& \textbf{80\%} 
& \textbf{62\%} 
& \textbf{84\%} 
& 46\% 
& \textbf{67\%} 
& {40\%} 
& \textbf{57\%} 
& \textbf{38\%} 
& \textbf{37\%} 
& \textbf{95\%} 
& \textbf{34\%} \\

\bottomrule
\end{tabular}
}

\vspace{-3pt}
\end{table*}

%% file: Tables/libero-plus-long.tex
\begin{table}[t]
\vspace{-4mm}

\caption{\textbf{Results on LIBERO-Plus~\cite{fei2025libero} across four task suites.}}
\centering
\footnotesize
\setlength{\tabcolsep}{5pt}
\label{tab:libero-plus-long}
\vspace{-5pt}

\resizebox{\columnwidth}{!}{
\begin{tabular}{l|cccc|c}
    \toprule
    \textbf{Method}
    & \textbf{Plus-S}
    & \textbf{Plus-O}
    & \textbf{Plus-G}
    & \textbf{Plus-10}
    & \textbf{Avg.} \\
    \midrule

    OpenVLA~\textit{[CoRL'24]}~\cite{openvla}
    & 19.4 & 14.0 & 15.1 & 14.3 & 15.6 \\

    OpenVLA-OFT~\textit{[RSS'25]}~\cite{oft}
    & 84.0 & 66.5 & 63.0 & 66.4 & 69.6 \\

    UniVLA~\textit{[RSS'25]}~\cite{univla}
    & 55.5 & 36.7 & 40.7 & 39.9 & 42.9 \\

    WorldVLA~\textit{[arXiv'25]}~\cite{cen2025worldvla}
    & 32.5 & 28.6 & 31.8 & 8.2 & 25.0 \\

    DreamVLA~\textit{[NeurIPS'25]}~\cite{zhang2026dreamvla}
    & 79.7 & 79.0 & 61.7 & 59.8 & 69.9 \\

    $\pi_0$~\textit{[RSS'25]}~\cite{pi0}
    & 60.7 & 61.4 & 44.9 & 48.4 & 53.6 \\

    $\pi_{0.5}$~\textit{[CoRL'25]}~\cite{pi05}
    & 79.9 & 87.8 & 69.0 & 64.9 & 73.6 \\

    Spatial Forcing~\textit{[ICLR'26]}~\cite{li2025spatial}
    & 52.9 & 31.0 & 28.2 & 5.4 & 29.1 \\

    VLA-Adapter~\textit{[AAAI'26]}~\cite{wang2025vlaadapter}
    & 85.0 & 46.3 & 56.0 & 50.4 & 59.1 \\

    GuidedVLA~\textit{[RSS'26]}~\cite{jia2026guidedvla}
    & 84.0 & 80.9 & 70.8 & 66.2 & 75.4 \\

    \midrule
    \rowcolor[HTML]{EFE9F7} 
    \textbf{UniMPA}
    & \textbf{91.0}
    & \textbf{89.7}
    & \textbf{84.3}
    & \textbf{76.5}
    & \textbf{85.3} \\

    \bottomrule
\end{tabular}
}
\vspace{-3pt}
\end{table}

%% file: Tables/vlabench.tex
\begin{table}[h]
\vspace{-4mm}

\caption{\textbf{Results on VLABench}~\cite{zhang2024vlabench}, including Select Painting, Select Book, Select Drink, Select Chemistry Tube, Add Condiment, and Insert Flower.}
\vspace{-5pt}
\centering
\footnotesize
\setlength{\tabcolsep}{2.3pt}
\label{tab:vlabench}

\resizebox{\columnwidth}{!}{
\begin{tabular}{l|cccccc}
\toprule
\textbf{Method}
& \textbf{Painting}
& \textbf{Book}
& \textbf{Drink}
& \textbf{Tube}
& \textbf{Condiment}
& \textbf{Flower} \\
\midrule

Octo~\textit{[RSS'24]}~\cite{octo}
& 6.2 & 0.0 & 0.0 & 1.5 & 3.1 & 1.5 \\

OpenVLA~\textit{[CoRL'24]}~\cite{openvla}
& \textbf{40.2} & 7.7 & 8.5 & 7.7 & 12.4 & 13.9 \\

RDT~\textit{[ICLR'25]}~\cite{liu2024rdt}
& 35.2 & 3.1 & 7.7 & 12.4 & 21.5 & 21.5 \\

$\pi_{0.5}$~\textit{[CoRL'25]}~\cite{pi05}
& 30.0 & 54.0 & 42.0 & 36.0 & \textbf{56.0} & 20.0 \\

FastV~\textit{[ECCV'24]}~\cite{fastv}
& 28.0 & 38.8 & 18.0 & 35.4 & 44.0 & 2.0 \\

SparseVLM~\textit{[ICML'25]}~\cite{zhang2024sparsevlm}
& 32.0 & 52.2 & 28.6 & 28.6 & 50.0 & 0.0 \\

DivPrune~\textit{[CVPR'25]}~\cite{alvar2025divprune}
& 22.0 & 8.0 & 12.2 & 6.0 & 6.0 & 0.0 \\

VLA-Cache~\textit{[NeurIPS'25]}~\cite{xu2025vlaCACHE}
& 32.0 & 42.9 & 32.0 & 30.0 & 42.0 & 10.0 \\

VLA-IAP~\textit{[arXiv'26]}~\cite{cheng2026vla}
& 36.0 & 40.4 & 42.0 & \textbf{41.7} & 55.0 & 12.0 \\

\midrule
\rowcolor[HTML]{EFE9F7}
\textbf{UniMPA}
& 38.0
& \textbf{59.1}
& \textbf{51.0}
& \textbf{41.7}
& 52.0
& \textbf{22.0} \\

\bottomrule
\end{tabular}
}
\vspace{-8pt}

\end{table}

%% file: Tables/real-world.tex
\begin{table*}[t]
\vspace{-4mm}
\caption{
\textbf{Real-world manipulation results on GALAXEA R1 Lite platform across seven task suites.}
For each task, we report cumulative progressive checkpoint success over 25 independent trials.
``$\dag$'' denotes our reproduced results.
}
\vspace{-5pt}
\label{tab:real_world}
\centering
\footnotesize
\setlength{\tabcolsep}{3.2pt}


\noindent
{%
\setlength{\fboxsep}{1pt}%
\colorbox[HTML]{EFEFEF}{%
    \makebox[\dimexpr\textwidth-2\fboxsep\relax][c]{%
        \fontsize{6.5pt}{7.5pt}\selectfont
        \textbf{Task Suite A: Semantic Rearrangement and Sorting}%
    }%
}%
}
\par\vspace{-0.4pt}
\setlength{\tabcolsep}{5.5pt}
\resizebox{\textwidth}{!}{%
\begin{tabular}{l|ccccc|cccc|cccc|c|c}
\toprule
\textbf{Method}
& \multicolumn{5}{c|}{\textbf{Fruit Placement}}
& \multicolumn{4}{c|}{\textbf{Object Position Swapping}}
& \multicolumn{4}{c|}{\textbf{Semantic Table Clearing}}
& \multicolumn{2}{c}{\textbf{Average}} \\

& \shortstack{Grasp\\Apple}
& \shortstack{+Apple\\$\rightarrow$Plate}
& \shortstack{+Grasp\\Banana}
& \shortstack{+Banana\\$\rightarrow$Plate}
& \textbf{CSR}

& \shortstack{Cup\\$\rightarrow$Temporary}
& \shortstack{+Can$\rightarrow$Cup\\Position}
& \shortstack{+Cup$\rightarrow$Can\\Position}
& \textbf{CSR}

& \shortstack{Tool\\$\rightarrow$Plate}
& \shortstack{+Fruit\\$\rightarrow$Bowl}
& \shortstack{+Trash\\$\rightarrow$Bin}
& \textbf{CSR}

& \textbf{TSR}
& \textbf{CSR} \\
\midrule

OpenVLA-OFT~\cite{oft}$\dag$
& 15{\fontsize{6pt}{7pt}\selectfont /25}
& 14{\fontsize{6pt}{7pt}\selectfont /25}
& 11{\fontsize{6pt}{7pt}\selectfont /25}
& 10{\fontsize{6pt}{7pt}\selectfont /25}
& 50.0\%
& 23{\fontsize{6pt}{7pt}\selectfont /25}
& 20{\fontsize{6pt}{7pt}\selectfont /25}
& 16{\fontsize{6pt}{7pt}\selectfont /25}
& 78.7\%
& 22{\fontsize{6pt}{7pt}\selectfont /25}
& 20{\fontsize{6pt}{7pt}\selectfont /25}
& 17{\fontsize{6pt}{7pt}\selectfont /25}
& 78.7\%
& 57.3\%
& 69.1\% \\

$\pi_{0.5}$~\cite{pi05}${\dagger}$
& 19{\fontsize{6pt}{7pt}\selectfont /25}
& 19{\fontsize{6pt}{7pt}\selectfont /25}
& 16{\fontsize{6pt}{7pt}\selectfont /25}
& 14{\fontsize{6pt}{7pt}\selectfont /25}
& 68.0\%
& \textbf{25{\fontsize{6pt}{7pt}\selectfont /25}}
& 22{\fontsize{6pt}{7pt}\selectfont /25}
& 19{\fontsize{6pt}{7pt}\selectfont /25}
& 88.0\%
& \textbf{24{\fontsize{6pt}{7pt}\selectfont /25}}
& 22{\fontsize{6pt}{7pt}\selectfont /25}
& \textbf{20{\fontsize{6pt}{7pt}\selectfont /25}}
& 88.0\%
& 70.7\%
& 81.3\% \\

\rowcolor[HTML]{F7EEEA}
\textbf{UniMPA}
& \textbf{22{\fontsize{6pt}{7pt}\selectfont /25}}
& \textbf{22{\fontsize{6pt}{7pt}\selectfont /25}}
& \textbf{20{\fontsize{6pt}{7pt}\selectfont /25}}
& \textbf{18{\fontsize{6pt}{7pt}\selectfont /25}}
& \textbf{82.0\%}
& \textbf{25{\fontsize{6pt}{7pt}\selectfont /25}}
& \textbf{24{\fontsize{6pt}{7pt}\selectfont /25}}
& \textbf{22{\fontsize{6pt}{7pt}\selectfont /25}}
& \textbf{94.7\%}
& \textbf{24{\fontsize{6pt}{7pt}\selectfont /25}}
& \textbf{23{\fontsize{6pt}{7pt}\selectfont /25}}
& \textbf{20{\fontsize{6pt}{7pt}\selectfont /25}}
& \textbf{89.3\%}
& \textbf{80.0\%}
& \textbf{88.7\%} \\
\bottomrule
\end{tabular}%
}

\vspace{4pt}


\noindent
{%
\setlength{\fboxsep}{1pt}%
\colorbox[HTML]{EFEFEF}{%
    \makebox[\dimexpr\textwidth-2\fboxsep\relax][c]{%
        \fontsize{6.5pt}{7.5pt}\selectfont
        \textbf{Task Suite B: Articulated and Container Interaction}%
    }%
}%
}
\par\vspace{-0.4pt}
\setlength{\tabcolsep}{8pt}
\resizebox{\textwidth}{!}{%
\begin{tabular}{l|cccc|cccc|cccc|c|c}
\toprule
\textbf{Method}
& \multicolumn{4}{c|}{\textbf{Microwave Manipulation}}
& \multicolumn{4}{c|}{\textbf{Pot Lid Replacement}}
& \multicolumn{4}{c|}{\textbf{Water Pouring}}
& \multicolumn{2}{c}{\textbf{Average}} \\

& \shortstack{Grasp\\Bowl}
& \shortstack{+Place\\Bowl Inside}
& \shortstack{+Close\\Door}
& \textbf{CSR}

& \shortstack{Remove\\Lid}
& \shortstack{+Place\\Object}
& \shortstack{+Replace\\Lid}
& \textbf{CSR}

& \shortstack{Grasp\\Kettle}
& \shortstack{+Pour\\Water}
& \shortstack{+Place\\Kettle}
& \textbf{CSR}

& \textbf{TSR}
& \textbf{CSR} \\
\midrule

OpenVLA-OFT~\cite{oft}$\dag$
& 18{\fontsize{6pt}{7pt}\selectfont /25}
& 15{\fontsize{6pt}{7pt}\selectfont /25}
& 14{\fontsize{6pt}{7pt}\selectfont /25}
& 62.7\%
& 17{\fontsize{6pt}{7pt}\selectfont /25}
& 14{\fontsize{6pt}{7pt}\selectfont /25}
& 12{\fontsize{6pt}{7pt}\selectfont /25}
& 57.3\%
& 19{\fontsize{6pt}{7pt}\selectfont /25}
& 15{\fontsize{6pt}{7pt}\selectfont /25}
& 14{\fontsize{6pt}{7pt}\selectfont /25}
& 64.0\%
& 53.3\%
& 61.3\% \\

$\pi_{0.5}$~\cite{pi05}${\dagger}$
& \textbf{25{\fontsize{6pt}{7pt}\selectfont /25}}
& 20{\fontsize{6pt}{7pt}\selectfont /25}
& 19{\fontsize{6pt}{7pt}\selectfont /25}
& 85.3\%
& 21{\fontsize{6pt}{7pt}\selectfont /25}
& 19{\fontsize{6pt}{7pt}\selectfont /25}
& 17{\fontsize{6pt}{7pt}\selectfont /25}
& 76.0\%
& 23{\fontsize{6pt}{7pt}\selectfont /25}
& 20{\fontsize{6pt}{7pt}\selectfont /25}
& 19{\fontsize{6pt}{7pt}\selectfont /25}
& 82.7\%
& 73.3\%
& 81.3\% \\

\rowcolor[HTML]{F7EEEA}
\textbf{UniMPA}
& 24{\fontsize{6pt}{7pt}\selectfont /25}
& \textbf{23{\fontsize{6pt}{7pt}\selectfont /25}}
& \textbf{21{\fontsize{6pt}{7pt}\selectfont /25}}
& \textbf{90.7\%}
& \textbf{24{\fontsize{6pt}{7pt}\selectfont /25}}
& \textbf{22{\fontsize{6pt}{7pt}\selectfont /25}}
& \textbf{19{\fontsize{6pt}{7pt}\selectfont /25}}
& \textbf{86.7\%}
& \textbf{25{\fontsize{6pt}{7pt}\selectfont /25}}
& \textbf{23{\fontsize{6pt}{7pt}\selectfont /25}}
& \textbf{23{\fontsize{6pt}{7pt}\selectfont /25}}
& \textbf{94.7\%}
& \textbf{84.0\%}
& \textbf{90.7\%} \\

\bottomrule
\end{tabular}%
}

\vspace{4pt}


\noindent
{%
\setlength{\fboxsep}{1pt}%
\colorbox[HTML]{EFEFEF}{%
    \makebox[\dimexpr\textwidth-2\fboxsep\relax][c]{%
        \fontsize{6.5pt}{7.5pt}\selectfont
        \textbf{Task Suite C: Precision Assembly and Geometric Manipulation}%
    }%
}%
}
\par\vspace{-0.4pt}
\setlength{\tabcolsep}{5pt}
\resizebox{\textwidth}{!}{%
\begin{tabular}{l|ccccc|ccc|ccccc|c|c}
\toprule
\textbf{Method}
& \multicolumn{5}{c|}{\textbf{Block Stacking}}
& \multicolumn{3}{c|}{\textbf{Ring Stacking}}
& \multicolumn{5}{c|}{\textbf{Plug Insertion and Activation}}
& \multicolumn{2}{c}{\textbf{Average}} \\

& \shortstack{Grasp\\Red Block}
& \shortstack{+Place at\\Center}
& \shortstack{+Place\\Blue Block}
& \shortstack{+Place\\Yellow Block}
& \textbf{CSR}

& \shortstack{Large Ring\\$\rightarrow$Pole}
& \shortstack{+Small Ring\\$\rightarrow$Pole}
& \textbf{CSR}

& \shortstack{Grasp\\Plug}
& \shortstack{+Align with\\Socket}
& \shortstack{+Insert\\Plug}
& \shortstack{+Press\\Switch}
& \textbf{CSR}

& \textbf{TSR}
& \textbf{CSR} \\
\midrule

OpenVLA-OFT~\cite{oft}$\dag$
& 19{\fontsize{6pt}{7pt}\selectfont /25}
& 17{\fontsize{6pt}{7pt}\selectfont /25}
& 14{\fontsize{6pt}{7pt}\selectfont /25}
& 13{\fontsize{6pt}{7pt}\selectfont /25}
& 63.0\%
& 21{\fontsize{6pt}{7pt}\selectfont /25}
& 18{\fontsize{6pt}{7pt}\selectfont /25}
& 78.0\%
& 16{\fontsize{6pt}{7pt}\selectfont /25}
& 13{\fontsize{6pt}{7pt}\selectfont /25}
& 10{\fontsize{6pt}{7pt}\selectfont /25}
& 8{\fontsize{6pt}{7pt}\selectfont /25}
& 47.0\%
& 52.0\%
& 62.7\% \\

$\pi_{0.5}$~\cite{pi05}${\dagger}$
& 22{\fontsize{6pt}{7pt}\selectfont /25}
& 21{\fontsize{6pt}{7pt}\selectfont /25}
& 19{\fontsize{6pt}{7pt}\selectfont /25}
& 18{\fontsize{6pt}{7pt}\selectfont /25}
& 80.0\%
& \textbf{24{\fontsize{6pt}{7pt}\selectfont /25}}
& \textbf{22{\fontsize{6pt}{7pt}\selectfont /25}}
& \textbf{92.0\%}
& 20{\fontsize{6pt}{7pt}\selectfont /25}
& 18{\fontsize{6pt}{7pt}\selectfont /25}
& 15{\fontsize{6pt}{7pt}\selectfont /25}
& 13{\fontsize{6pt}{7pt}\selectfont /25}
& 66.0\%
& 70.7\%
& 79.3\% \\

\rowcolor[HTML]{F7EEEA}
\textbf{UniMPA}
& \textbf{24{\fontsize{6pt}{7pt}\selectfont /25}}
& \textbf{24{\fontsize{6pt}{7pt}\selectfont /25}}
& \textbf{23{\fontsize{6pt}{7pt}\selectfont /25}}
& \textbf{21{\fontsize{6pt}{7pt}\selectfont /25}}
& \textbf{92.0\%}
& \textbf{24{\fontsize{6pt}{7pt}\selectfont /25}}
& 21{\fontsize{6pt}{7pt}\selectfont /25}
& 90.0\%
& \textbf{24{\fontsize{6pt}{7pt}\selectfont /25}}
& \textbf{23{\fontsize{6pt}{7pt}\selectfont /25}}
& \textbf{21{\fontsize{6pt}{7pt}\selectfont /25}}
& \textbf{18{\fontsize{6pt}{7pt}\selectfont /25}}
& \textbf{86.0\%}
& \textbf{80.0\%}
& \textbf{89.3\%} \\

\bottomrule
\end{tabular}%
}

\vspace{4pt}


\noindent
{%
\setlength{\fboxsep}{1pt}%
\colorbox[HTML]{EFEFEF}{%
    \makebox[\dimexpr\textwidth-2\fboxsep\relax][c]{%
        \fontsize{6.5pt}{7.5pt}\selectfont
        \textbf{Task Suite D: Deformable and Tool-Mediated Manipulation}%
    }%
}%
}
\par\vspace{-0.4pt}
\setlength{\tabcolsep}{4.5pt}
\resizebox{\textwidth}{!}{%
\begin{tabular}{l|cccc|ccccc|ccccc|c|c}
\toprule
\textbf{Method}
& \multicolumn{4}{c|}{\textbf{Towel Folding}}
& \multicolumn{5}{c|}{\textbf{Table Sweeping}}
& \multicolumn{5}{c|}{\textbf{Medicine Bottling}}
& \multicolumn{2}{c}{\textbf{Average}} \\

& \shortstack{First\\Fold}
& \shortstack{+Second\\Fold}
& \shortstack{+Final\\Fold}
& \textbf{CSR}

& \shortstack{Grasp\\Broom}
& \shortstack{+Sweep\\Region A}
& \shortstack{+Sweep\\Region B}
& \shortstack{+Trash Inside\\Dustpan}
& \textbf{CSR}

& \shortstack{Grasp\\Medicine}
& \shortstack{+Place\\into Bottle}
& \shortstack{+Grasp\\Cap}
& \shortstack{+Screw\\on Cap}
& \textbf{CSR}

& \textbf{TSR}
& \textbf{CSR} \\
\midrule

OpenVLA-OFT~\cite{oft}$\dag$
& 17{\fontsize{6pt}{7pt}\selectfont /25}
& 14{\fontsize{6pt}{7pt}\selectfont /25}
& 12{\fontsize{6pt}{7pt}\selectfont /25}
& 57.3\%
& 18{\fontsize{6pt}{7pt}\selectfont /25}
& 15{\fontsize{6pt}{7pt}\selectfont /25}
& 12{\fontsize{6pt}{7pt}\selectfont /25}
& 11{\fontsize{6pt}{7pt}\selectfont /25}
& 56.0\%
& 20{\fontsize{6pt}{7pt}\selectfont /25}
& 13{\fontsize{6pt}{7pt}\selectfont /25}
& 12{\fontsize{6pt}{7pt}\selectfont /25}
& 7{\fontsize{6pt}{7pt}\selectfont /25}
& 52.0\%
& 40.0\%
& 55.1\% \\

$\pi_{0.5}$~\cite{pi05}${\dagger}$
& 21{\fontsize{6pt}{7pt}\selectfont /25}
& 19{\fontsize{6pt}{7pt}\selectfont /25}
& 18{\fontsize{6pt}{7pt}\selectfont /25}
& 77.3\%
& 22{\fontsize{6pt}{7pt}\selectfont /25}
& 20{\fontsize{6pt}{7pt}\selectfont /25}
& 18{\fontsize{6pt}{7pt}\selectfont /25}
& 17{\fontsize{6pt}{7pt}\selectfont /25}
& 77.0\%
& 22{\fontsize{6pt}{7pt}\selectfont /25}
& 18{\fontsize{6pt}{7pt}\selectfont /25}
& 17{\fontsize{6pt}{7pt}\selectfont /25}
& 13{\fontsize{6pt}{7pt}\selectfont /25}
& 70.0\%
& 64.0\%
& 74.8\% \\

\rowcolor[HTML]{F7EEEA}
\textbf{UniMPA}
& \textbf{24{\fontsize{6pt}{7pt}\selectfont /25}}
& \textbf{23{\fontsize{6pt}{7pt}\selectfont /25}}
& \textbf{20{\fontsize{6pt}{7pt}\selectfont /25}}
& \textbf{89.3\%}
& \textbf{24{\fontsize{6pt}{7pt}\selectfont /25}}
& \textbf{23{\fontsize{6pt}{7pt}\selectfont /25}}
& \textbf{21{\fontsize{6pt}{7pt}\selectfont /25}}
& \textbf{19{\fontsize{6pt}{7pt}\selectfont /25}}
& \textbf{87.0\%}
& \textbf{23{\fontsize{6pt}{7pt}\selectfont /25}}
& \textbf{22{\fontsize{6pt}{7pt}\selectfont /25}}
& \textbf{21{\fontsize{6pt}{7pt}\selectfont /25}}
& \textbf{17{\fontsize{6pt}{7pt}\selectfont /25}}
& \textbf{83.0\%}
& \textbf{74.7\%}
& \textbf{86.4\%} \\

\bottomrule
\end{tabular}%
}

\vspace{4pt}


\noindent
{%
\setlength{\fboxsep}{1pt}%
\colorbox[HTML]{EFEFEF}{%
    \makebox[\dimexpr\textwidth-2\fboxsep\relax][c]{%
        \fontsize{6.5pt}{7.5pt}\selectfont
        \textbf{Task Suite E: Bimanual Coordination}%
    }%
}%
}
\par\vspace{-0.4pt}
\setlength{\tabcolsep}{6.5pt}
\resizebox{\textwidth}{!}{%
\begin{tabular}{l|cccc|cccc|cccc|c|c}
\toprule
\textbf{Method}
& \multicolumn{4}{c|}{\textbf{Bimanual Transfer}}
& \multicolumn{4}{c|}{\textbf{Bimanual Bag Packing}}
& \multicolumn{4}{c|}{\textbf{Tray Transportation}}
& \multicolumn{2}{c}{\textbf{Average}} \\

& \shortstack{Right-Arm\\Grasp}
& \shortstack{+Hand-\\over}
& \shortstack{+Place\\in Bowl}
& \textbf{CSR}

& \shortstack{Hold Bag\\Open}
& \shortstack{+Insert\\Toy}
& \shortstack{+Close\\Bag}
& \textbf{CSR}

& \shortstack{Dual-Arm\\Grasp}
& \shortstack{+Transport\\Tray}
& \shortstack{+Place\\Toy}
& \textbf{CSR}

& \textbf{TSR}
& \textbf{CSR} \\
\midrule

OpenVLA-OFT~\cite{oft}$\dag$
& 18{\fontsize{6pt}{7pt}\selectfont /25}
& 13{\fontsize{6pt}{7pt}\selectfont /25}
& 11{\fontsize{6pt}{7pt}\selectfont /25}
& 56.0\%
& 17{\fontsize{6pt}{7pt}\selectfont /25}
& 13{\fontsize{6pt}{7pt}\selectfont /25}
& 12{\fontsize{6pt}{7pt}\selectfont /25}
& 56.0\%
& 19{\fontsize{6pt}{7pt}\selectfont /25}
& 16{\fontsize{6pt}{7pt}\selectfont /25}
& 13{\fontsize{6pt}{7pt}\selectfont /25}
& 64.0\%
& 48.0\%
& 58.7\% \\

$\pi_{0.5}$~\cite{pi05}${\dagger}$
& 22{\fontsize{6pt}{7pt}\selectfont /25}
& 18{\fontsize{6pt}{7pt}\selectfont /25}
& 17{\fontsize{6pt}{7pt}\selectfont /25}
& 76.0\%
& 20{\fontsize{6pt}{7pt}\selectfont /25}
& 17{\fontsize{6pt}{7pt}\selectfont /25}
& 17{\fontsize{6pt}{7pt}\selectfont /25}
& 72.0\%
& 22{\fontsize{6pt}{7pt}\selectfont /25}
& 20{\fontsize{6pt}{7pt}\selectfont /25}
& 18{\fontsize{6pt}{7pt}\selectfont /25}
& 80.0\%
& 69.3\%
& 76.0\% \\

\rowcolor[HTML]{F7EEEA}
\textbf{UniMPA}
& \textbf{24{\fontsize{6pt}{7pt}\selectfont /25}}
& \textbf{21{\fontsize{6pt}{7pt}\selectfont /25}}
& \textbf{21{\fontsize{6pt}{7pt}\selectfont /25}}
& \textbf{88.0\%}
& \textbf{21{\fontsize{6pt}{7pt}\selectfont /25}}
& \textbf{20{\fontsize{6pt}{7pt}\selectfont /25}}
& \textbf{20{\fontsize{6pt}{7pt}\selectfont /25}}
& \textbf{81.3\%}
& \textbf{24{\fontsize{6pt}{7pt}\selectfont /25}}
& \textbf{23{\fontsize{6pt}{7pt}\selectfont /25}}
& \textbf{21{\fontsize{6pt}{7pt}\selectfont /25}}
& \textbf{90.7\%}
& \textbf{82.7\%}
& \textbf{86.7\%} \\

\bottomrule
\end{tabular}%
}

\vspace{4pt}


\noindent
{%
\setlength{\fboxsep}{1pt}%
\colorbox[HTML]{EFEFEF}{%
    \makebox[\dimexpr\textwidth-2\fboxsep\relax][c]{%
        \fontsize{6.5pt}{7.5pt}\selectfont
        \textbf{Task Suite F: Dynamic and Reactive Manipulation}%
    }%
}%
}
\par\vspace{-0.4pt}
\setlength{\tabcolsep}{5.5pt}
\resizebox{\textwidth}{!}{%
\begin{tabular}{l|ccccc|cccc|ccc|c|c}
\toprule
\textbf{Method}
& \multicolumn{5}{c|}{\textbf{Conveyor Interception}}
& \multicolumn{4}{c|}{\textbf{Rolling-Ball Interception}}
& \multicolumn{3}{c|}{\textbf{Dynamic Placement}}
& \multicolumn{2}{c}{\textbf{Average}} \\

& \shortstack{Intercept\\First Target}
& \shortstack{+Place\\in Plate}
& \shortstack{+Intercept\\Second Target}
& \shortstack{+Place\\in Plate}
& \textbf{CSR}

& \shortstack{Intercept\\Ball}
& \shortstack{+Stabilize\\Grasp}
& \shortstack{+Place\\in Bowl}
& \textbf{CSR}

& \shortstack{Grasp\\Banana}
& \shortstack{+Place Behind\\Apple}
& \textbf{CSR}

& \textbf{TSR}
& \textbf{CSR} \\
\midrule

OpenVLA-OFT~\cite{oft}$\dag$
& 15{\fontsize{6pt}{7pt}\selectfont /25}
& 11{\fontsize{6pt}{7pt}\selectfont /25}
& 8{\fontsize{6pt}{7pt}\selectfont /25}
& 7{\fontsize{6pt}{7pt}\selectfont /25}
& 41.0\%
& 14{\fontsize{6pt}{7pt}\selectfont /25}
& 9{\fontsize{6pt}{7pt}\selectfont /25}
& 8{\fontsize{6pt}{7pt}\selectfont /25}
& 41.3\%
& 19{\fontsize{6pt}{7pt}\selectfont /25}
& 13{\fontsize{6pt}{7pt}\selectfont /25}
& 64.0\%
& 37.3\%
& 48.8\% \\

$\pi_{0.5}$~\cite{pi05}${\dagger}$
& 19{\fontsize{6pt}{7pt}\selectfont /25}
& 17{\fontsize{6pt}{7pt}\selectfont /25}
& 13{\fontsize{6pt}{7pt}\selectfont /25}
& 12{\fontsize{6pt}{7pt}\selectfont /25}
& 61.0\%
& 19{\fontsize{6pt}{7pt}\selectfont /25}
& 14{\fontsize{6pt}{7pt}\selectfont /25}
& 13{\fontsize{6pt}{7pt}\selectfont /25}
& 61.3\%
& \textbf{23{\fontsize{6pt}{7pt}\selectfont /25}}
& \textbf{18{\fontsize{6pt}{7pt}\selectfont /25}}
& \textbf{82.0\%}
& 57.3\%
& 68.1\% \\

\rowcolor[HTML]{F7EEEA}
\textbf{UniMPA}
& \textbf{23{\fontsize{6pt}{7pt}\selectfont /25}}
& \textbf{22{\fontsize{6pt}{7pt}\selectfont /25}}
& \textbf{19{\fontsize{6pt}{7pt}\selectfont /25}}
& \textbf{17{\fontsize{6pt}{7pt}\selectfont /25}}
& \textbf{81.0\%}
& \textbf{20{\fontsize{6pt}{7pt}\selectfont /25}}
& \textbf{18{\fontsize{6pt}{7pt}\selectfont /25}}
& \textbf{18{\fontsize{6pt}{7pt}\selectfont /25}}
& \textbf{74.7\%}
& \textbf{23{\fontsize{6pt}{7pt}\selectfont /25}}
& \textbf{18{\fontsize{6pt}{7pt}\selectfont /25}}
& \textbf{82.0\%}
& \textbf{70.7\%}
& \textbf{79.2\%} \\

\bottomrule
\end{tabular}%
}

\vspace{4pt}


\noindent
{%
\setlength{\fboxsep}{1pt}%
\colorbox[HTML]{EFEFEF}{%
    \makebox[\dimexpr\textwidth-2\fboxsep\relax][c]{%
        \fontsize{6.5pt}{7.5pt}\selectfont
        \textbf{Task Suite G: Long-Horizon Composition and Recovery}%
    }%
}%
}
\par\vspace{-0.4pt}
\setlength{\tabcolsep}{2.5pt}
\resizebox{\textwidth}{!}{%
\begin{tabular}{l|cccccc|ccccc|cccccc|c|c}
\toprule
\textbf{Method}
& \multicolumn{6}{c|}{\textbf{Stack Recovery}}
& \multicolumn{5}{c|}{\textbf{Drawer Cleanup}}
& \multicolumn{6}{c|}{\textbf{Folding Recovery}}
& \multicolumn{2}{c}{\textbf{Average}} \\

& \shortstack{Place\\Red Block}
& \shortstack{+Place\\Blue Block}
& \shortstack{+Recover\\Grasp}
& \shortstack{+Place\\Blue Block}
& \shortstack{+Place\\Yellow Block}
& \textbf{CSR}

& \shortstack{Open\\Drawer}
& \shortstack{+Insert\\Toy A}
& \shortstack{+Insert\\Toy B}
& \shortstack{+Close\\Drawer}
& \textbf{CSR}

& \shortstack{Grasp\\T-shirt}
& \shortstack{+First\\Fold}
& \shortstack{+Second\\Fold}
& \shortstack{+Realign\\T-shirt}
& \shortstack{+Final\\Fold}
& \textbf{CSR}

& \textbf{TSR}
& \textbf{CSR} \\
\midrule

OpenVLA-OFT~\cite{oft}$\dag$
& 21{\fontsize{6pt}{7pt}\selectfont /25}
& 17{\fontsize{6pt}{7pt}\selectfont /25}
& 11{\fontsize{6pt}{7pt}\selectfont /25}
& 9{\fontsize{6pt}{7pt}\selectfont /25}
& 8{\fontsize{6pt}{7pt}\selectfont /25}
& 52.8\%
& 17{\fontsize{6pt}{7pt}\selectfont /25}
& 15{\fontsize{6pt}{7pt}\selectfont /25}
& 13{\fontsize{6pt}{7pt}\selectfont /25}
& 12{\fontsize{6pt}{7pt}\selectfont /25}
& 57.0\%
& 20{\fontsize{6pt}{7pt}\selectfont /25}
& 12{\fontsize{6pt}{7pt}\selectfont /25}
& 10{\fontsize{6pt}{7pt}\selectfont /25}
& 7{\fontsize{6pt}{7pt}\selectfont /25}
& 4{\fontsize{6pt}{7pt}\selectfont /25}
& 42.4\%
& 32.0\%
& 50.7\% \\

$\pi_{0.5}$~\cite{pi05}${\dagger}$
& \textbf{24{\fontsize{6pt}{7pt}\selectfont /25}}
& \textbf{23{\fontsize{6pt}{7pt}\selectfont /25}}
& 16{\fontsize{6pt}{7pt}\selectfont /25}
& 14{\fontsize{6pt}{7pt}\selectfont /25}
& 13{\fontsize{6pt}{7pt}\selectfont /25}
& 72.0\%
& 21{\fontsize{6pt}{7pt}\selectfont /25}
& 19{\fontsize{6pt}{7pt}\selectfont /25}
& 17{\fontsize{6pt}{7pt}\selectfont /25}
& 17{\fontsize{6pt}{7pt}\selectfont /25}
& 74.0\%
& 22{\fontsize{6pt}{7pt}\selectfont /25}
& 17{\fontsize{6pt}{7pt}\selectfont /25}
& 14{\fontsize{6pt}{7pt}\selectfont /25}
& 12{\fontsize{6pt}{7pt}\selectfont /25}
& 9{\fontsize{6pt}{7pt}\selectfont /25}
& 59.2\%
& 52.0\%
& 68.4\% \\

\rowcolor[HTML]{F7EEEA}
\textbf{UniMPA}
& \textbf{24{\fontsize{6pt}{7pt}\selectfont /25}}
& \textbf{23{\fontsize{6pt}{7pt}\selectfont /25}}
& \textbf{21{\fontsize{6pt}{7pt}\selectfont /25}}
& \textbf{20{\fontsize{6pt}{7pt}\selectfont /25}}
& \textbf{18{\fontsize{6pt}{7pt}\selectfont /25}}
& \textbf{84.8\%}
& \textbf{23{\fontsize{6pt}{7pt}\selectfont /25}}
& \textbf{21{\fontsize{6pt}{7pt}\selectfont /25}}
& \textbf{21{\fontsize{6pt}{7pt}\selectfont /25}}
& \textbf{20{\fontsize{6pt}{7pt}\selectfont /25}}
& \textbf{85.0\%}
& \textbf{23{\fontsize{6pt}{7pt}\selectfont /25}}
& \textbf{22{\fontsize{6pt}{7pt}\selectfont /25}}
& \textbf{20{\fontsize{6pt}{7pt}\selectfont /25}}
& \textbf{19{\fontsize{6pt}{7pt}\selectfont /25}}
& \textbf{16{\fontsize{6pt}{7pt}\selectfont /25}}
& \textbf{80.0\%}
& \textbf{72.0\%}
& \textbf{83.3\%} \\

\bottomrule
\end{tabular}%
}

\vspace{-8pt}
\end{table*}

%% file: Tables/real-world-aloha.tex
\begin{table*}[t]
\caption{\textbf{Real-world manipulation results on AgileX Cobot Magic platform across seven tasks.}
For each task, we report TSR and CSR over 25 independent trials.
``$\dag$'' denotes our reproduced results.}
\vspace{-5pt}
\centering
\footnotesize
\setlength{\tabcolsep}{4pt}
\label{tab:aloha}
\resizebox{\textwidth}{!}{
    \begin{tabular}{l|cc|cc|cc|cc|cc|cc|cc|cc}
    \toprule
    \multirow{2}{*}{\textbf{Method}}
    & \multicolumn{2}{c|}{\textbf{Fruit Placement}}
    & \multicolumn{2}{c|}{\textbf{Microwave Manipulation}}
    & \multicolumn{2}{c|}{\textbf{Block Stacking}}
    & \multicolumn{2}{c|}{\textbf{Towel Folding}}
    & \multicolumn{2}{c|}{\textbf{Bimanual Transfer}}
    & \multicolumn{2}{c|}{\textbf{Conveyor Interception}}
    & \multicolumn{2}{c|}{\textbf{Stack Recovery}}
    & \multicolumn{2}{c}{\textbf{Average}} \\
    
    & TSR & CSR
    & TSR & CSR
    & TSR & CSR
    & TSR & CSR
    & TSR & CSR
    & TSR & CSR
    & TSR & CSR
    & TSR & CSR \\
    \midrule

    OpenVLA-OFT~\cite{oft}$\dag$
    & 32.0\% & 50.0\%
    & 48.0\% & 62.7\%
    & 40.0\% & 59.0\%
    & 44.0\% & 58.7\%
    & 36.0\% & 57.0\%
    & 28.0\% & 44.0\%
    & 32.0\% & 55.2\%
    & 37.1\% & 55.2\% \\

    $\pi_{0.5}$~\cite{pi05}${\dagger}$
    & 52.0\% & 67.0\%
    & 72.0\% & 84.0\%
    & 68.0\% & 78.0\%
    & 72.0\% & 80.0\%
    & 72.0\% & 81.0\%
    & 40.0\% & 60.0\%
    & 60.0\% & 74.4\%
    & 62.3\% & 74.9\% \\

    \midrule
    \rowcolor[HTML]{F7EEEA} 
    \textbf{UniMPA} 
    & \textbf{64.0\%} & \textbf{80.0\%}
    & \textbf{80.0\%} & \textbf{90.7\%}
    & \textbf{76.0\%} & \textbf{90.0\%}
    & \textbf{76.0\%} & \textbf{88.0\%}
    & \textbf{80.0\%} & \textbf{86.7\%}
    & \textbf{76.0\%} & \textbf{85.0\%}
    & \textbf{72.0\%} & \textbf{84.8\%}
    & \textbf{74.9\%} & \textbf{86.4\%} \\
    
    \bottomrule
    \end{tabular}
}
\vspace{-8pt}

\end{table*}

%% file: Tables/ablation-perdition.tex
\begin{table}[t]
\centering
\caption{\textbf{Ablation study on Persistent-Selective Future Prediction.} We isolate latent- and pixel-level future modeling and compare dense pixel supervision with the proposed persistent latent plus selectively triggered pixel prediction.}
\vspace{-5pt}

\label{tab:ablation_prediction}
\setlength{\tabcolsep}{4.5pt}
\renewcommand{\arraystretch}{1.05}
\resizebox{\columnwidth}{!}{%
    \begin{tabular}{l|ccccc|cc}
    \toprule
    \textbf{Prediction Design}
    & \textbf{Spatial}
    & \textbf{Object}
    & \textbf{Goal}
    & \textbf{Long}
    & \textbf{Avg.}
    & \textbf{RW TSR}
    & \textbf{RW CSR} \\
    \midrule
    
    Only Latent Prediction
    & 97.8 & 98.4 & 96.2 & 88.6 & 95.3
    & 64.6 & 78.2 \\
    
    Only Pixel Prediction
    & 98.8 & 99.0 & 97.2 & 92.6 & 96.9
    & 68.6 & 81.2 \\
    
    Dense Pixel Prediction
    & 99.0 & 99.2 & 97.6 & 93.4 & 97.3
    & 70.3 & 83.0 \\
    
    \textbf{Persistent + Selective (Ours)}
    & \textbf{99.6}
    & \textbf{99.8}
    & \textbf{98.8}
    & \textbf{96.0}
    & \textbf{98.6}
    & \textbf{74.9}
    & \textbf{86.4} \\
    \bottomrule
    \end{tabular}%
}
\vspace{-5pt}

\end{table}

%% file: Tables/ablation-gate.tex
\begin{table}[t]
\centering
\caption{\textbf{Ablation study on Transition-Critical Trigger strategies.} We compare random, action-only, latent-only, and the proposed combined latent-and-action trigger.}
\vspace{-5pt}

\label{tab:ablation_trigger}
\setlength{\tabcolsep}{4.5pt}
\renewcommand{\arraystretch}{1.05}
\resizebox{\columnwidth}{!}{%
    \begin{tabular}{l|ccccc|cc}
    \toprule
    \textbf{Trigger Strategy}
    & \textbf{Spatial}
    & \textbf{Object}
    & \textbf{Goal}
    & \textbf{Long}
    & \textbf{Avg.}
    & \textbf{RW TSR}
    & \textbf{RW CSR} \\
    \midrule
    
    Random Gate
    & 98.0 & 98.6 & 96.2 & 91.2 & 96.0
    & 65.1 & 79.1 \\

    Action-only Gate
    & 98.8 & 99.0 & 97.4 & 92.8 & 97.0
    & 68.6 & 81.3 \\
    
    Latent-only Gate
    & 99.0 & 99.2 & 97.8 & 93.6 & 97.4
    & 69.7 & 83.0 \\
    
    \textbf{Latent + Action Gate (Ours)}
    & \textbf{99.6}
    & \textbf{99.8}
    & \textbf{98.8}
    & \textbf{96.0}
    & \textbf{98.6}
    & \textbf{74.9}
    & \textbf{86.4} \\
    
    \bottomrule
    \end{tabular}%
}
\vspace{-5pt}

\end{table}

%% file: Tables/ablation-memory.tex
\begin{table}[t]
\centering
\caption{\textbf{Ablation study on bidirectional memory design.} We remove the Visual-Action and Action-Visual Memory Banks and jointly to evaluate their complementary roles.}
\vspace{-5pt}

\label{tab:ablation_memory}
\setlength{\tabcolsep}{4.5pt}
\renewcommand{\arraystretch}{1.05}
\resizebox{\columnwidth}{!}{%
    \begin{tabular}{l|ccccc|cc}
    \toprule
    \textbf{Memory Design}
    & \textbf{Spatial}
    & \textbf{Object}
    & \textbf{Goal}
    & \textbf{Long}
    & \textbf{Avg.}
    & \textbf{RW TSR}
    & \textbf{RW CSR} \\
    \midrule
    
    w/o Memory
    & 97.0 & 97.6 & 95.2 & 88.0 & 94.5
    & 61.1 & 75.7 \\

    w/o Visual-Action Bank
    & 98.8 & 99.0 & 96.6 & 91.4 & 96.5
    & 66.9 & 80.1 \\
    
    w/o Action-Visual Bank
    & 99.0 & 99.4 & 97.2 & 92.6 & 97.1
    & 68.0 & 81.4 \\
    
    \textbf{Bidirectional Banks (Ours)}
    & \textbf{99.6}
    & \textbf{99.8}
    & \textbf{98.8}
    & \textbf{96.0}
    & \textbf{98.6}
    & \textbf{74.9}
    & \textbf{86.4} \\
    
    \bottomrule
    \end{tabular}%
}
\vspace{-5pt}

\end{table}

%% file: Tables/ablation-coarse2fine.tex
\begin{table}[t]
\centering
\caption{ \textbf{Ablation study on temporal retrieval strategies.} Coarse-to-fine temporal retrieval outperforms retrieval without temporal modeling and fixed local-window search. }
\vspace{-5pt}

\label{tab:ablation_temporal_retrieval}
\setlength{\tabcolsep}{4.5pt}
\renewcommand{\arraystretch}{1.05}
\resizebox{\columnwidth}{!}{%
    \begin{tabular}{l|ccccc|cc}
    \toprule
    \textbf{Temporal Retrieval}
    & \textbf{Spatial}
    & \textbf{Object}
    & \textbf{Goal}
    & \textbf{Long}
    & \textbf{Avg.}
    & \textbf{RW TSR}
    & \textbf{RW CSR} \\
    \midrule
    
    w/o Temporal Modeling
    & 98.4 & 98.8 & 96.4 & 90.0 & 95.9
    & 65.7 & 79.4 \\

    Fixed Local Window
    & 98.6 & 98.8 & 97.2 & 92.2 & 96.7
    & 68.6 & 81.6 \\
    
    \textbf{Coarse-to-Fine Temporal (Ours)}
    & \textbf{99.6}
    & \textbf{99.8}
    & \textbf{98.8}
    & \textbf{96.0}
    & \textbf{98.6}
    & \textbf{74.9}
    & \textbf{86.4} \\
    
    \bottomrule
    \end{tabular}%
}
\vspace{-5pt}

\end{table}

%% file: Tables/ablation-bank-pretraining.tex
\begin{table}[t]
\centering
\caption{ \textbf{Ablation study on memory pretraining objectives.} Future-oriented visual and action reconstruction together with retrieval-simulation training jointly shape the memory into a transition-aware repository whose retrieved values preserve task-relevant future evolution. }
\vspace{-5pt}

\label{tab:ablation_memory_pretraining}
\setlength{\tabcolsep}{4.5pt}
\renewcommand{\arraystretch}{1.05}
\resizebox{\columnwidth}{!}{%
    \begin{tabular}{l|ccccc|cc}
    \toprule
    \textbf{Memory Pretraining}
    & \textbf{Spatial}
    & \textbf{Object}
    & \textbf{Goal}
    & \textbf{Long}
    & \textbf{Avg.}
    & \textbf{RW TSR}
    & \textbf{RW CSR} \\
    \midrule
    
    Current-State Reconstruction
    & 98.2 & 98.6 & 96.6 & 90.8 & 96.1
    & 62.9 & 76.7 \\
    
    w/o Visual Future Recon.
    & 98.4 & 98.8 & 96.8 & 91.4 & 96.4
    & 64.6 & 78.5 \\
    
    w/o Action Future Recon.
    & 98.6 & 99.0 & 97.2 & 92.0 & 96.7
    & 66.9 & 80.6 \\
    
    w/o Pairing Objective
    & 98.8 & 99.0 & 97.6 & 92.8 & 97.1
    & 69.1 & 82.3 \\
    
    \textbf{Full Pretraining (Ours)}
    & \textbf{99.6}
    & \textbf{99.8}
    & \textbf{98.8}
    & \textbf{96.0}
    & \textbf{98.6}
    & \textbf{74.9}
    & \textbf{86.4} \\
    
    \bottomrule
    \end{tabular}%
}
\vspace{-5pt}

\end{table}

%% file: Tables/ablation-proposer.tex
\begin{table}[t]
\centering
\caption{ \textbf{Ablation study on action prior design.}
Fusing the retrieved prototype with recent action history yields the strongest performance. }
\vspace{-5pt}

\label{tab:ablation_action_prior}
\setlength{\tabcolsep}{4.5pt}
\renewcommand{\arraystretch}{1.05}
    \resizebox{\columnwidth}{!}{%
    \begin{tabular}{l|ccccc|cc}
    \toprule
    \textbf{Action Prior Design}
    & \textbf{Spatial}
    & \textbf{Object}
    & \textbf{Goal}
    & \textbf{Long}
    & \textbf{Avg.}
    & \textbf{RW TSR}
    & \textbf{RW CSR} \\
    \midrule
    
    NN Action Copy
    & 94.6 & 96.2 & 91.8 & 82.4 & 91.3
    & 52.0 & 67.5 \\
    
    w/o Action Proposer
    & 98.4 & 98.8 & 96.6 & 91.0 & 96.2
    & 64.6 & 78.6 \\
    
    Prior as Condition Only
    & 98.8 & 99.0 & 97.4 & 92.8 & 97.0
    & 69.1 & 81.6 \\

    \textbf{Prototype-Biased Flow (Ours)}
    & \textbf{99.6}
    & \textbf{99.8}
    & \textbf{98.8}
    & \textbf{96.0}
    & \textbf{98.6}
    & \textbf{74.9}
    & \textbf{86.4} \\
    
    \bottomrule
    \end{tabular}%
}
\vspace{-5pt}

\end{table}

%% file: supp_arxiv.tex
\title{Appendix of UniMPA}

\maketitle

\IEEEdisplaynontitleabstractindextext

\IEEEpeerreviewmaketitle

\appendices
\setcounter{subsection}{0}
\setcounter{subsubsection}{0}
\renewcommand{\thesubsection}{\Alph{subsection}}
\renewcommand{\thesubsubsection}{\thesubsection\arabic{subsubsection}}

\subsection{Model and Training Details}
\label{app:model_training_details}
This section provides supplementary implementation details for UniMPA. We first describe the joint-space form of the Transition-Critical Trigger Gate used for bimanual tasks whose actions are represented by arm joint positions, and then summarize the model and training configurations used in our experiments.

\subsubsection{Joint-Space Trigger Gate for Bimanual Tasks}
\label{app:joint_space_gate}
\begin{figure}[t]
    \centering
    \includegraphics[width=1\columnwidth]{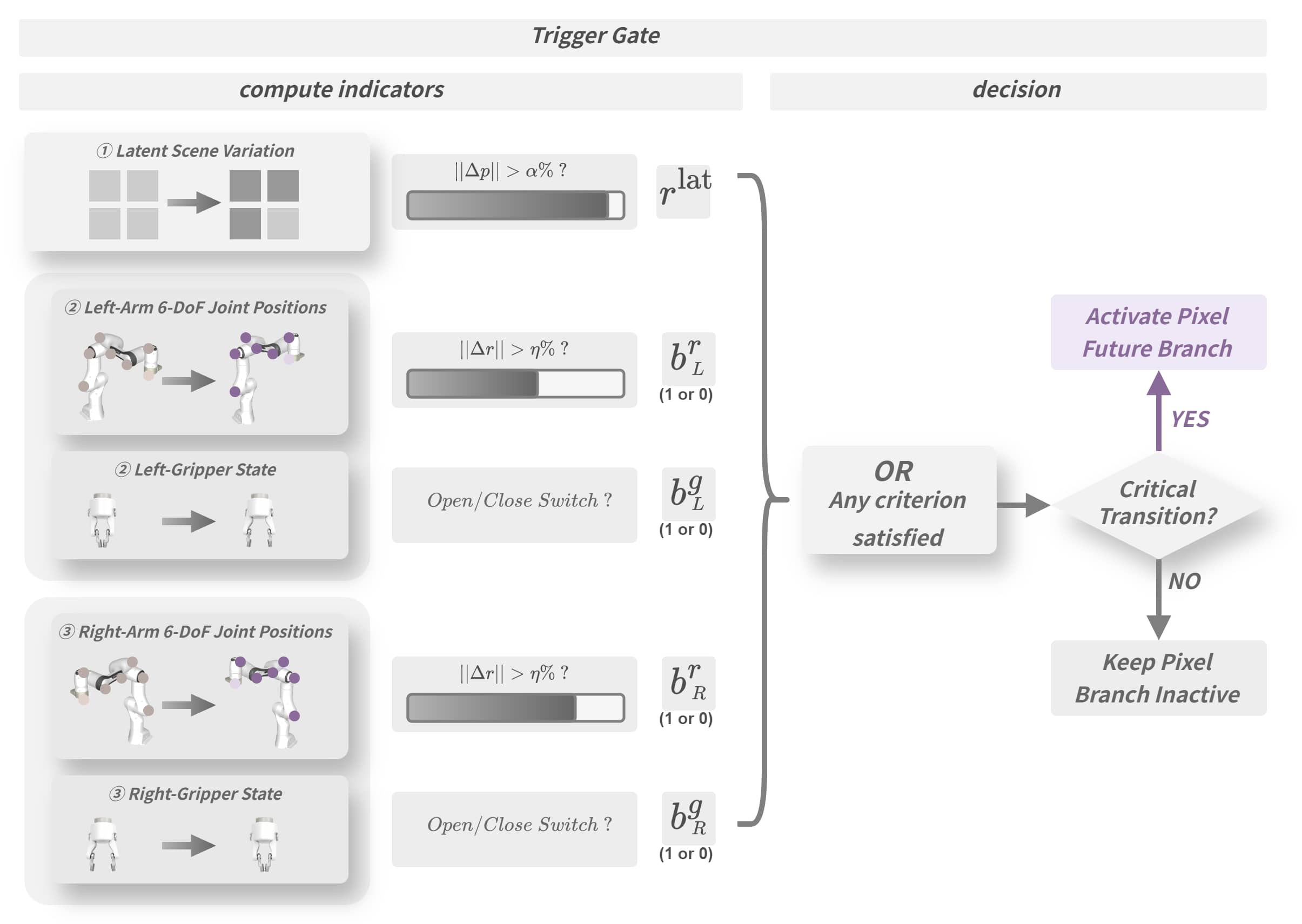}
    \vspace{-18pt}
    \caption{\textbf{Joint-space Transition-Critical Trigger Gate for bimanual control.} The gate combines latent scene variation with left- and right-arm 6-DoF joint-position changes and gripper-state switches. Pixel future prediction is activated when any criterion indicates a critical transition and otherwise remains inactive.}
    \vspace{-1pt}
\label{fig:supp_gate}
\end{figure}

For bimanual embodiments controlled in joint space, the Cartesian translation and rotation indicators used for end-effector action representations are replaced by arm-wise joint-position variation, while the latent-scene and gripper-state criteria remain unchanged. As illustrated in Fig.~\ref{fig:supp_gate}, the gate therefore combines predicted semantic change with left-arm motion, left-gripper switching, right-arm motion, and right-gripper switching.

The semantic transition score follows the main formulation and is computed from the predicted latent displacement of the main camera view:
\begin{equation}
\small
r^{\mathrm{lat}}_t
=
\frac{
\left\|
\hat{\mathbf{F}}^{\mathrm{main}}_{t+\Delta}
-
\mathbf{F}^{\mathrm{main}}_t
\right\|_2
}{
\left\|
\mathbf{F}^{\mathrm{main}}_t
\right\|_2
+
\epsilon
},
\label{eq:app_joint_latent_transition_score}
\end{equation}
where $\epsilon$ prevents numerical instability. A large $r^{\mathrm{lat}}_t$ indicates that the world expert anticipates a substantial task-relevant visual-state change.

For a dual-arm action, we use the joint-space representation
\begin{equation}
\small
\mathbf{a}_{t+k}
=
\left[
\mathbf{q}^{L}_{t+k},
 g^{L}_{t+k},
\mathbf{q}^{R}_{t+k},
 g^{R}_{t+k}
\right],
\qquad
\mathbf{q}^{L}_{t+k},\mathbf{q}^{R}_{t+k}\in\mathbb{R}^{6},
\label{eq:app_bimanual_joint_action}
\end{equation}
where $\mathbf{q}^{L}_{t+k}$ and $\mathbf{q}^{R}_{t+k}$ denote the commanded 6-DoF joint positions of the left and right arms, respectively, and $g^{L}_{t+k}$ and $g^{R}_{t+k}$ denote the corresponding gripper commands. We define the inter-step joint changes as
\begin{equation}
\small
\Delta\mathbf{q}^{s}_{t+k}
=
\mathbf{q}^{s}_{t+k}-\mathbf{q}^{s}_{t+k-1},
\qquad s\in\{L,R\}.
\label{eq:app_joint_position_change}
\end{equation}
The arm-motion and gripper-switch indicators are then
\begin{equation}
\small
\left\{
\begin{aligned}
 b^{q}_{s,t}
 &=
 \mathbf{1}\!\left[
 \max_{0\leq k<K}
 \left\|\Delta\mathbf{q}^{s}_{t+k}\right\|_2
 >\eta_q
 \right],
 \\
 b^{g}_{s,t}
 &=
 \mathbf{1}\!\left[
 \max_{0\leq k<K}
 \left|g^{s}_{t+k}-g^{s}_{t+k-1}\right|
 >\eta_g
 \right],
 \qquad s\in\{L,R\},
\end{aligned}
\right.
\label{eq:app_bimanual_joint_trigger}
\end{equation}
where $\eta_q$ and $\eta_g$ are the joint-motion and gripper-switch thresholds, respectively.

The final binary gate activates pixel-level future prediction whenever any semantic or motor criterion is satisfied:
\begin{equation}
\small
\begin{aligned}
m_t
=
\mathbf{1}\!\Big[
& r^{\mathrm{lat}}_t>\eta_{\mathrm{lat}}
 \;\lor\;
 b^{q}_{L,t}=1
 \;\lor\;
 b^{g}_{L,t}=1
\\[-1mm]
& \lor\;
 b^{q}_{R,t}=1
 \;\lor\;
 b^{g}_{R,t}=1
\Big].
\end{aligned}
\label{eq:app_bimanual_binary_trigger_gate}
\end{equation}
Thus, the joint-space gate preserves the same persistent-selective principle as the end-effector formulation while adapting the action-derived evidence to bimanual joint control. The latent score captures semantic state evolution, whereas the arm-wise joint and gripper indicators explicitly identify motor transitions associated with contact, reconfiguration, and coordinated manipulation.

\subsubsection{Hierarchical Cross-Expert Attention}
\label{app:cross_expert_attention}

UniMPA couples the vision-language backbone, World Expert, and Action Expert through a layer-wise shared attention operation rather than introducing separate cross-attention modules between experts.
The World Expert participates in both training and inference: it converts stochastic world tokens into a world-transition representation that remains visible to the Action Expert.
The latent- and pixel-prediction heads attached to the World Expert are training-only supervision modules and are disabled at inference; removing these decoding heads does not remove the World Expert or its world-token representations from the action-generation path.
Let
$\mathbf{H}^{\ell}_{B}$,
$\mathbf{H}^{\ell}_{W}$, and
$\mathbf{H}^{\ell}_{A}$
denote the hidden tokens of the backbone, World Expert, and Action Expert at Transformer layer $\ell$, respectively.
During training, the three streams are logically arranged as
\begin{equation}
\mathbf{H}^{\ell}
=
\left[
\mathbf{H}^{\ell}_{B};
\mathbf{H}^{\ell}_{W};
\mathbf{H}^{\ell}_{A}
\right],
\label{eq:app_attention_token_order}
\end{equation}
where $\mathbf{H}^{\ell}_{B}$ contains the visual-language prefix,
$\mathbf{H}^{\ell}_{W}$ contains the stochastic world tokens,
and $\mathbf{H}^{\ell}_{A}$ contains the action-chunk tokens.
In the $\pi_{0.5}$ configuration used by UniMPA, image and language tokens belong to the same attention block, while the world and action streams each form one additional block.

\begin{figure}[t]
    \centering
    \includegraphics[width=1\columnwidth]{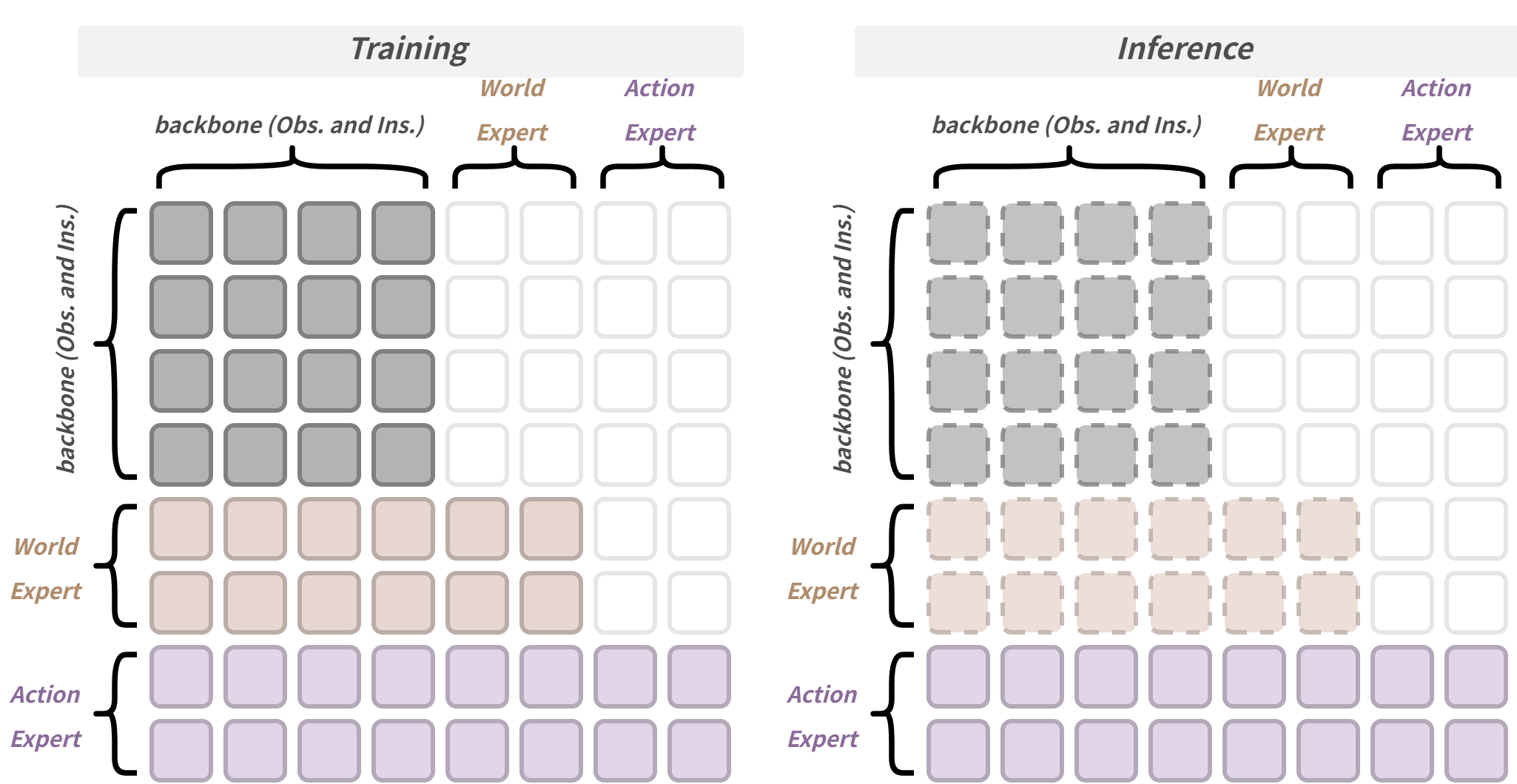}
    \vspace{-18pt}
    \caption{\textbf{Cross-expert attention in training and inference.}
    During training, the backbone, World Expert, and Action Expert participate in a shared self-attention operation with hierarchical block-wise visibility.
    During inference, the backbone and World Expert are each evaluated once per control step, and their layer-wise key/value tensors are cached; each denoising step then recomputes only the Action Expert.
    Action tokens attend jointly to the cached visual-language tokens, cached world tokens, and the current action chunk.
    The World Expert's latent- and pixel-prediction heads are disabled at inference, so no explicit future feature or image is decoded, but the world-transition representation remains active in action generation.}
    \vspace{-1pt}
\label{fig:supp_atten}
\end{figure}

\textbf{Block-wise visibility.}
The implementation constructs a one-dimensional block-boundary mask $m\in\{0,1\}^{N}$ and converts it into a two-dimensional attention mask through cumulative block indices. Specifically,
\begin{equation}
l_i=\sum_{j=1}^{i}m_j,
\qquad
M_{ij}
=
p_i p_j\,
\mathbf{1}\!\left[l_j\leq l_i\right],
\label{eq:app_block_attention_mask}
\end{equation}
where $p_i$ is the padding-validity indicator. The backbone prefix uses zeros in $m$; the world block uses $[1,0,\ldots,0]$; and the action block uses $[1,0,\ldots,0]$. Consequently, the cumulative levels are $0$, $1$, and $2$ for backbone, world, and action tokens, respectively. At the block level, the training mask is therefore
\begin{equation}
\mathbf{M}_{\mathrm{tr}}
=
\begin{bmatrix}
1 & 0 & 0 \\
1 & 1 & 0 \\
1 & 1 & 1
\end{bmatrix},
\label{eq:app_training_attention_visibility}
\end{equation}
Thus, backbone tokens attend only to the backbone,
World tokens attend to both the backbone and World block,
and Action tokens attend to all three streams.
Tokens within the same block share the same visibility level and therefore use full bidirectional attention rather than token-wise causal attention, as shown in Fig.~\ref{fig:supp_atten}.

\textbf{Layer-wise joint attention.}
Importantly, shared visibility does not imply shared Transformer parameters.
At each layer, the three streams independently apply their own normalization and query, key, and value projections,
\begin{equation}
\begin{aligned}
\mathbf{Q}^{\ell}_{r}
&=
\mathbf{W}^{\ell}_{Q,r}\widetilde{\mathbf{H}}^{\ell}_{r},
&
\mathbf{K}^{\ell}_{r}
&=
\mathbf{W}^{\ell}_{K,r}\widetilde{\mathbf{H}}^{\ell}_{r},
\\
\mathbf{V}^{\ell}_{r}
&=
\mathbf{W}^{\ell}_{V,r}\widetilde{\mathbf{H}}^{\ell}_{r},
&
r &\in\{B,W,A\}.
\end{aligned}
\label{eq:app_stream_qkv}
\end{equation}
The projected tensors are then concatenated along the token dimension,
\begin{equation}
\begin{aligned}
\mathbf{Q}^{\ell}
&=
[\mathbf{Q}^{\ell}_{B};
 \mathbf{Q}^{\ell}_{W};
 \mathbf{Q}^{\ell}_{A}],\\
\mathbf{K}^{\ell}
&=
[\mathbf{K}^{\ell}_{B};
 \mathbf{K}^{\ell}_{W};
 \mathbf{K}^{\ell}_{A}],\\
\mathbf{V}^{\ell}
&=
[\mathbf{V}^{\ell}_{B};
 \mathbf{V}^{\ell}_{W};
 \mathbf{V}^{\ell}_{A}],
\end{aligned}
\label{eq:app_concat_qkv}
\end{equation}
and participate in a single multi-head self-attention operation,
\begin{equation}
\mathbf{O}^{\ell}
=
\operatorname{Attn}
\!\left(
\mathbf{Q}^{\ell},
\mathbf{K}^{\ell},
\mathbf{V}^{\ell};
\mathbf{M}_{\mathrm{tr}}
\right).
\label{eq:app_shared_attention}
\end{equation}
The attention output is split according to the original token ranges, after which each stream applies its own output projection, residual connection, normalization, and MLP.
Consequently, cross-expert interaction occurs at every Transformer depth rather than only after a complete World Expert forward pass. The World Expert reads the current backbone representation, while the Action Expert simultaneously accesses both the current backbone and world-transition representations.

\textbf{Inference visibility and caching.}
At inference time, UniMPA preserves the same hierarchical information flow from the vision-language backbone through the World Expert to the Action Expert.
For each control step, the vision-language backbone and World Expert are evaluated once.
Their layer-wise key/value tensors are then cached and reused throughout the iterative flow-matching process.
The World Expert therefore remains part of the inference computation graph, whereas its latent- and pixel-prediction heads are not invoked because explicit future-feature or future-image decoding is unnecessary for action generation.

Denoting the cached backbone tensors as
$\mathbf{K}^{\mathrm{cache}}_{B}$ and
$\mathbf{V}^{\mathrm{cache}}_{B}$,
and the cached World-Expert tensors as
$\mathbf{K}^{\mathrm{cache}}_{W}$ and
$\mathbf{V}^{\mathrm{cache}}_{W}$,
the action attention at Euler denoising step $s$ becomes
\begin{equation}
\begin{aligned}
\mathbf{O}^{(s)}_{A}
=
\operatorname{Attn}\Big(
&\mathbf{Q}^{(s)}_{A},
\left[
\mathbf{K}^{\mathrm{cache}}_{B};
\mathbf{K}^{\mathrm{cache}}_{W};
\mathbf{K}^{(s)}_{A}
\right], \\
&\left[
\mathbf{V}^{\mathrm{cache}}_{B};
\mathbf{V}^{\mathrm{cache}}_{W};
\mathbf{V}^{(s)}_{A}
\right]
\Big).
\end{aligned}
\label{eq:app_inference_attention}
\end{equation}
Thus, at every denoising step, each action token attends to the cached visual-language context, the cached world-transition representation, and the complete current action chunk.
Only the Action Expert states must be recomputed across denoising steps; the backbone and World Expert are each evaluated once per control step and subsequently reused through their cached layer-wise representations.
This cached implementation retains the complete anticipate--ground--refine inference path while avoiding repeated backbone and World-Expert computation during iterative action refinement.

\subsubsection{Model Configuration}
\label{app:model_configuration}

Table~\ref{tab:model_configuration} summarizes the architectural and implementation configuration used throughout our experiments. 
UniMPA is instantiated on the $\pi_{0.5}$ backbone, while the World Expert follows an 18-layer Transformer architecture compatible with the action expert and operates on 16 learnable transition queries initialized to zero. 
For future prediction and memory construction, we use frozen V-JEPA~2 representations as the latent visual space, together with a lightweight Transformer decoder for pixel prediction and stateful Mamba encoders for temporal memory modeling. 
The memory space is maintained at a hidden dimension of 512, with cross-modal interaction implemented by multi-head attention and causal These architectural settings are shared across all simulation and real-world experiments; only benchmark-dependent training and data configurations are varied as described below.

\input{Supp-Tables/configuration_model}

\subsubsection{Training Configuration}
\label{app:training_configuration}

Table~\ref{tab:training_configuration} provides the complete optimization, data-processing, and benchmark-specific training settings. 
Both training stages use AdamW with $\beta_1=0.9$ and $\beta_2=0.95$. 
Stage~1 pretrains the memory components for 10 epochs across all benchmarks, using truncated backpropagation through time over 8 steps and fixed reconstruction, retrieval, and cross-modal pairing loss weights. 
During Stage~1 retrieval-simulation training, we exclude the query entry itself from the candidate set, such that $i \notin \mathcal{C}(i)$.
Stage~2 then fine-tunes the policy while keeping the pretrained memory encoders and stored memory entries frozen; the training length, learning rate, warm-up schedule, and prediction-loss weights are adjusted according to the benchmark scale and data regime. 
We additionally report the camera configuration, image preprocessing, robot embodiment, and computational budget for each benchmark to make the simulation and real-world training protocols directly reproducible.

\input{Supp-Tables/configuration_training}

\input{Supp-Tables/21tasks_suite}

\input{Supp-Tables/21tasks_stage}

\subsection{Real-World Task Organization}
\label{Experimental}
We conduct real-world evaluation on two bimanual robot platforms, AgileX Cobot Magic and GALAXEA R1 Lite, to assess UniMPA across different embodiments and manipulation regimes. The evaluation is organized along two complementary axes. First, tasks are grouped into seven suites according to their dominant physical interaction pattern: \textit{(A)} semantic rearrangement and sorting, \textit{(B)} articulated and container interaction, \textit{(C)} precision assembly and geometric manipulation, \textit{(D)} deformable and tool-mediated manipulation, \textit{(E)} bimanual coordination, \textit{(F)} dynamic and reactive manipulation, and \textit{(G)} long-horizon composition and recovery. Second, each task is annotated with orthogonal properties that describe the transition, memory, coordination, and recovery challenges it contains. This two-level organization separates \emph{what physical interaction dominates a task} from \emph{which UniMPA capabilities are stressed during its execution}.

\subsubsection{Evaluation Protocol and Metrics}
\label{app:real_metrics}

All compared methods are evaluated under the same randomized initial-state distributions, language instructions, camera configurations, observation modalities, and robot-control interfaces. Unless otherwise specified, each task is evaluated with $N=25$ independent trials. Initial object poses and robot configurations are randomized within predefined task-specific safety ranges that are fixed before evaluation and shared across methods. A trial is completed without human intervention. For dynamic tasks, the target follows the same predefined motion distribution for all methods. For recovery tasks, the prescribed external disturbance is part of the task definition; successful completion requires the policy to recover and finish the remaining checkpoints.

For task $i$ with $S_i$ ordered checkpoints, let $c_{i,j}^{(n)}\in\{0,1\}$ indicate whether checkpoint $j$ is reached in trial $n$. We first define the cumulative checkpoint indicator
\begin{equation}
\small
\tilde{c}_{i,j}^{(n)}
=
\prod_{k=1}^{j} c_{i,k}^{(n)},
\label{eq:app_progressive_indicator}
\end{equation}
so that a later checkpoint is successful only when all earlier checkpoints have been completed. The success rate is
\begin{equation}
\small
\mathrm{SR}_{i,j}
=
\frac{1}{N}
\sum_{n=1}^{N}
\tilde{c}_{i,j}^{(n)}.
\label{eq:app_checkpoint_sr}
\end{equation}

We report two complementary real-world metrics. The \textbf{Task Success Rate (TSR)} is the final-checkpoint success rate,
\begin{equation}
\small
\mathrm{TSR}_{i}
=
\mathrm{SR}_{i,S_i},
\label{eq:app_tsr}
\end{equation}
whereas the \textbf{Cumulative Success Rate (CSR)} averages the cumulative checkpoint success rates over the complete task,
\begin{equation}
\small
\mathrm{CSR}_{i}
=
\frac{1}{S_i}
\sum_{j=1}^{S_i}
\mathrm{SR}_{i,j}.
\label{eq:app_csr}
\end{equation}
CSR therefore distinguishes early failures from failures that occur near task completion. Suite-level and platform-level TSR/CSR values are obtained by averaging the corresponding task-level metrics.

\subsubsection{Suite-Level Organization Across Platforms}
\label{app:real_suite_organization}
Table~\ref{tab:real_taxonomy} gives the suite-level view of the real-world evaluation. GALAXEA R1 Lite contains three tasks from each suite, yielding 21 tasks in total, whereas AgileX Cobot Magic evaluates one representative task from each suite, yielding seven tasks. Rather than treating the seven suites as arbitrary semantic categories, Table~\ref{tab:real_taxonomy} also summarizes the primary evaluation focus associated with each interaction type, ranging from language-conditioned rearrangement and articulated-state transitions to bimanual coordination, dynamic adaptation, and long-horizon recovery.

\subsubsection{Task Definitions and Progressive Checkpoints}
\label{app:real_task_checkpoints}
Table~\ref{tab:app_galaxea_tasks_ad} instantiates the suite taxonomy at the task level. Each of the 21 tasks is assigned to its dominant interaction suite and decomposed into an ordered sequence of observable physical checkpoints. Tasks evaluated on both platforms are marked as ``Both'', while the remaining tasks are evaluated on GALAXEA R1 Lite only. The checkpoint sequence provides a consistent description of task progress across methods: a later stage is considered reached only after the preceding stages have been completed. This decomposition connects the task definitions in this section directly to the TSR and CSR metrics introduced in Sec.~\ref{app:real_metrics}, allowing final task completion and intermediate execution progress to be evaluated under the same task specification.

\providecommand{\tagTA}{\textsc{TA}}
\providecommand{\tagCT}{\textsc{CT}}
\providecommand{\tagLP}{\textsc{LP}}
\providecommand{\tagVR}{\textsc{VR}}
\providecommand{\tagAR}{\textsc{AR}}
\providecommand{\tagBI}{\textsc{BI}}
\providecommand{\tagDY}{\textsc{DY}}
\providecommand{\tagRC}{\textsc{RC}}

\input{Supp-Tables/21tasks_tag}

\subsubsection{Orthogonal Task Properties and Language Instructions}
\label{app:real_task_properties}
The suite assignment captures the dominant physical interaction, but it does not fully describe the challenges that may co-occur within a task. For example, a long-horizon task may simultaneously contain critical contact transitions, require temporally coherent retrieval, and involve recovery from an intermediate disturbance. We therefore introduce the orthogonal task properties summarized in Table~\ref{tab:real_tags}. The tags characterize transition ambiguity (\tagTA), critical transitions (\tagCT), long-horizon progress (\tagLP), visual-to-action retrieval (\tagVR), action-to-visual retrieval (\tagAR), bimanual coordination (\tagBI), dynamic interaction (\tagDY), and recovery (\tagRC). Their correspondence to UniMPA components makes explicit which aspects of the method are exercised by each task, and allows multiple properties to be associated with the same task.

As shown in Table~\ref{tab:app_instruction}, the language-instruction table that follows completes the task description by reporting the task-property tags together with the natural-language commands used during evaluation. The instruction specifies the task-level goal presented to the policy, while the ordered checkpoints in Table~\ref{tab:app_galaxea_tasks_ad} define the observable physical progression used for evaluation. All compared methods receive the same fixed instruction for a given task. The four tables in this section form a coherent hierarchy: the taxonomy table establishes suite-level coverage, the checkpoint table specifies task-level execution stages, the property table identifies cross-cutting challenges and their correspondence to UniMPA, and the instruction table defines the language-conditioning interface.

\input{Supp-Tables/21tasks_instruction}

\subsection{Supplementary RoboTwin~2.0 Results}
\label{app:robotwin_full}

The main paper reports results on 11 representative tasks from the RoboTwin~2.0 Randomized setting, while we provide an extended evaluation here to further characterize the task-level performance distribution. Each task is evaluated with 100 rollouts under the same randomized protocol, and Table~\ref{tab:robotwin_random_allresults} reports the success rate of each compared method together with the average performance for each task group. This expanded evaluation is particularly informative because RoboTwin~2.0 spans manipulation tasks with substantially different levels of difficulty. Some tasks remain challenging for nearly all methods, whereas others permit considerably higher success rates and thus more clearly expose differences in robustness and generalization.

Across the 44 evaluated tasks, UniMPA achieves an average success rate of \textbf{35.84\%}, outperforming the strongest compared baseline, $\pi_{0.5}$ (27.57\%), by \textbf{8.27 percentage points}. It also surpasses HALO (26.86\%), BagelVLA (22.59\%), and $\pi_0$ (18.30\%) by 8.98, 13.25, and 17.54 points, respectively. The gains are distributed across diverse interaction patterns rather than being concentrated in a small set of tasks. For example, UniMPA achieves 84\% on \textit{Grab Roller}, 75\% on \textit{Place Container Plate}, 67\% on \textit{Open Laptop} and \textit{Press Stapler}, and 95\%/94\% on the two bottle-shaking tasks. Meanwhile, the detailed results also retain challenging cases involving ranking, handover, and stacking, indicating that substantial room for improvement remains on difficult manipulation behaviors. These per-task results complement the representative evaluation in the main paper and provide a broader view of UniMPA's robustness under randomized bimanual manipulation.

\input{Supp-Tables/supp-robotwinv2}

\begin{figure*}[t]
\vspace{-4mm}
\centering
\includegraphics[width=1.0\textwidth]{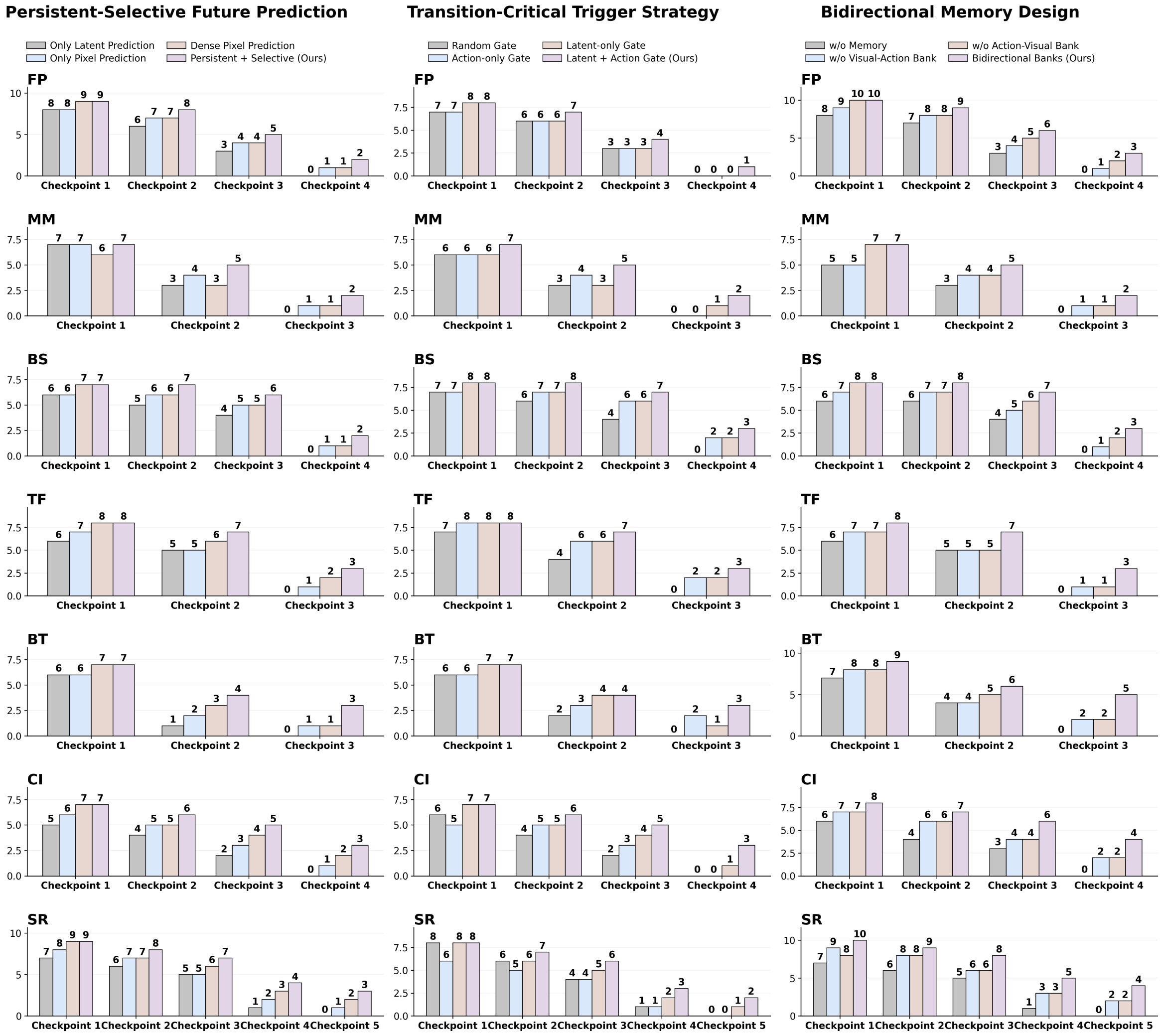} 
    \caption{\textbf{Checkpoint-level real-world ablations for future prediction, trigger strategy, and bidirectional memory design.} Results are shown on the seven AgileX tasks FP, MM, BS, TF, BT, CI, and SR. Within each task row, the minimum checkpoint count across all variants and checkpoints is shifted to zero, and each bar reports the number of additional successful trials relative to this task-specific minimum. Left: persistent-selective future prediction consistently preserves larger progression margins than latent-only, pixel-only, or dense-pixel alternatives. Middle: combining latent scene variation with action-derived transition indicators yields stronger checkpoint progression than Random, Action-only, or Latent-only gates. Right: the complete bidirectional memory maintains the strongest progression, while removing memory or either retrieval direction produces cumulative losses toward later checkpoints.}
\label{fig:supp-realworld-taskab1-3}
\end{figure*}

\begin{figure*}[t]
\vspace{-4mm}
\centering
\includegraphics[width=1.0\textwidth]{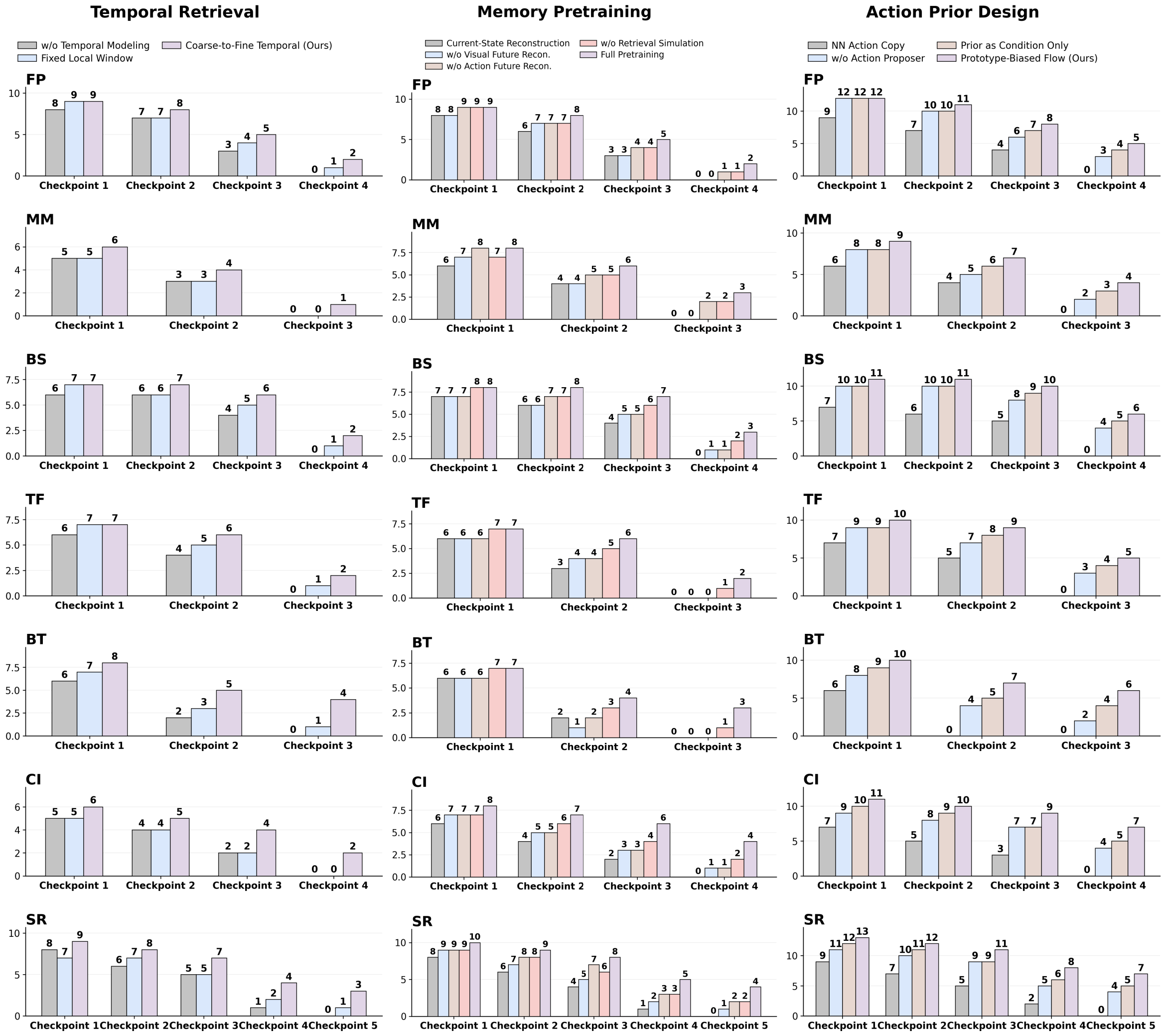} 
    \caption{\textbf{Checkpoint-level real-world ablations for temporal retrieval, memory pretraining, and action-prior design.} The visualization follows the same task-wise normalization as Fig.~\ref{fig:supp-realworld-taskab1-3}: the minimum checkpoint count in each task row is mapped to zero and all other bars show additional successful trials above that reference. Left: coarse-to-fine temporal retrieval preserves larger margins than fixed-window retrieval or removing temporal modeling. Middle: full memory pretraining benefits from future reconstruction, with the advantage becoming clearer at later checkpoints. Right: Prototype-Biased Flow substantially outperforms nearest-neighbor action copying and weaker uses of the retrieved prior, indicating that executable prototypes are most effective when they bias the flow source and are subsequently refined by the action expert.}
\label{fig:supp-realworld-taskab4-6}
\end{figure*}

\subsection{Supplementary Real-World Ablation Results}
\label{app:real_ablation}

We further provide checkpoint-level real-world ablations for the seven AgileX Cobot Magic tasks used in the main ablation study: Fruit Placement (FP), Microwave Manipulation (MM), Block Stacking (BS), Towel Folding (TF), Bimanual Transfer (BT), Conveyor Interception (CI), and Stack Recovery (SR). In addition to the aggregate TSR and CSR reported in the main paper, Figs.~\ref{fig:supp-realworld-taskab1-3} and~\ref{fig:supp-realworld-taskab4-6} visualize how different variants progress through the ordered checkpoints defined in Table~\ref{tab:app_galaxea_tasks_ad}. For each task row, we subtract the minimum checkpoint count among all compared variants from every bar in that row. The lowest observation is therefore mapped to zero and the remaining bars show the number of additional successful trials relative to this task-specific reference. This normalization removes the large scale difference caused by task difficulty and makes the relative checkpoint advantage of each design choice directly visible. The plots should therefore be read as \emph{relative progression margins}, rather than as absolute success rates.

Fig.~\ref{fig:supp-realworld-taskab1-3} summarizes the first three ablation groups. For future prediction, persistent latent prediction alone reaches 64.6\% TSR / 78.2\% CSR, pixel-only prediction reaches 68.6\% / 81.2\%, and dense pixel prediction reaches 70.3\% / 83.0\%, whereas the persistent-selective formulation achieves 74.9\% / 86.4\%. The checkpoint profiles show that the full design maintains its advantage not only at final completion but also across intermediate stages, supporting the complementary roles of persistent latent evolution and selectively activated fine-grained pixel supervision. For the trigger strategy, Random, Action-only, and Latent-only gates obtain 65.1\% / 79.1\%, 68.6\% / 81.3\%, and 69.7\% / 83.0\%, respectively, while the combined latent-action gate reaches 74.9\% / 86.4\%. The gains across multiple checkpoints indicate that semantic scene variation and motor-transition evidence provide complementary signals for locating critical transitions. The bidirectional-memory ablation shows an even larger degradation when memory is removed entirely (61.1\% / 75.7\%). Retaining only the Action-Visual or Visual-Action direction improves performance to 68.0\% / 81.4\% and 66.9\% / 80.1\%, respectively, but both remain below the complete bidirectional banks, demonstrating that predicted visual transitions and recent action evolution contribute distinct retrieval cues.

Fig.~\ref{fig:supp-realworld-taskab4-6} covers temporal retrieval, memory pretraining, and action-prior construction. Replacing coarse-to-fine retrieval with a fixed local window reduces performance from 74.9\% / 86.4\% to 68.6\% / 81.6\%, while removing temporal modeling further decreases it to 65.7\% / 79.4\%. This degradation is especially visible at later checkpoints, where phase-consistent retrieval becomes increasingly important. The pretraining variants exhibit a consistent hierarchy: current-state reconstruction obtains 62.9\% / 76.7\%, removing visual-future reconstruction gives 64.6\% / 78.5\%, and removing action-future reconstruction gives 66.9\% / 80.6\%; full memory pretraining reaches 74.9\% / 86.4\%. Finally, the action-prior ablation produces the largest gap. Nearest-neighbor action copying reaches only 52.0\% / 67.5\%, while removing the Action Proposer and using the prior only as a condition reach 64.6\% / 78.6\% and 69.1\% / 81.6\%, respectively. Prototype-Biased Flow recovers 74.9\% / 86.4\%, supporting the use of retrieved experience as a source-distribution bias that is subsequently refined rather than directly copied. Collectively, the checkpoint visualizations show that the improvements of UniMPA are accumulated throughout execution and become particularly pronounced at later, transition-sensitive stages.

\begin{figure*}[t]
\vspace{-4mm}
\centering
\includegraphics[width=1.0\textwidth]{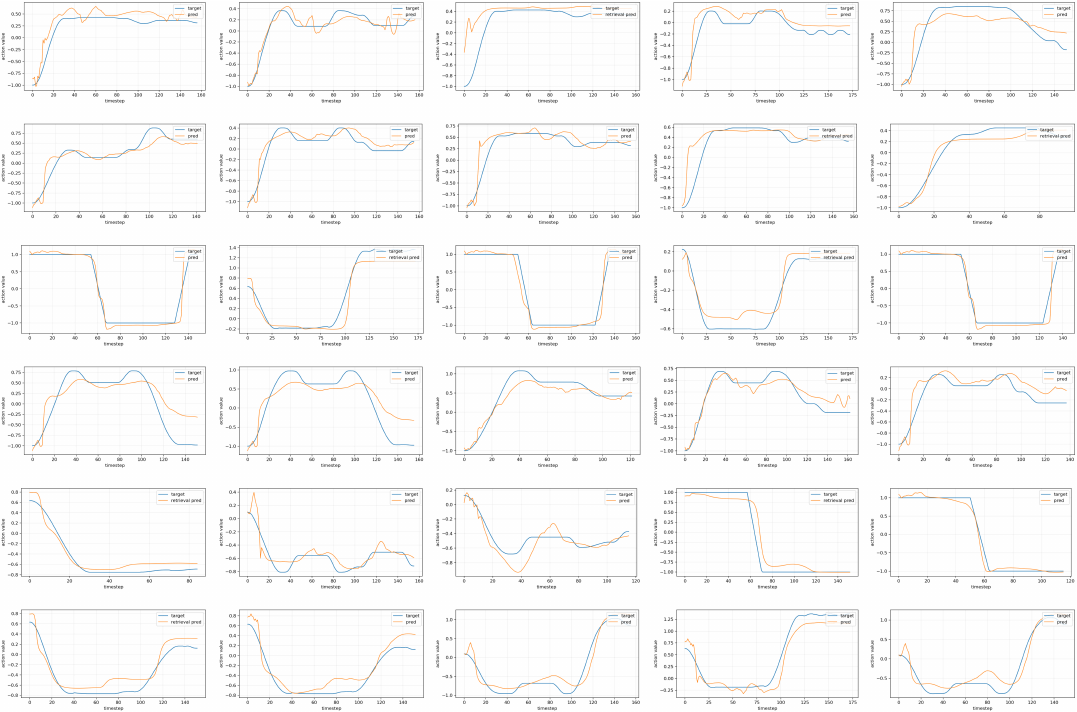} 
    \caption{\textbf{Joint-space action generation on representative real-world trajectories.} Generated actions are compared with target commands across representative joint-position and gripper dimensions. UniMPA follows the dominant temporal evolution of the target trajectories, including sustained motion trends, direction changes, and gripper-state transitions. Unlike Cartesian end-effector visualizations, these plots use the joint-space action representation adopted by the bimanual embodiments in Sec.~\ref{app:joint_space_gate}, directly illustrating temporally coherent joint-space action generation under the proposed action-manifold refinement.}
\label{fig:supp_action}
\end{figure*}

\begin{figure*}[t]
\vspace{-4mm}
\centering
\includegraphics[width=1.0\textwidth]{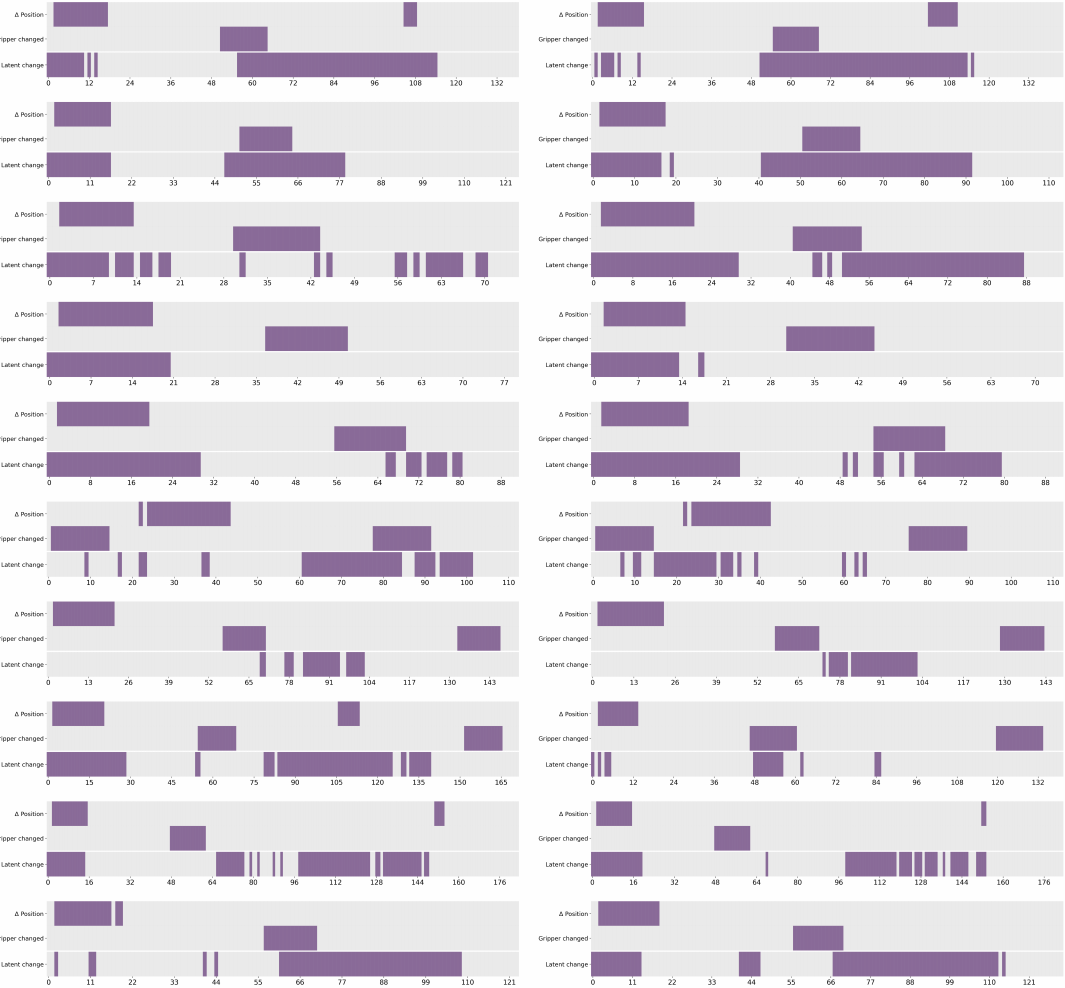} 
    \caption{\textbf{Joint-space transition-critical evidence for selective pixel prediction.} The visualization shows the temporal activation of arm joint-position variation, gripper-state switching, and latent visual change on representative bimanual trajectories. Following the Joint-Space Trigger Gate in Sec.~\ref{app:joint_space_gate}, pixel-level future supervision is activated whenever any criterion indicates a critical transition. The highlighted intervals concentrate around motion reconfiguration and interaction-sensitive periods, showing that the persistent-selective prediction principle is preserved when the action-derived trigger evidence is expressed in the joint-space control representation used by the robots.}
\label{fig:supp_vis_gate}
\end{figure*}

\begin{figure*}[t]
\vspace{-4mm}
\centering
\includegraphics[width=1.0\textwidth]{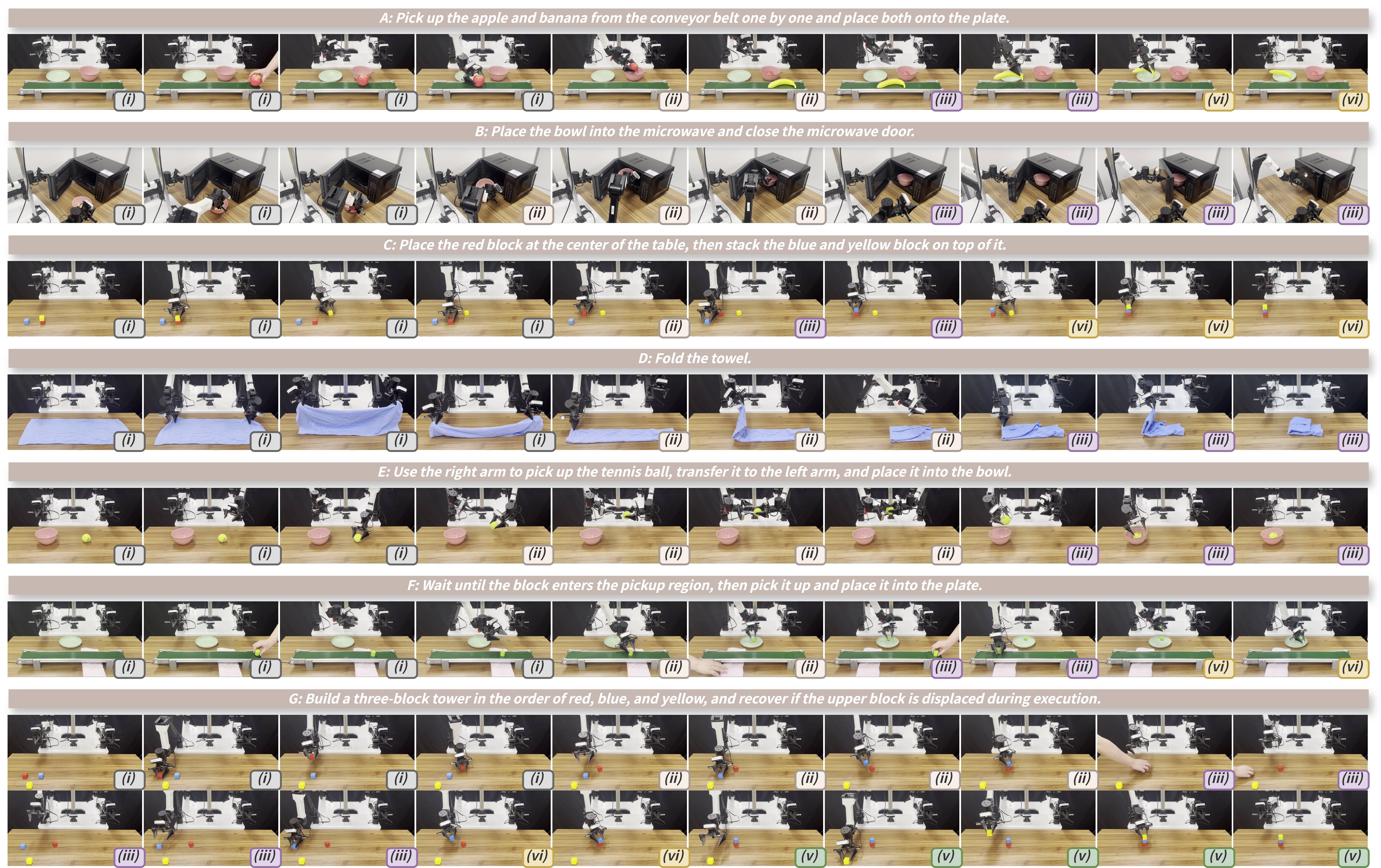} 
    \caption{\textbf{Qualitative real-world rollouts for seven representative tasks on AgileX Cobot Magic.} One task is selected from each real-world suite: Fruit Placement (A), Microwave Manipulation (B), Block Stacking (C), Towel Folding (D), Bimanual Transfer (E), Conveyor Interception (F), and Stack Recovery (G). Each row shows a temporally ordered execution, and the Roman numerals denote the ordered progressive checkpoints defined in Table~\ref{tab:app_galaxea_tasks_ad}. The trajectories provide a cross-suite view of semantic rearrangement, articulated interaction, precision manipulation, deformable-object manipulation, bimanual coordination, dynamic interception, and long-horizon recovery on AgileX Cobot Magic. The Stack Recovery rollout additionally includes the controlled displacement and subsequent recovery process.}
\label{fig:supp_realworld_aloha}
\end{figure*}

\begin{figure*}[t]
\vspace{-4mm}
\centering
\includegraphics[width=1.0\textwidth]{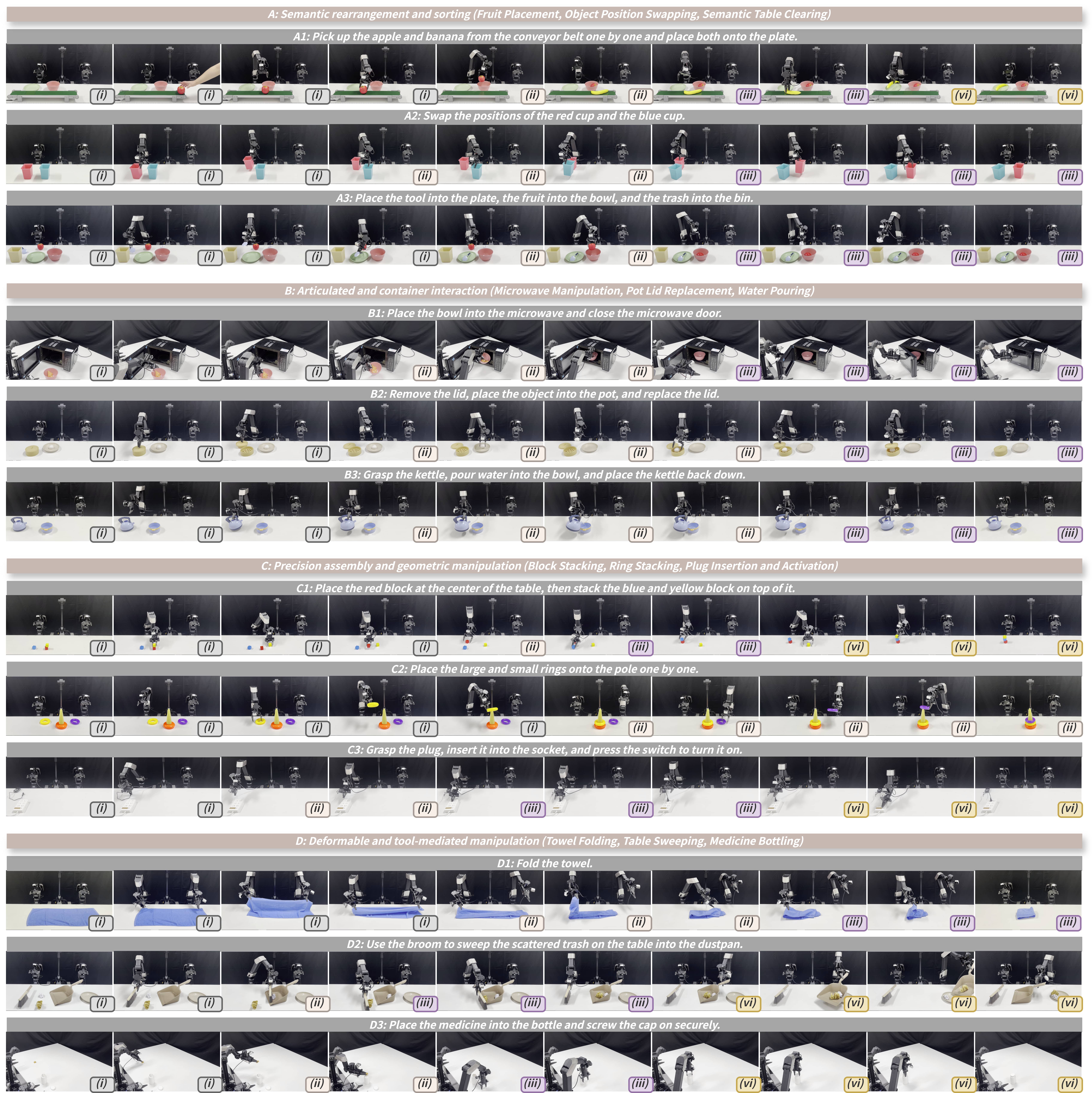} 
    \caption{\textbf{Qualitative real-world rollouts for Suites A--D on GALAXEA R1 Lite.} The figure shows representative temporally ordered executions for semantic rearrangement and sorting (A), articulated and container interaction (B), precision assembly and geometric manipulation (C), and deformable and tool-mediated manipulation (D). Each row corresponds to one task, and the Roman numerals mark the ordered progressive checkpoints defined in Table~\ref{tab:app_galaxea_tasks_ad}. The sequences visualize how UniMPA maintains task phase across multi-step manipulation while resolving interaction-critical transitions such as grasping, articulated-state changes, stacking, insertion, pouring, folding, and tool-mediated contact.}
\label{fig:supp_realworld_ad}
\end{figure*}

\begin{figure*}[t]
\vspace{-4mm}
\centering
\includegraphics[width=1.0\textwidth]{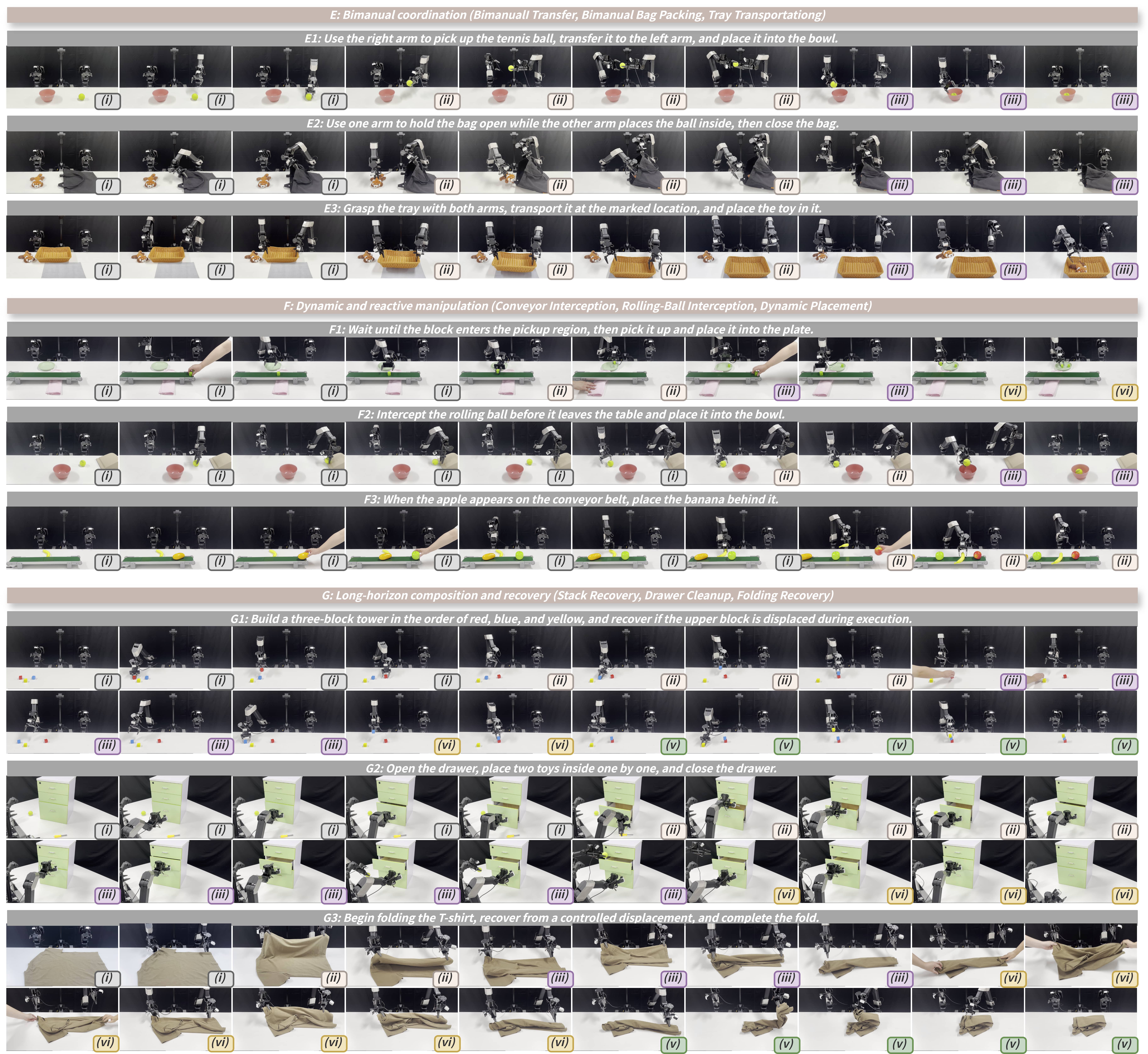} 
    \caption{\textbf{Qualitative real-world rollouts for Suites E--G on GALAXEA R1 Lite.} Representative executions are shown for bimanual coordination (E), dynamic and reactive manipulation (F), and long-horizon composition and recovery (G). Roman numerals indicate the ordered progressive checkpoints used for evaluation. The trajectories highlight inter-arm coordination and handover, adaptation to moving or newly appearing objects, and persistent execution across long-horizon stages. Recovery sequences additionally include the prescribed controlled disturbance and the subsequent return to task progress, illustrating continued execution from the perturbed state rather than task restart.}
\label{fig:supp_realworld_eg}
\end{figure*}

\begin{figure*}[t]
\vspace{-4mm}
\centering
\includegraphics[width=1.0\textwidth]{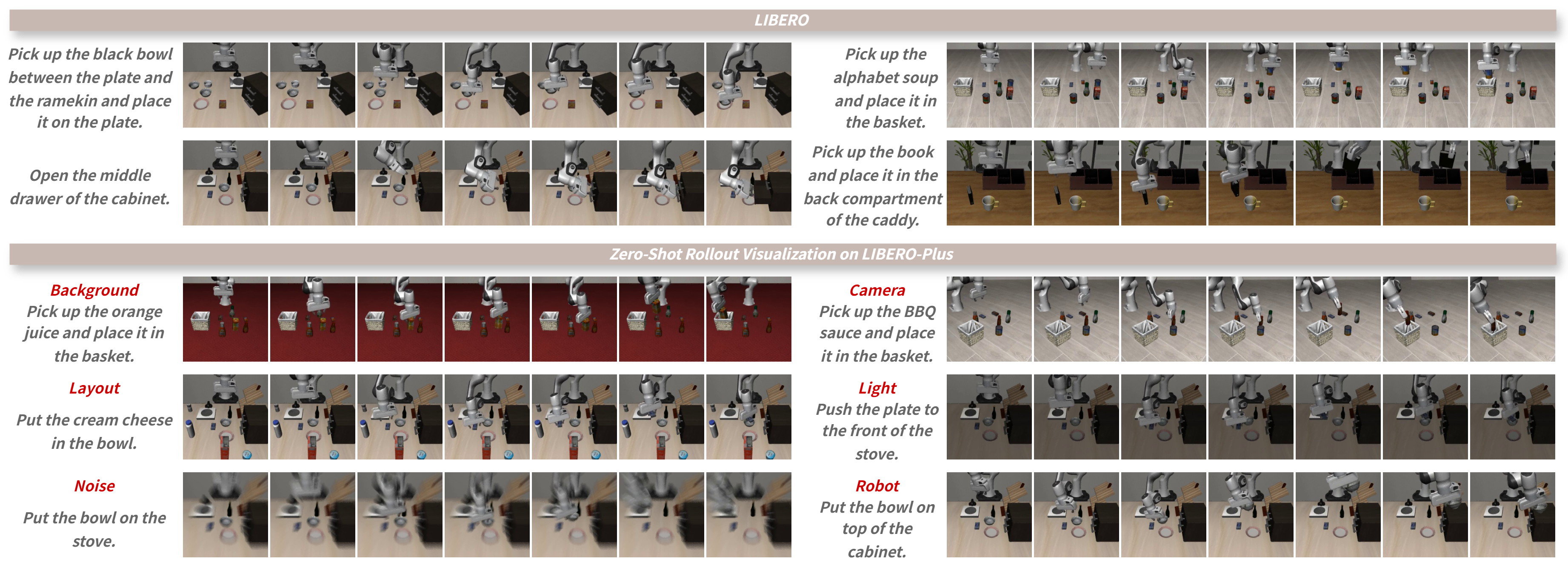} 
\caption{
\textbf{Qualitative rollouts on LIBERO and LIBERO-Plus.}
The upper panel shows representative LIBERO executions, while the lower panel visualizes zero-shot rollouts on LIBERO-Plus under background, camera, layout, illumination, noise, and robot shifts.
UniMPA maintains coherent task execution across these distribution shifts.
}
\label{fig:supp_libero_rollout}
\end{figure*}

\begin{figure*}[t]
\vspace{-4mm}
\centering
\includegraphics[width=1.0\textwidth]{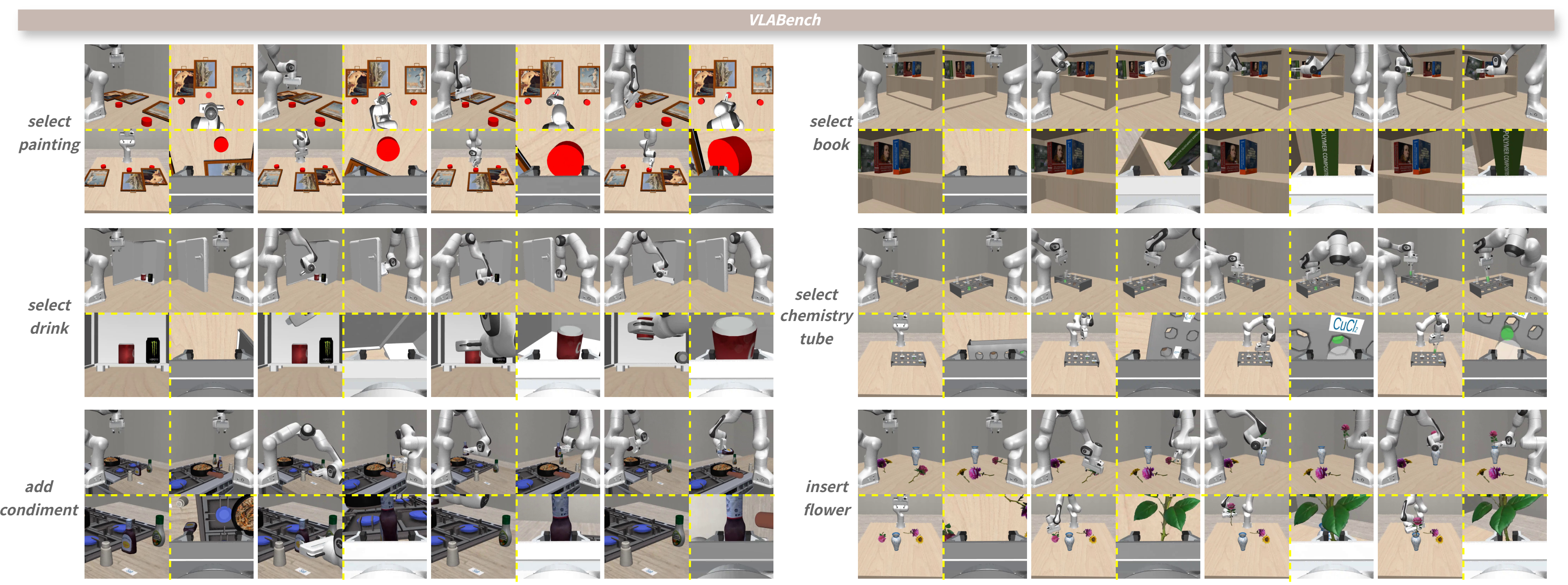} 
\caption{
\textbf{Qualitative rollouts on VLABench.}
Representative executions are shown for Select Painting, Select Book, Select Drink, Select Chemistry Tube, Add Condiment, and Insert Flower, covering diverse semantic, spatial, and multi-stage manipulation tasks.
}
\label{fig:supp_vlabench_rollout}
\end{figure*}

\begin{figure*}[t]
\vspace{-4mm}
\centering
\includegraphics[width=1.0\textwidth]{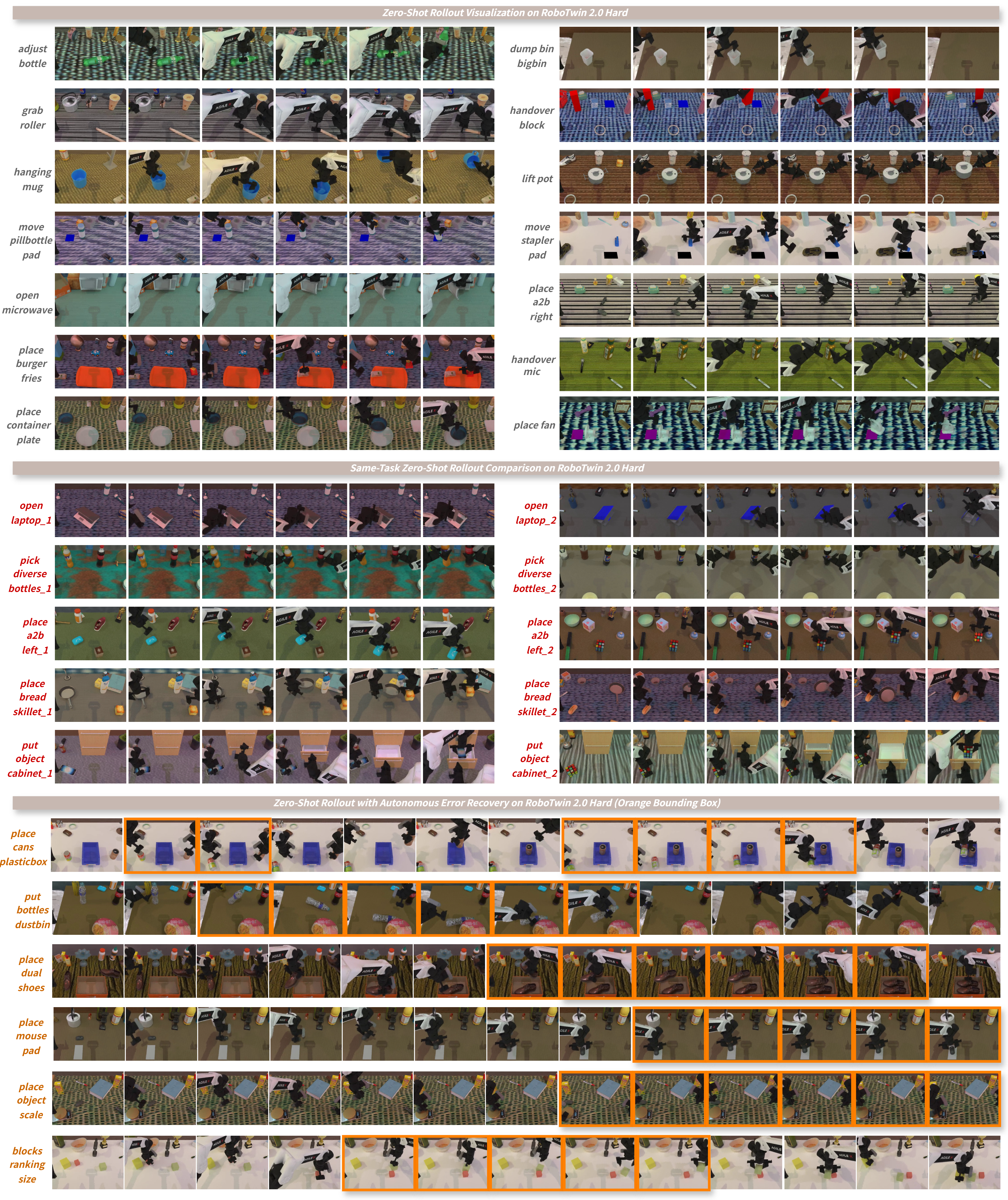} 
\caption{
\textbf{Zero-shot rollouts on RoboTwin~2.0 Hard.}
The upper panel shows representative hard tasks, the middle panel compares different instances of the same tasks, and the lower panel highlights autonomous error recovery with orange bounding boxes.
UniMPA maintains robust execution and recovers from intermediate deviations without restarting.
}
\label{fig:supp_robotwin_hard_rollout}
\end{figure*}

\subsection{Joint-Space Action and Trigger Visualization}
\label{app:joint_space_action_trigger_visualization}

To complement the joint-space formulation in Sec.~\ref{app:joint_space_gate}, we provide two additional qualitative diagnostics in Figs.~\ref{fig:supp_action} and~\ref{fig:supp_vis_gate}. These visualizations are included specifically to match the action representation used by the bimanual embodiments in the experiments, where UniMPA generates joint-position commands together with gripper commands rather than Cartesian end-effector translation and rotation. The main paper introduces action generation and transition-critical triggering in a representation-agnostic manner and uses Cartesian motion cues to explain the underlying mechanism. Here, we instantiate the same analysis directly in joint space so that the supplementary visual evidence is consistent with the actual control interface used by AgileX Cobot Magic and GALAXEA R1 Lite. Fig.~\ref{fig:supp_action} examines the temporal behavior of the generated joint-space actions, while Fig.~\ref{fig:supp_vis_gate} visualizes the transition-critical evidence derived from joint-position variation, gripper-state switching, and latent visual change. Together, they connect joint-space action generation to the Joint-Space Trigger Gate defined in Sec.~\ref{app:joint_space_gate}.

Fig.~\ref{fig:supp_action} compares representative generated action trajectories with their target commands across joint-position and gripper dimensions. The predicted trajectories follow the dominant temporal evolution of the targets, including sustained motion trends, direction changes, and discrete gripper-state transitions. This visualization complements the action-generation analysis in the main paper by showing that the Action Expert produces temporally coherent commands in the joint-space representation actually executed by the bimanual robots, rather than relying on a Cartesian end-effector abstraction.

Fig.~\ref{fig:supp_vis_gate} further visualizes the temporal evidence used to determine when fine-grained pixel-level future supervision is necessary. For representative bimanual trajectories, the activation intervals indicate substantial arm joint-position variation, gripper-state switching, or latent visual change. As defined in Eq.~\eqref{eq:app_bimanual_binary_trigger_gate}, pixel future prediction is activated whenever any of these semantic or motor criteria is satisfied. The resulting activations concentrate around motion reconfiguration and interaction-sensitive periods, illustrating how the persistent-selective prediction principle from the main paper is transferred to embodiment-consistent joint-space cues.

\subsection{Qualitative Rollout Visualization}
\label{app:rollout_visualization}

\subsubsection{Real-World Rollout Visualization}
\label{app:real_rollout_visualization}

Figs.~\ref{fig:supp_realworld_aloha}--\ref{fig:supp_realworld_eg} provide qualitative real-world rollout visualizations on both bimanual platforms. Fig.~\ref{fig:supp_realworld_aloha} presents seven representative tasks on AgileX Cobot Magic, selecting one task from each of the seven suites introduced in Sec.~\ref{Experimental}. Figs.~\ref{fig:supp_realworld_ad} and~\ref{fig:supp_realworld_eg} then provide the complete 21-task visualization on GALAXEA R1 Lite, with three tasks from each suite. In all figures, each row presents temporally ordered observations from a representative execution, and the Roman numerals overlaid on the frames correspond to the ordered progressive checkpoints defined in Table~\ref{tab:app_galaxea_tasks_ad}. The visualized stages therefore follow the same task progression used by the TSR/CSR evaluation protocol rather than being independently selected qualitative keyframes.

Fig.~\ref{fig:supp_realworld_aloha} provides a compact cross-suite view on AgileX Cobot Magic. The seven rows correspond to Fruit Placement (Suite A), Microwave Manipulation (Suite B), Block Stacking (Suite C), Towel Folding (Suite D), Bimanual Transfer (Suite E), Conveyor Interception (Suite F), and Stack Recovery (Suite G). This one-task-per-suite selection matches the seven-task AgileX evaluation used in the main paper and covers semantic rearrangement, articulated interaction, precision manipulation, deformable-object manipulation, bimanual coordination, dynamic interception, and long-horizon recovery within a single platform. In particular, the Stack Recovery sequence includes the controlled intermediate displacement and subsequent continuation of the task, making the recovery behavior visible.

Fig.~\ref{fig:supp_realworld_ad} covers Suites A--D. The trajectories show semantically conditioned object transfer in Suite A, articulated and container-state changes in Suite B, contact-sensitive alignment and stacking in Suite C, and continuous geometric evolution during deformable or tool-mediated manipulation in Suite D. Across these tasks, the ordered frames make visible the transition structure that motivates persistent future prediction and transition-aligned memory: visually similar states may occur at different phases, while interaction-critical events such as grasping, insertion, pouring, folding, and tool contact produce localized state changes that must be resolved precisely.

Fig.~\ref{fig:supp_realworld_eg} covers Suites E--G and emphasizes coordination, online adaptation, and recovery. Suite E visualizes dual-arm role coordination and handover, Suite F contains moving-object and dynamically conditioned interactions, and Suite G contains longer compositions in which task progress must be maintained across multiple dependent stages. In the recovery examples, the rollout explicitly includes the prescribed intermediate disturbance followed by continued execution, illustrating that successful behavior requires resuming from the altered state rather than restarting the task. These visualizations complement the quantitative metrics by exposing the physical transitions and multi-stage dependencies represented by the progressive checkpoints.

\subsubsection{Simulation Benchmark Rollout Visualization}
\label{app:simulation_rollout_visualization}

Beyond the real-world evaluation, Figs.~\ref{fig:supp_libero_rollout}--\ref{fig:supp_robotwin_hard_rollout} extend the qualitative analysis to the simulation benchmarks.
The visualizations cover standard and distribution-shifted LIBERO tasks, diverse language-conditioned manipulation scenarios in VLABench, and challenging zero-shot execution on RoboTwin~2.0 Hard.
Together, they provide complementary qualitative evidence of UniMPA's robustness to distribution shifts, generalization across manipulation categories, and ability to maintain executable behavior under unseen and challenging conditions.

For LIBERO, Fig.~\ref{fig:supp_libero_rollout} juxtaposes representative standard rollouts with zero-shot execution under LIBERO-Plus distribution shifts.
The upper panel contains standard LIBERO executions involving object relocation, articulated manipulation, and semantic object-receptacle interaction, whereas the lower panel introduces background, camera, layout, illumination, observation-noise, and robot shifts.
Despite these perturbations, the rollout sequences remain task-directed and temporally coherent, illustrating that UniMPA can preserve manipulation behavior when visual appearance, scene geometry, observation quality, or embodiment differs from the training distribution.

Qualitative results on VLABench are summarized in Fig.~\ref{fig:supp_vlabench_rollout} through six representative tasks: Select Painting, Select Book, Select Drink, Select Chemistry Tube, Add Condiment, and Insert Flower.
These tasks cover semantic target identification, object transport, spatial alignment, and multi-stage manipulation across substantially different scenes and object categories.
The temporally ordered trajectories demonstrate that UniMPA consistently converts language-conditioned task goals into coherent action sequences across diverse manipulation settings.

The most challenging zero-shot cases are further examined on RoboTwin~2.0 Hard in Fig.~\ref{fig:supp_robotwin_hard_rollout}.
Representative successful rollouts across diverse hard tasks are shown in the upper panel, while the middle panel compares paired instances of the same tasks and illustrates consistent execution under changes in scene configuration and initialization.
The lower panel focuses on autonomous error recovery, where the orange bounding boxes highlight intermediate deviations and subsequent corrective execution.
Rather than restarting the rollout after an error, the policy adjusts its behavior from the resulting state and continues toward task completion.
These examples qualitatively complement the quantitative RoboTwin~2.0 results by demonstrating zero-shot task transfer, instance-level robustness, and autonomous recovery under challenging bimanual manipulation conditions.

\subsection{Limitations and Future Work} \label{app:limitations_future}

\textbf{\textcolor[HTML]{9673A6}{Asynchronous memory construction.}}
UniMPA formulation supports memory expansion asynchronously across rollouts.
Once an execution is completed, the newly collected observations and actions can be passed through the pretrained Visual--Action and Action--Visual memory-construction pipelines to generate additional memory entries, which are then appended to the corresponding banks and become available to subsequent executions without retraining either the policy or the memory encoders.
However, experience collected during the current rollout can only benefit future rollouts, rather than the remaining decisions within the ongoing execution.

\textbf{\textcolor[HTML]{D79B00}{Future Work 1: Synchronous online memory construction.}}
A natural extension is to eliminate this temporal separation by constructing and updating memory synchronously during execution.
Specifically, UniMPA could incrementally process newly observed states and executed actions through the same pretrained memory-construction pipelines, dynamically inserting the resulting entries into the Visual--Action and Action--Visual banks as the rollout proceeds.
These newly acquired memories could then be retrieved immediately at subsequent timesteps of the same execution, allowing the policy to exploit task-specific interaction history, intermediate corrections, and newly encountered state transitions online.
Such a formulation would extend UniMPA from cross-rollout experience accumulation toward within-rollout continual memory adaptation while preserving the existing bidirectional memory architecture.

\textbf{\textcolor[HTML]{9673A6}{Fixed temporal prediction scale.}} 
UniMPA currently adopts a predefined future offset and action horizon for latent prediction, pixel prediction, memory retrieval, and action generation. This shared temporal scale provides a simple and consistent interface across the World Expert, memory banks, and Action Expert, but different manipulation phases can exhibit substantially different temporal structure: free-space motion may evolve slowly, whereas contact, insertion, grasp switching, or recovery may require much finer temporal resolution. 

\textbf{\textcolor[HTML]{D79B00}{Future Work 2: Adaptive multi-horizon prediction.}} 
Future versions of UniMPA can extend the World Expert from a single prediction scale to a hierarchy of short-, medium-, and long-horizon future representations. The predicted transition magnitude, memory-retrieval confidence, and task phase could jointly determine which horizon is emphasized at each timestep. Short-horizon predictions would resolve contact-sensitive transitions, while longer-horizon predictions would preserve task-level progress and guide retrieval over extended temporal dependencies. This would generalize the current persistent-selective prediction principle from selective prediction modalities to selective temporal scales.

\textbf{\textcolor[HTML]{9673A6}{Binary transition-critical triggering.}} 
The current Trigger Gate makes a binary decision from latent scene variation and action-derived motion cues, activating pixel-level future prediction when any transition criterion exceeds its corresponding threshold. This formulation is lightweight and directly interpretable, and is particularly suitable for distinguishing redundant states from interaction-critical transitions. At the same time, manipulation events naturally vary in their degree and type of uncertainty rather than forming only two discrete categories. 

\textbf{\textcolor[HTML]{D79B00}{Future Work 3: Uncertainty-aware adaptive computation.}} 
UniMPA can therefore be extended from binary triggering to uncertainty-aware computation allocation. Instead of only deciding whether pixel prediction is activated, the gate could estimate transition uncertainty and dynamically determine the prediction resolution, number of pixel tokens, decoding depth, or temporal horizon allocated to the current state. Different trigger patterns could additionally distinguish contact, gripper switching, articulated motion, deformation, and unexpected scene changes. In this way, the existing Transition-Critical Trigger Gate would become a general mechanism for allocating world-model computation according to the predicted difficulty of the upcoming physical transition.

\textbf{\textcolor[HTML]{9673A6}{Single-prototype action-manifold guidance.}} 
The current Prototype-Biased Flow formulation uses retrieved executable experience to construct an informative action proposal that biases the source distribution before refinement by the Action Expert. This formulation avoids direct action copying and preserves the generative flexibility of flow matching. Nevertheless, manipulation trajectories can be genuinely multimodal: similar visual transitions may admit different grasp directions, arm assignments, approach trajectories, or recovery strategies. 

\textbf{\textcolor[HTML]{D79B00}{Future Work 4: Multimodal and phase-conditioned action priors.}} 
The action-manifold component can be extended to maintain multiple retrieved prototypes and represent them as a mixture of executable action hypotheses rather than collapsing retrieval into a single proposal. Prototype weights could be conditioned on predicted future states, temporal phase, retrieval similarity, and action-history consistency, after which flow matching could refine the selected or softly combined hypotheses. This extension would preserve the core idea of Prototype-Biased Flow while allowing UniMPA to represent multiple physically valid modes and to switch between them when the scene evolution makes one mode preferable.

%% file: Supp-Tables/configuration_model.tex
\begin{table*}[t]
\centering
\caption{\textbf{Model Configuration of UniMPA.}}
\label{tab:model_configuration}
\footnotesize
\setlength{\tabcolsep}{6pt}
\renewcommand{\arraystretch}{1.10}

\begin{tabularx}{\textwidth}{p{4.0cm} X}
\toprule
\textbf{Component} & \textbf{Configuration} \\
\midrule

Backbone
& $\pi_{0.5}$ with a PaliGemma-2B vision-language backbone and a 311M Gemma action expert \\

World Expert
& 18 Transformer layers, hidden dimension 1024, 8 attention heads, FFN dimension 4096 \\

World Tokens
& 16 learnable transition queries initialized to zero \\

Latent Prediction
& Frozen V-JEPA~2 features; 16 tokens per view with feature dimension 1408 \\

Pixel Prediction
& 64 pixel tokens; 4-layer, 8-head Transformer decoder; $32\times32$ patches; $256\times256$ output \\

Memory Representation
& Hidden dimension 512; V-JEPA~2 visual keys projected from 1408 to 512 dimensions \\

Temporal Memory Encoder
& 2-layer stateful Mamba with state dimension 16; $32\times32$ visual patches and 64 temporal tokens per view \\

Cross-Modal Fusion
& 8-head cross-attention \\

Memory Retrieval
& Cosine similarity with temperature $\tau=0.1$; maximum causal history of 32 entries \\

Trigger Gate
& Activated when latent-change score exceeds $0.2-0.5$ or a translation, rotation, or gripper-state change is detected \\

\bottomrule
\end{tabularx}
\end{table*}

%% file: Supp-Tables/configuration_training.tex
\begin{table*}[t]
\centering
\caption{\textbf{Training Configuration of UniMPA.}}
\label{tab:training_configuration}
\footnotesize
\setlength{\tabcolsep}{5.0pt}
\renewcommand{\arraystretch}{1.10}

\begin{tabular}{lcccc}
\toprule
\textbf{Setting}
& \textbf{LIBERO}
& \textbf{RoboTwin 2.0}
& \textbf{VLABench}
& \textbf{Real World} \\
\midrule

\multicolumn{5}{l}{\textit{\textbf{Stage 1: Memory Pretraining}}} \\
\midrule

Epochs
& 10 & 10 & 10 & 10 \\

Batch size
& 16 & 8 & 8 & 8 \\

Optimizer
& \multicolumn{4}{c}{AdamW, $\beta_1=0.9$, $\beta_2=0.95$} \\

Learning rate
& \multicolumn{4}{c}{$1\times10^{-4}$} \\

Truncated BPTT
& \multicolumn{4}{c}{8 steps} \\

Loss weights
& \multicolumn{4}{c}{$\lambda_{\mathrm{rec}}=1.0,\;
\lambda_{\mathrm{ret}}=1.0,\;
\lambda_{\mathrm{pair}}=0.5$} \\

\midrule
\multicolumn{5}{l}{\textit{\textbf{Stage 2: Policy Fine-Tuning}}} \\
\midrule

Training steps
& 30k & 30k & 30k & 20k \\

Batch size
& 16 & 8 & 8 & 8 \\

Optimizer
& \multicolumn{4}{c}{AdamW, $\beta_1=0.9$, $\beta_2=0.95$} \\

Learning rate
& $5\times10^{-5}$ & $5\times10^{-5}$ & $5\times10^{-6}$ & $5\times10^{-5}$ \\

Warm-up steps
& 10k & 10k & 10k & 5k \\

Frozen modules
& \multicolumn{4}{c}{Memory encoders and stored memory entries} \\

$\lambda_{\mathrm{lat}}$
& 1.0 & 0.01 & 0.01 & 0.01 \\

$\lambda_{\mathrm{pix}}$
& 1.0 & 0.01 & 0.01 & 0.01 \\

Training protocol
& Benchmark-specific
& Benchmark-specific
& Benchmark-specific
& Task-suite-specific \\

GPUs
& 4 & 8 & 8 & 16 \\

Training budget
& 25\% & 50\% & 50\% & -- \\

$\eta_{\mathrm{lat}}$
& 0.20 & 0.45 & 0.20 & 0.50 \\

$\eta_p$, $\eta_r$
& \multicolumn{4}{c}{0.8} \\

\midrule
\multicolumn{5}{l}{\textit{\textbf{Data and Preprocessing}}} \\
\midrule

Input cameras
& 2 & 3 & 3 & 3 \\

Input resolution
& \multicolumn{4}{c}{$224\times224$} \\

Visual augmentation
& \multicolumn{4}{c}{
0.95 random crop, $\pm5^{\circ}$ rotation, and color jitter} \\

Number of Arms
& 1 & 2 & 1 & 2 \\

Robotic Arm Type
& Franka Emika Panda & Aloha-AgileX & Franka Emika Panda & Aloha-AgileX \& GALAXEA R1 Lite  \\

\bottomrule
\end{tabular}
\end{table*}

%% file: Supp-Tables/21tasks_suite.tex
\begin{table*}[t]
\caption{
\textbf{Taxonomy of the 21 real-world manipulation tasks.}
Tasks are organized into seven suites according to their dominant physical interaction patterns.
}
\label{tab:real_taxonomy}
\centering
\footnotesize
\setlength{\tabcolsep}{4pt}
\renewcommand{\arraystretch}{1.12}

\begin{tabularx}{\textwidth}{
C{0.45cm}
p{5.5cm}
C{1.5cm}
C{1.5cm}
X
}
\toprule
\textbf{Suite}
& \textbf{Interaction Type}
& \multicolumn{2}{c}{\textbf{Task$\times$Episode}}
& \textbf{Primary Evaluation Focus} \\
\cmidrule(lr){3-4}
&
&
\textbf{GALAXEA}
&
\textbf{AgileX}
& \\
\midrule

A
& Semantic rearrangement and sorting
& $3\times50$
& $1\times50$
& Language-conditioned target transitions, object--receptacle correspondence, and transferable pick--transport--release patterns. \\

B
& Articulated and container interaction
& $3\times50$
& $1\times50$
& Phase ambiguity, articulated-state transitions, container interaction, and temporally coherent memory retrieval. \\

C
& Precision assembly and geometric manipulation
& $3\times50$
& $1\times50$
& Fine-grained alignment, insertion, stacking, contact-sensitive prediction, and action-prototype refinement. \\

D
& Deformable and tool-mediated manipulation
& $3\times75$
& $1\times75$
& Continuous state evolution, deformable-object manipulation, tool-mediated contact, and selective pixel-level transition modeling. \\

E
& Bimanual coordination
& $3\times75$
& $1\times75$
& Inter-arm coordination, handover, stabilization, coordinated motion, and visually grounded action-history retrieval. \\

F
& Dynamic and reactive manipulation
& $3\times100$
& $1\times100$
& Dynamic target interaction, interception, online adaptation, and action switching under changing scene states. \\

G
& Long-horizon composition and recovery
& $3\times100$
& $1\times100$
& Persistent task-progress modeling, phase-consistent retrieval, recovery from controlled disturbances, and long-horizon execution. \\

\bottomrule
\end{tabularx}
\end{table*}

%% file: Supp-Tables/21tasks_stage.tex
\begin{table*}[t]
\centering
\caption{\textbf{Real-world task definitions and progressive checkpoints for Suites A--G.} ``Both'' indicates that the corresponding task is evaluated on both AgileX Cobot Magic and GALAXEA R1 Lite; ``GALAXEA'' indicates GALAXEA R1 Lite only.}
\label{tab:app_galaxea_tasks_ad}
\scriptsize
\setlength{\tabcolsep}{4.0pt}
\renewcommand{\arraystretch}{1.10}
\begin{tabularx}{\textwidth}{C{0.45cm} p{3.05cm} C{1.3cm} X}
\toprule
\textbf{Suite} & \textbf{Task} & \textbf{Platform} & \textbf{Ordered Progressive Checkpoints} \\
\midrule

\multirow{3}{*}{A}
& Fruit Placement 
& Both 
& \textbf{\textit{(i)}} Grasp Apple 
$\rightarrow$ \textbf{\textit{(ii)}} (Apple $\rightarrow$ Plate) 
$\rightarrow$ \textbf{\textit{(iii)}} Grasp Banana 
$\rightarrow$ \textbf{\textit{(iv)}} (Banana $\rightarrow$ Plate) \\

& Object Position Swapping 
& GALAXEA 
& \textbf{\textit{(i)}} (Cup $\rightarrow$ Temporary) 
$\rightarrow$ \textbf{\textit{(ii)}} (Can $\rightarrow$ Cup Position) 
$\rightarrow$ \textbf{\textit{(iii)}} (Cup $\rightarrow$ Can Position) \\

& Semantic Table Clearing 
& GALAXEA 
& \textbf{\textit{(i)}} (Tool $\rightarrow$ Plate)
$\rightarrow$ \textbf{\textit{(ii)}} (Fruit $\rightarrow$ Bowl)
$\rightarrow$ \textbf{\textit{(iii)}} (Trash $\rightarrow$ Bin) \\

\midrule
\multirow{3}{*}{B}
& Microwave Manipulation 
& Both 
& \textbf{\textit{(i)}} Grasp Bowl 
$\rightarrow$ \textbf{\textit{(ii)}} Place Bowl Inside 
$\rightarrow$ \textbf{\textit{(iii)}} Close Door \\

& Pot Lid Replacement 
& GALAXEA 
& \textbf{\textit{(i)}} Remove Lid 
$\rightarrow$ \textbf{\textit{(ii)}} Place Object 
$\rightarrow$ \textbf{\textit{(iii)}} Replace Lid \\

& Water Pouring 
& GALAXEA 
& \textbf{\textit{(i)}} Grasp Kettle 
$\rightarrow$ \textbf{\textit{(ii)}} Pour Water 
$\rightarrow$ \textbf{\textit{(iii)}} Place Kettle \\

\midrule
\multirow{3}{*}{C}
& Block Stacking 
& Both 
& \textbf{\textit{(i)}} Grasp Red Block 
$\rightarrow$ \textbf{\textit{(ii)}} Place at Center 
$\rightarrow$ \textbf{\textit{(iii)}} Place Blue Block 
$\rightarrow$ \textbf{\textit{(iv)}} Place Yellow Block \\

& Ring Stacking 
& GALAXEA 
& \textbf{\textit{(i)}} (Large Ring $\rightarrow$ Pole) 
$\rightarrow$ \textbf{\textit{(ii)}} (Small Ring $\rightarrow$ Pole) \\

& Plug Insertion and Activation 
& GALAXEA 
& \textbf{\textit{(i)}} Grasp Plug 
$\rightarrow$ \textbf{\textit{(ii)}} Align with Socket 
$\rightarrow$ \textbf{\textit{(iii)}} Insert Plug 
$\rightarrow$ \textbf{\textit{(iv)}} Press Switch \\

\midrule
\multirow{3}{*}{D}
& Towel Folding 
& Both 
& \textbf{\textit{(i)}} First Fold 
$\rightarrow$ \textbf{\textit{(ii)}} Second Fold 
$\rightarrow$ \textbf{\textit{(iii)}} Final Fold \\

& Table Sweeping 
& GALAXEA 
& \textbf{\textit{(i)}} Grasp Broom 
$\rightarrow$ \textbf{\textit{(ii)}} Sweep Region A 
$\rightarrow$ \textbf{\textit{(iii)}} Sweep Region B 
$\rightarrow$ \textbf{\textit{(iv)}} Trash Inside Dustpan \\

& Medicine Bottling 
& GALAXEA 
& \textbf{\textit{(i)}} Grasp Medicine 
$\rightarrow$ \textbf{\textit{(ii)}} Place into Bottle 
$\rightarrow$ \textbf{\textit{(iii)}} Grasp Cap 
$\rightarrow$ \textbf{\textit{(iv)}} Screw on Cap \\

\midrule
\multirow{3}{*}{E}
& Bimanual Transfer 
& Both 
& \textbf{\textit{(i)}} Right-Arm Grasp 
$\rightarrow$ \textbf{\textit{(ii)}} Hand-over 
$\rightarrow$ \textbf{\textit{(iii)}} Place in Bowl \\

& Bimanual Bag Packing 
& GALAXEA 
& \textbf{\textit{(i)}} Hold Bag Open 
$\rightarrow$ \textbf{\textit{(ii)}} Insert Ball 
$\rightarrow$ \textbf{\textit{(iii)}} Close Bag \\

& Tray Transportation 
& GALAXEA 
& \textbf{\textit{(i)}} Dual-Arm Grasp 
$\rightarrow$ \textbf{\textit{(ii)}} Transport Tray 
$\rightarrow$ \textbf{\textit{(iii)}} Place Toy \\

\midrule
\multirow{3}{*}{F}
& Conveyor Interception 
& Both 
& \textbf{\textit{(i)}} Intercept First Target 
$\rightarrow$ \textbf{\textit{(ii)}} Place in Plate 
$\rightarrow$ \textbf{\textit{(iii)}} Intercept Second Target 
$\rightarrow$ \textbf{\textit{(iv)}} Place in Plate \\

& Rolling-Ball Interception 
& GALAXEA 
& \textbf{\textit{(i)}} Intercept Ball 
$\rightarrow$ \textbf{\textit{(ii)}} Stabilize Grasp 
$\rightarrow$ \textbf{\textit{(iii)}} Place in Bowl \\

& Dynamic Placement 
& GALAXEA 
& \textbf{\textit{(i)}} Grasp Banana 
$\rightarrow$ \textbf{\textit{(ii)}} Place Behind Apple \\

\midrule
\multirow{3}{*}{G}
& Stack Recovery 
& Both 
& \textbf{\textit{(i)}} Place Red Block 
$\rightarrow$ \textbf{\textit{(ii)}} Place Blue Block 
$\rightarrow$ \textbf{\textit{(iii)}} Recover Grasp 
$\rightarrow$ \textbf{\textit{(iv)}} Place Blue Block 
$\rightarrow$ \textbf{\textit{(v)}} Place Yellow Block \\

& Drawer Cleanup 
& GALAXEA 
& \textbf{\textit{(i)}} Open Drawer 
$\rightarrow$ \textbf{\textit{(ii)}} Insert Toy A 
$\rightarrow$ \textbf{\textit{(iii)}} Insert Toy B 
$\rightarrow$ \textbf{\textit{(iv)}} Close Drawer \\

& Folding Recovery 
& GALAXEA 
& \textbf{\textit{(i)}} Grasp T-shirt 
$\rightarrow$ \textbf{\textit{(ii)}} First Fold 
$\rightarrow$ \textbf{\textit{(iii)}} Second Fold 
$\rightarrow$ \textbf{\textit{(iv)}} Realign T-shirt 
$\rightarrow$ \textbf{\textit{(v)}} Final Fold \\

\bottomrule
\end{tabularx}
\end{table*}

%% file: Supp-Tables/21tasks_tag.tex
\begin{table*}[t]
\caption{
\textbf{Orthogonal task properties and their correspondence to UniMPA.}
The tags characterize complementary challenges that may appear across different real-world task suites rather than defining mutually exclusive categories.
}
\label{tab:real_tags}
\centering
\footnotesize
\setlength{\tabcolsep}{4.5pt}
\renewcommand{\arraystretch}{1.12}

\begin{tabularx}{\textwidth}{C{1.0cm} p{3.9cm} X}
\toprule
\textbf{Tag}
& \textbf{Property}
& \textbf{Description and Corresponding UniMPA Component} \\
\midrule

\tagTA
& Transition Ambiguity
& Similar observations or object configurations may correspond to different execution phases or intended future transitions, motivating transition-aware reasoning and temporally aligned memory retrieval rather than appearance-only matching. \\

\tagCT
& Critical Transition
& The task contains interaction-intensive events such as contact, grasping, releasing, insertion, articulated manipulation, or deformation, where the Trigger Gate activates fine-grained pixel-level future prediction. \\

\tagLP
& Long-Horizon Progress
& Successful execution requires persistent tracking of task phase and accumulated progress across temporally dependent manipulation stages, motivating persistent latent future modeling and temporally coherent retrieval. \\

\tagVR
& Visual-to-Action Retrieval
& Predicted visual transitions query the Visual-Action Memory Bank to retrieve historically executable visual-action experience associated with compatible state evolution. \\

\tagAR
& Action-to-Visual Retrieval
& Recent action evolution queries the Action-Visual Memory Bank to retrieve visually grounded and temporally compatible action prototypes for action-manifold refinement. \\

\tagBI
& Bimanual Coordination
& The task requires coordinated dual-arm execution, including role assignment, stabilization, handover, or synchronized object manipulation. \\

\tagDY
& Dynamic Interaction
& The task involves state changes that unfold during execution, such as moving-target interception or dynamically changing object relations, requiring future-transition anticipation and online action adjustment. \\

\tagRC
& Recovery
& The task contains controlled disturbances or intermediate execution deviations and requires the policy to resume task progress without restarting the complete manipulation sequence. \\

\bottomrule
\end{tabularx}
\end{table*}

%% file: Supp-Tables/21tasks_instruction.tex
\begin{table*}[t]
\centering
\caption{\textbf{Real-world task definitions and language instructions.}}
\label{tab:app_instruction}
\scriptsize
\setlength{\tabcolsep}{4.0pt}
\renewcommand{\arraystretch}{1.10}
    \begin{tabularx}{\textwidth}{C{0.45cm} p{3cm} C{2.5cm} X}
    \toprule
    \textbf{Suite} & \textbf{Task} & \textbf{Tags} & \textbf{Language Instruction} \\
    \midrule
    
    \multirow{3}{*}{A}
    
    & Fruit Placement
    & \tagCT, \tagVR, \tagAR
    & Pick up the apple and banana from the conveyor belt one by one and place both onto the plate. \\

    & Object Position Swapping
    & \tagTA, \tagLP, \tagAR
    & Swap the positions of the cup and the can without disturbing the remaining objects. \\
    
    & Semantic Table Clearing
    & \tagLP, \tagVR, \tagAR
    & Place the tool into the plate, the fruit into the bowl, and the trash into the bin. \\

    \midrule
    
    \multirow{3}{*}{B}
    
    & Microwave Manipulation
    & \tagTA, \tagCT, \tagLP, \tagVR, \tagAR
    & Place the bowl into the microwave and close the microwave door. \\

    & Pot Lid Replacement
    & \tagTA, \tagCT, \tagLP, \tagAR
    & Remove the lid, place the object into the pot, and replace the lid. \\

    & Water Pouring
    & \tagTA, \tagCT, \tagLP, \tagAR
    & Grasp the kettle, pour water into the bowl, and place the kettle back down. \\

    \midrule

    \multirow{3}{*}{C}

    & Block Stacking
    & \tagCT, \tagVR, \tagAR
    & Place the red block at the center of the table, then stack the blue and yellow blocks on top of it. \\

    & Ring Stacking
    & \tagTA, \tagCT, \tagAR
    & Place the large and small rings onto the pole one by one. \\

    & Plug Insertion and Activation
    & \tagCT, \tagTA, \tagVR, \tagAR
    & Grasp the plug, insert it into the socket, and press the switch to turn it on. \\
    
    \midrule

    \multirow{3}{*}{D}

    & Towel Folding
    & \tagTA, \tagCT, \tagLP, \tagAR
    & Fold the towel. \\

    & Table Sweeping
    & \tagCT, \tagLP, \tagAR
    & Use the broom to sweep the scattered trash on the table into the dustpan. \\

    & Medicine Bottling
    & \tagCT, \tagLP, \tagVR, \tagAR
    & Place the medicine into the bottle and screw the cap on securely. \\

    \midrule

    \multirow{3}{*}{E}

    & Bimanual Transfer
    & \tagTA, \tagCT, \tagAR, \tagBI
    & Use the right arm to pick up the tennis ball, transfer it to the left arm, and place it into the bowl. \\

    & Bimanual Bag Packing
    & \tagCT, \tagLP, \tagAR, \tagBI
    & Use one arm to hold the bag open while the other arm places the ball inside, then close the bag. \\

    & Tray Transportation
    & \tagLP, \tagAR, \tagBI
    & Grasp the tray with both arms, transport it to the marked location, and place the toy in it. \\

    \midrule

    \multirow{3}{*}{F}

    & Conveyor Interception
    & \tagCT, \tagVR, \tagAR, \tagDY
    & Wait until the block enters the pickup region, then pick it up and place it into the plate. \\

    & Rolling-Ball Interception
    & \tagCT, \tagAR, \tagDY
    & Intercept the rolling ball before it leaves the table and place it into the bowl. \\

    & Dynamic Placement
    & \tagLP, \tagVR, \tagAR, \tagDY
    & When the apple appears on the conveyor belt, place the banana behind it. \\

    \midrule

    \multirow{3}{*}{G}

    & Stack Recovery
    & \tagCT, \tagLP, \tagAR, \tagRC
    & Build a three-block tower in the order of red, blue, and yellow, and recover if the upper block is displaced during execution. \\

    & Drawer Cleanup
    & \tagTA, \tagLP, \tagVR, \tagAR
    & Open the drawer, place two toys inside one by one, and close the drawer. \\

    & Folding Recovery
    & \tagCT, \tagLP, \tagAR, \tagRC
    & Begin folding the T-shirt, recover from a controlled displacement, and complete the fold. \\

    \bottomrule
    \end{tabularx}
\end{table*}

%% file: Supp-Tables/supp-robotwinv2.tex
\begin{table*}[htbp]
\vspace{-4mm}
\caption{\textbf{Per-task results on RoboTwin 2.0 benchmark~\cite{chen2025robotwin}.}
All policies are trained on the Clean setting and evaluated without task-specific adaptation under the
\textbf{Randomized (hard) zero-shot} setting, with 100 rollouts per task.}
\vspace{-5pt}
\centering
\footnotesize
\label{tab:robotwin_random_allresults}

{
\setlength{\tabcolsep}{7.0pt}
\resizebox{\textwidth}{!}{
\begin{tabular}{l|ccccccccccc|c}
\toprule
Method
& \shortstack{Adjust\\Bottle}
& \shortstack{Beat Block\\Hammer}
& \shortstack{Blocks Rank.\\Size}
& \shortstack{Click\\Alarmclock}
& \shortstack{Click\\Bell}
& \shortstack{Dump Bin\\Bigbin}
& \shortstack{Grab\\Roller}
& \shortstack{Handover\\Block}
& \shortstack{Handover\\Mic}
& \shortstack{Lift\\Pot}
& \shortstack{Move Can\\Pot}
& Average \\
\midrule

ACT~\textit{[RSS'23]}~\cite{zhao2023act}
& 23\% & 3\% & 0\% & 4\% & 3\% & 1\% & 25\% & 0\% & 0\% & 0\% & 4\%
& 5.73\% \\

DP~\textit{[RSS'23]}~\cite{diff}
& 0\% & 0\% & 0\% & 5\% & 0\% & 0\% & 0\% & 0\% & 0\% & 0\% & 0\%
& 0.45\% \\

DP3~\textit{[RSS'24]}~\cite{ze20243d}
& 3\% & 8\% & 0\% & 14\% & 0\% & 53\% & 2\% & 0\% & 3\% & 0\% & 6\%
& 8.09\% \\

RDT~\textit{[ICLR'25]}~\cite{liu2024rdt}
& 75\% & \textbf{37\%} & 0\% & 12\% & 9\% & 32\% & 43\% & 14\% & 31\% & 9\% & 12\%
& 24.91\% \\

$\pi_{0}$~\textit{[RSS'25]}~\cite{pi0}
& 56\% & 21\% & 1\% & 11\% & 3\% & 24\% & 80\% & 8\% & 13\% & \textbf{36\%} & \textbf{21\%}
& 24.91\% \\

UP-VLA~\textit{[ICML'25]}~\cite{zhang2025up}
& 17\% & 16\% & 0\% & 41\% & 72\% & 35\% & 28\% & 0\% & 0\% & 0\% & 0\%
& 19.00\% \\

ADV~\textit{[arXiv'26]}~\cite{zhao2026action}
& 26\% & 0\% & 0\% & 22\% & 13\% & 15\% & 46\% & 6\% & 2\% & 9\% & 11\%
& 13.64\% \\

TwinVLA~\textit{[ICLR'26]}~\cite{im2025twinvla}
& 35\% & 10\% & 0\% & 1\% & 13\% & 34\% & 22\% & 0\% & 2\% & 7\% & 5\%
& 11.73\% \\

BagelVLA~\textit{[RSS'26]}~\cite{hu2026bagelvla}
& 14\% & 16\% & 2\% & 20\% & 35\% & 51\% & 41\% & 0\% & 8\% & 32\% & 0\%
& 19.91\% \\

HALO~\textit{[ICML'26]}~\cite{shou2026halo}
& 9\% & 11\% & 2\% & 14\% & 10\% & 28\% & 57\% & \textbf{36\%} & \textbf{61\%} & 34\% & 15\%
& 25.18\% \\

$\pi_{0.5}$~\textit{[CoRL'25]}~\cite{pi05}
& 54\% & 0\% & \textbf{11\%} & \textbf{62\%} & \textbf{85\%}
& 30\% & 63\% & 0\% & 2\% & 0\% & 0\%
& 27.91\% \\

\midrule
\rowcolor[HTML]{EFE9F7}
\textbf{UniMPA}
& \textbf{80\%} & 24\% & 4\% & 28\% & 24\%
& \textbf{62\%} & \textbf{84\%} & 12\% & 34\% & 8\% & 13\%
& \textbf{33.91\%} \\

\bottomrule
\end{tabular}
}
}

\vspace{2mm}

{
\setlength{\tabcolsep}{5.0pt}
\resizebox{\textwidth}{!}{
\begin{tabular}{l|ccccccccccc|c}
\toprule
Method
& \shortstack{Move Pillbottle\\Pad}
& \shortstack{Move Playingcard\\Away}
& \shortstack{Open\\Laptop}
& \shortstack{Open\\Microwave}
& \shortstack{Pick Diverse\\Bottles}
& \shortstack{Pick Dual\\Bottles}
& \shortstack{Place A2B\\Left}
& \shortstack{Place A2B\\Right}
& \shortstack{Place Bread\\Basket}
& \shortstack{Place Bread\\Skillet}
& \shortstack{Place Burger\\Fries}
& Average \\
\midrule

ACT~\textit{[RSS'23]}~\cite{zhao2023act}
& 0\% & 0\% & 0\% & 0\% & 0\% & 0\% & 0\% & 0\% & 0\% & 0\% & 0\%
& 0.00\% \\

DP~\textit{[RSS'23]}~\cite{diff}
& 0\% & 0\% & 0\% & 0\% & 0\% & 0\% & 0\% & 0\% & 0\% & 0\% & 0\%
& 0.00\% \\

DP3~\textit{[RSS'24]}~\cite{ze20243d}
& 0\% & 3\% & 7\% & 22\% & 1\% & 1\% & 2\% & 0\% & 1\% & 0\% & 18\%
& 5.00\% \\

RDT~\textit{[ICLR'25]}~\cite{liu2024rdt}
& 0\% & 11\% & 32\% & 20\% & 0\% & 13\% & 1\% & 1\% & 2\% & 1\% & 27\%
& 9.82\% \\

$\pi_{0}$~\textit{[RSS'25]}~\cite{pi0}
& 1\% & 22\% & 46\% & \textbf{50\%} & 6\% & 12\% & 1\% & 6\% & 4\% & 1\% & 4\%
& 13.91\% \\

UP-VLA~\textit{[ICML'25]}~\cite{zhang2025up}
& 7\% & 13\% & 21\% & 7\% & 18\% & 31\% & 4\% & 1\% & 20\% & 16\% & 26\%
& 14.91\% \\

ADV~\textit{[arXiv'26]}~\cite{zhao2026action}
& 1\% & 17\% & 27\% & 39\% & 2\% & 9\% & 0\% & 0\% & 2\% & 1\% & 11\%
& 9.91\% \\

TwinVLA~\textit{[ICLR'26]}~\cite{im2025twinvla}
& 2\% & 35\% & 17\% & 1\% & 8\% & 12\% & 5\% & 1\% & 3\% & 1\% & 13\%
& 8.91\% \\

BagelVLA~\textit{[RSS'26]}~\cite{hu2026bagelvla}
& 1\% & 30\% & 37\% & 0\% & \textbf{34\%} & \textbf{56\%}
& 12\% & 11\% & 29\% & \textbf{26\%} & 11\%
& 22.45\% \\

HALO~\textit{[ICML'26]}~\cite{shou2026halo}
& \textbf{26\%} & \textbf{53\%} & 37\% & 24\% & 17\% & 30\%
& 8\% & 9\% & 26\% & 23\% & 37\%
& 26.36\% \\

$\pi_{0.5}$~\textit{[CoRL'25]}~\cite{pi05}
& 20\% & 14\% & 3\% & 14\% & 14\% & 20\%
& 12\% & 6\% & \textbf{38\%} & 19\% & \textbf{45\%}
& 18.64\% \\

\midrule
\rowcolor[HTML]{EFE9F7}
\textbf{UniMPA}
& 15\% & 46\% & \textbf{67\%} & 24\% & 18\% & 25\%
& \textbf{20\%} & \textbf{14\%} & 31\% & 16\% & 40\%
& \textbf{28.73\%} \\

\bottomrule
\end{tabular}
}
}

\vspace{2mm}

{
\setlength{\tabcolsep}{5.0pt}
\resizebox{\textwidth}{!}{
\begin{tabular}{l|ccccccccccc|c}
\toprule
Method
& \shortstack{Place Can\\Basket}
& \shortstack{Place Cans\\Plasticbox}
& \shortstack{Place Container\\Plate}
& \shortstack{Place Dual\\Shoes}
& \shortstack{Place Empty\\Cup}
& \shortstack{Place\\Fan}
& \shortstack{Place Object\\Basket}
& \shortstack{Place Object\\Scale}
& \shortstack{Place Object\\Stand}
& \shortstack{Place Phone\\Stand}
& \shortstack{Place\\Shoe}
& Average \\
\midrule

ACT~\textit{[RSS'23]}~\cite{zhao2023act}
& 0\% & 0\% & 1\% & 0\% & 0\% & 0\% & 0\% & 0\% & 0\% & 0\% & 0\%
& 0.09\% \\

DP~\textit{[RSS'23]}~\cite{diff}
& 0\% & 0\% & 0\% & 0\% & 0\% & 0\% & 0\% & 0\% & 0\% & 0\% & 0\%
& 0.00\% \\

DP3~\textit{[RSS'24]}~\cite{ze20243d}
& 2\% & 3\% & 1\% & 0\% & 1\% & 1\% & 0\% & 0\% & 0\% & 2\% & 2\%
& 1.09\% \\

RDT~\textit{[ICLR'25]}~\cite{liu2024rdt}
& 6\% & 5\% & 17\% & 4\% & 7\% & 2\% & 17\% & 0\% & 5\% & 6\% & 7\%
& 6.91\% \\

$\pi_{0}$~\textit{[RSS'25]}~\cite{pi0}
& 5\% & 2\% & 45\% & 0\% & 11\% & 10\% & 2\% & 0\% & 11\% & 7\% & 6\%
& 9.00\% \\

UP-VLA~\textit{[ICML'25]}~\cite{zhang2025up}
& 0\% & 24\% & 48\% & 0\% & 27\% & 1\% & 1\% & 4\% & 24\% & 0\% & 12\%
& 12.82\% \\

ADV~\textit{[arXiv'26]}~\cite{zhao2026action}
& 2\% & 0\% & 11\% & 0\% & 3\% & 1\% & 4\% & 0\% & 0\% & 6\% & 2\%
& 2.64\% \\

TwinVLA~\textit{[ICLR'26]}~\cite{im2025twinvla}
& 0\% & 8\% & 4\% & 3\% & 1\% & 0\% & 3\% & 0\% & 2\% & 2\% & 4\%
& 2.45\% \\

BagelVLA~\textit{[RSS'26]}~\cite{hu2026bagelvla}
& 0\% & 5\% & 58\% & 0\% & 34\% & 5\% & 3\% & 0\% & 21\% & 2\% & 29\%
& 14.27\% \\

HALO~\textit{[ICML'26]}~\cite{shou2026halo}
& \textbf{34\%} & \textbf{47\%} & 22\% & 3\% & 28\%
& 9\% & \textbf{25\%} & 5\% & 33\% & 10\% & 18\%
& 21.27\% \\

$\pi_{0.5}$~\textit{[CoRL'25]}~\cite{pi05}
& 7\% & 27\% & 58\% & 3\% & 53\% & 8\%
& \textbf{25\%} & 20\% & 45\% & 7\% & 41\%
& 26.73\% \\

\midrule
\rowcolor[HTML]{EFE9F7}
\textbf{UniMPA}
& 17\% & 18\% & \textbf{75\%} & \textbf{7\%} & \textbf{57\%}
& \textbf{19\%} & 16\% & \textbf{38\%} & \textbf{59\%}
& \textbf{37\%} & \textbf{49\%}
& \textbf{35.64\%} \\

\bottomrule
\end{tabular}
}
}

\vspace{2mm}

{
\setlength{\tabcolsep}{6.0pt}
\resizebox{\textwidth}{!}{
\begin{tabular}{l|ccccccccccc|c}
\toprule
Method
& \shortstack{Press\\Stapler}
& \shortstack{Put Bottles\\Dustbin}
& \shortstack{Put Object\\Cabinet}
& \shortstack{Rotate\\QRCode}
& \shortstack{Scan\\Object}
& \shortstack{Shake Bottle\\Horiz.}
& \shortstack{Shake\\Bottle}
& \shortstack{Stack Bowls\\Three}
& \shortstack{Stack Bowls\\Two}
& \shortstack{Stamp\\Seal}
& \shortstack{Turn\\Switch}
& Average \\
\midrule

ACT~\textit{[RSS'23]}~\cite{zhao2023act}
& 6\% & 1\% & 0\% & 0\% & 0\% & 4\% & 10\% & 0\% & 0\% & 0\% & 2\%
& 2.09\% \\

DP~\textit{[RSS'23]}~\cite{diff}
& 0\% & 0\% & 0\% & 0\% & 0\% & 18\% & 8\% & 0\% & 0\% & 0\% & 1\%
& 2.45\% \\

DP3~\textit{[RSS'24]}~\cite{ze20243d}
& 3\% & \textbf{21\%} & 1\% & 1\% & 1\% & 25\% & 19\% & 5\% & 6\% & 0\% & 8\%
& 8.18\% \\

RDT~\textit{[ICLR'25]}~\cite{liu2024rdt}
& 24\% & 4\% & \textbf{18\%} & 5\% & 1\% & 51\% & 45\% & 17\% & 30\% & 0\% & 15\%
& 19.09\% \\

$\pi_{0}$~\textit{[RSS'25]}~\cite{pi0}
& 29\% & 13\% & \textbf{18\%} & 15\% & 1\% & 51\% & 60\% & 24\% & 41\% & 4\% & 23\%
& 25.36\% \\

UP-VLA~\textit{[ICML'25]}~\cite{zhang2025up}
& 56\% & 0\% & 0\% & 2\% & 23\% & 68\% & 54\% & 1\% & 12\% & 2\% & 26\%
& 22.18\% \\

ADV~\textit{[arXiv'26]}~\cite{zhao2026action}
& 15\% & 9\% & 0\% & 9\% & 0\% & 42\% & 37\% & 26\% & 48\% & 0\% & 9\%
& 17.73\% \\

TwinVLA~\textit{[ICLR'26]}~\cite{im2025twinvla}
& 26\% & 4\% & 16\% & 3\% & 4\% & 55\% & 58\% & 15\% & 11\% & 1\% & 15\%
& 18.91\% \\

BagelVLA~\textit{[RSS'26]}~\cite{hu2026bagelvla}
& 58\% & 10\% & 0\% & 21\% & \textbf{32\%} & 73\% & 74\% & 13\% & 52\% & 8\% & 30\%
& 33.73\% \\

HALO~\textit{[ICML'26]}~\cite{shou2026halo}
& 64\% & 13\% & 8\% & 11\% & 24\% & 66\% & 73\% & 25\% & 49\% & 21\% & 27\%
& 34.64\% \\

$\pi_{0.5}$~\textit{[CoRL'25]}~\cite{pi05}
& \textbf{72\%} & 10\% & 7\% & 10\% & 7\% & 94\%
& \textbf{96\%} & 24\% & 52\% & 15\% & 20\%
& 37.00\% \\

\midrule
\rowcolor[HTML]{EFE9F7}
\textbf{UniMPA}
& 67\% & 18\% & 15\% & \textbf{22\%} & 20\%
& \textbf{95\%} & 94\% & \textbf{38\%} & \textbf{69\%}
& \textbf{24\%} & \textbf{34\%}
& \textbf{45.09\%} \\

\bottomrule
\end{tabular}
}
}

\vspace{-3pt}
\end{table*}